\documentclass{article}
\usepackage[preprint]{neurips_2026}
\newif\ifanon
\anonfalse
\usepackage[T1]{fontenc}
\usepackage{graphicx}
\usepackage{textcomp}
\usepackage{amsmath,amssymb}
\usepackage{amsthm}
\usepackage{booktabs}
\usepackage{mathtools}
\usepackage{tikz}
\usetikzlibrary{arrows.meta,positioning,fit}
\usepackage{subcaption}
\usepackage{url}

\theoremstyle{plain}
\newtheorem{theorem}{Theorem}
\newtheorem{lemma}{Lemma}
\newtheorem{proposition}{Proposition}
\newtheorem{corollary}{Corollary}
\theoremstyle{definition}
\newtheorem{definition}{Definition}
\newtheorem{example}{Example}
\theoremstyle{remark}
\newtheorem{remark}{Remark}
\providecommand{\qed}{\hfill\qedsymbol}
\providecommand{\Description}[1]{}
\DeclareMathSymbol{\Gamma}{\mathalpha}{letters}{"00}
\DeclareMathSymbol{\Delta}{\mathalpha}{letters}{"01}
\DeclareMathSymbol{\Theta}{\mathalpha}{letters}{"02}
\DeclareMathSymbol{\Lambda}{\mathalpha}{letters}{"03}
\DeclareMathSymbol{\Xi}{\mathalpha}{letters}{"04}
\DeclareMathSymbol{\Pi}{\mathalpha}{letters}{"05}
\DeclareMathSymbol{\Sigma}{\mathalpha}{letters}{"06}
\DeclareMathSymbol{\Upsilon}{\mathalpha}{letters}{"07}
\DeclareMathSymbol{\Phi}{\mathalpha}{letters}{"08}
\DeclareMathSymbol{\Psi}{\mathalpha}{letters}{"09}
\DeclareMathSymbol{\Omega}{\mathalpha}{letters}{"0A}

\definecolor{maxcolor}{RGB}{0,90,160}
\definecolor{mincolor}{RGB}{170,40,90}
\definecolor{ourcolor}{RGB}{0,125,70}
\newcommand{\ours}[1]{\textcolor{ourcolor}{#1}}

\newcommand{\R}{{\mathbb{R}}}
\newcommand{\Q}{{\mathbb{Q}}}
\newcommand{\N}{{\mathbb{N}}}
\newcommand{\Rbar}{{\overline{\mathbb{R}}}}
\newcommand{\Dist}{\Delta}
\newcommand{\dirac}[1]{\mathsf{Dirac}(#1)}
\newcommand{\Prob}{{\mathbb{P}}}
\newcommand{\E}{{\mathbb{E}}}
\newcommand{\Laws}{\mathcal{D}(\Rbar)}
\newcommand{\law}{{\mathcal{L}}}
\newcommand{\lawset}{{\mathbb{L}}}

\newcommand{\push}[1]{(S_{#1})_{\#}}
\newcommand{\rmdp}{{\mathcal{M}}}
\newcommand{\sg}{{\mathcal{G}}}
\newcommand{\fwd}[1]{\mathfrak{G}(#1)}
\newcommand{\bwd}[1]{\mathfrak{R}(#1)}
\newcommand{\pmax}{{\mathsf{max}}}
\newcommand{\pmin}{{\mathsf{min}}}
\newcommand{\Smax}{{S_{\pmax}}}
\newcommand{\Smin}{{S_{\pmin}}}
\newcommand{\Amax}{{A_{\pmax}}}
\newcommand{\Amin}{{A_{\pmin}}}
\newcommand{\Oa}{{\mathcal{O}_{\mathsf{a}}}}
\newcommand{\Oe}{{\mathcal{O}_{\mathsf{e}}}}
\newcommand{\Omax}{{\mathcal{O}_{\pmax}}}
\newcommand{\Omin}{{\mathcal{O}_{\pmin}}}
\newcommand{\Za}{{\mathcal{Z}_{\mathsf{a}}}}
\newcommand{\Ze}{{\mathcal{Z}_{\mathsf{e}}}}
\newcommand{\Zmax}{{\mathcal{Z}_{\pmax}}}
\newcommand{\Zmin}{{\mathcal{Z}_{\pmin}}}
\newcommand{\Ha}{{\mathcal{H}_{\mathsf{a}}}}
\newcommand{\He}{{\mathcal{H}_{\mathsf{e}}}}

\newcommand{\Unc}{{\mathcal{P}}}
\newcommand{\anew}{\top}
\newcommand{\Tot}{\mathsf{Tot}}
\newcommand{\Disc}{\mathsf{Disc}}
\newcommand{\LimAvg}{\mathsf{LimAvg}}
\newcommand{\MV}{\mathsf{MV}}
\newcommand{\CVaR}{\mathsf{CVaR}}
\newcommand{\ERisk}{\mathsf{ERisk}}

\newcommand{\Var}{\mathsf{Var}}
\newcommand{\Ber}{\mathsf{Ber}}
\newcommand{\Val}{\mathsf{Val}}
\newcommand{\Cyl}{\mathsf{Cyl}}
\newcommand{\proj}{\mathsf{proj}}
\newcommand{\Pclass}{\mathsf{P}}
\newcommand{\NP}{\mathsf{NP}}
\newcommand{\coNP}{\mathsf{coNP}}
\newcommand{\PSPACE}{\mathsf{PSPACE}}
\newcommand{\EXPTIME}{\mathsf{EXPTIME}}
\newcommand{\NEXPTIME}{\mathsf{NEXPTIME}}

\newcommand{\conv}{\mathrm{conv}}

\usepackage[most]{tcolorbox}
\usepackage{multirow}
\usepackage{enumitem}
\usepackage[hidelinks]{hyperref}

\title{Reward Objectives for Robust (PO)MDPs: A Law-Preserving Equivalence with (Partially Observable) Games}

\ifanon
\author{Anonymous Author(s)}
\else
\author{%
  Durgam Latha \\
  Indian Institute of Technology Bombay \\
  Mumbai, India \\
  \And
  Dion Reji \\
  Indian Institute of Technology Bombay \\
  Mumbai, India \\
  \And
  {\DJ}or{\dj}e {\v{Z}}ikeli{\'c} \\
  Nanyang Technological University \\
  Singapore, Singapore \\
  \And
  Shankaranarayanan Krishna \\
  Indian Institute of Technology Bombay \\
  Mumbai, India
}
\fi

\begin{document}

\maketitle

\begin{abstract}
Robust partially observable Markov decision processes model agents
that act under partial observability while adversarial environment resolves uncertainty in the transition probabilities. Reward objectives such as discounted-sum, long-run average and cumulative reward under finite and infinite horizons allow quantifying the utility of an agent. We consider the problem of solving RPOMDPs under all four reward objectives, where the goal is to decide whether the  value of a risk measure with respect to a reward objective exceeds a given threshold. We consider different risk measures, such as (1)~expected value, (2)~mean-variance (MV), (3)~conditional value-at-risk (CVaR), and (4)~entropic risk (ERisk). Our results can be summarized as follows. First, we formalize the class of {\em law-invariant risk measures}, that subsume all four risk measures above. Second, we propose a {\em unified framework for solving RPOMDPs} with discounted-sum, long-run average and cumulative reward objectives under any law-invariant risk measure, by reducing the problem to solving partially observable stochastic games (POSGs). Our unified framework applies to $(s,a)$-rectangular RPOMDPs with polytopic uncertainty sets, which we show to be {\em equivalent} to POSGs, for each of the four reward objectives and any law-invariant risk measure. We prove equivalence by designing logspace reductions between the RPOMDP and POSG models, that map strategies to strategies and preserve the distribution of the reward objective, up to a constant rescaling
factor. Third, as a corollary, we derive new computational complexity results for R(PO)MDPs and POSGs. In particular, we show that solving $(s,a)$-rectangular RMDPs is as hard as solving simple stochastic games.
\end{abstract}

\section{Introduction}
\label{sec:intro}

Partially observable Markov decision processes (POMDPs) are the standard model of an agent acting under partial observability in the presence of stochastic uncertainty~\cite{Kaelbling1998}. The key limitation of POMDPs, however, is that their transition probabilities are assumed to be fully known. This is a strong assumption in the settings where transition probabilities are e.g.~estimated from data. Robust MDPs (RMDPs)~\cite{Iyengar2005,Nilim2005,Wiesemann2013} and, more generally, robust POMDPs (RPOMDPs)~\cite{Osogami2015,Bovy2024}, account for this by replacing fully known transition probabilities with an uncertainty set of transition probabilities, with the adversarial environment being able to choose transition probabilities from this uncertainty set. Recent years have seen significant interest in solving R(PO)MDPs with reward objectives, such as discounted-sum reward~\cite{NilimG03,Iyengar2005,KaufmanS13,TamarMX14,GhavamzadehPC16,HoPW21,AsadiCGKMP26arxiv} or long-run average reward objectives~\cite{TewariB07,GC2023,GoyalG23,WangVAPZ23,Chatterjee2024,MeggendorferWW25}. The goal in solving an RPOMDP is to compute a strategy that maximizes the expected value of the reward objective. Existing works largely focused on designing efficient algorithms for solving RMDPs. Solving RPOMDPs, which also account for partial observability, has been considered in~\cite{Bovy2024,Osogami2015,ChamieM18,Suilen0CT20,NakaoJS21,Cubuktepe0JMST21,GaleslootA0J025,krale2025on}.

These works all focus on computing a strategy that maximizes the {\em expected value} $(\E)$ of the reward objective. However, while providing valuable information on the average-case performance of the strategy, expected value alone does not consider the variability of possible outcomes. Two strategies that achieve the same expected reward objective value may carry very different risks. This is why expected value is sometimes replaced by other risk measures such as mean-variance~\cite{MannorTsitsiklis2013}, conditional value-at-risk (CVaR)~\cite{Rockafellar2000,Chow2015}, or entropic risk~\cite{HowardMatheson1972,BaierEntropicRisk2023}.
However, computing strategies that optimize these risk measures has so far mostly been studied for MDPs, but not for robust models such as RMDPs or RPOMDPs. The goal of this work is to address this gap.

We present a {\em unified framework for solving RPOMDPs with reward objectives and different risk measures}, by reducing the problem to solving partially observable stochastic games (POSGs). Our framework applies to discounted-sum, long-run average and total reward under finite and infinite horizons objectives, and under any law-invariant risk measure. A risk measure is said to be {\em law-invariant}, if it only depends on the probability distribution of the value of the reward objective over RPOMDP runs. We show that common risk measures such as (1) expected value ($\E$), (2) mean-variance (MV),
(3) conditional value-at-risk (CVaR), and (4) entropic risk (ERisk) are all law-invariant. Hence, our unified framework applies to all these risk measures.

We consider $(s,a)$-rectangular RPOMDPs with polytopic uncertainty sets, which we show to be {\em equivalent} to POSGs. For each of the four reward objectives and under any law-invariant risk measure, we design two logspace reductions: one from RPOMDPs to POSGs, and one from POSGs to RPOMDPs. Our reductions map strategies to strategies and are {\em law preserving}, meaning that the image of every strategy achieves the same distribution of the reward objective value (up to the reward objective being normalized by a constant rescaling factor), thus mapping optimal strategies to optimal strategies under any law-invariant risk measure. Hence, we can reduce the problem of solving an RPOMDP to solving a POSG under the same reward objective and risk measure and vice-versa.  In the case of RMDPs (with full observability), our reductions establish equivalence with turn-based stochastic games (TBSGs)~\cite{Shapley1953, Condon1992}.

Building upon our equivalence result, we study the \emph{threshold problems} for law-invariant risk measures such as $\E$, $\MV$, $\CVaR$, and $\ERisk$ with rational parameters, and for the finite and infinite horizon total, long-run average and discounted-sum reward objectives. We consider the non-strict and strict threshold problems: given a model (RPOMDP/RMDP/POSG/TBSG), a risk measure $\Psi$, a reward objective $X$, and a rational $t$, decide whether the agent in the RPOMDP/RMDP (or the maximizing player in POSG/TBSG) can guarantee a value of $\Psi$ on the reward objective $X$ that is greater than $t$ (strict threshold), or greater or equal than $t$ (non-strict threshold).  Table~\ref{tab:results} summarizes our novel complexity results. Under partial observation, both variants are $\PSPACE$-hard and decidable for finite horizon total reward, and undecidable for infinite horizon total reward and for long-run average reward objectives. For discounted-sum rewards, values can be approximated to any precision, which makes the strict problem recursively enumerable; it is nevertheless undecidable for $\E$ and $\CVaR$, while for discounted mean-variance and entropic risk the decidability of the threshold problems remains open.
Under full observation, both variants are SSG-hard for RMDPs with total or long-run average rewards for all four risk measures, and for discounted-sum rewards for the expected value. For the expected discounted-sum and the long-run average reward, the SSG bound is tight, since both problems lie in $\NP\cap\coNP$. Here, SSG stands for the complexity of solving simple stochastic games, which is known to be in $\NP\cap\coNP$~\cite{Condon1992} but the question of whether there exists a polynomial algorithm is an open problem.

Our contributions can be summarized as follows:
\begin{enumerate}
    \item {\em Law-invariant risk measures.} We formalize law-invariant risk measures for reward objectives in RPOMDPs. These include (1) expected value ($\E$), (2) mean-variance (MV), (3) conditional value-at-risk (CVaR), and (4) entropic risk (ERisk).
    \item {\em Equivalence of RPOMDPs and POSGs.}  We present a {\em unified framework for solving RPOMDPs with reward objectives and law-invariant risk measures}, by reducing the problem to solving POSGs. Our framework applies to $(s,a)$-rectangular polytopic RPOMDPs with discounted-sum, long-run average, total and finite-horizon reward objectives. It is based on our novel {\em equivalence result} between RPOMDPs and POSGs obtained via logspace reductions in both directions. In the case of RMDPs (with full observability), we obtain equivalence with TBSGs.
    \item {\em Uniform complexity landscape.} We derive new computational complexity results and provide a complexity landscape for RPOMDPs (and POSGs) under reward objectives and law-invariant risk measures.
\end{enumerate}

\section{Preliminaries}
\label{sec:prelim}

We write $\N=\{0,1,2,\dots\}$, $\Q$, and $\R$ for the sets of natural, rational, and real numbers.
For a finite set $Y$, we use $\Dist(Y)$ to denote the set of probability distributions over $Y$,
and $\dirac{y}$ the distribution with all probability mass on $y \in Y$. We write
$\Rbar=[-\infty,+\infty]$ and $\Laws$ for the set of Borel probability measures
on $\Rbar$, which we call \emph{laws}. For a random variable
$X\colon\Omega\to\Rbar$ on a probability space $(\Omega,\mathcal{F},\Prob)$ and a Borel set
$B\subseteq\Rbar$, let $X^{-1}(B)=\{\omega\in\Omega : X(\omega)\in B\}$ be the set
of outcomes that $X$ maps into $B$. The \emph{law} of $X$ is the measure
$\law_X(B)=\Prob(X^{-1}(B))=\Prob(X\in B)$. For $c>0$ and a law $\mu$, we write $\push{c}\mu$ for the law of $cX$ when $X$ has law $\mu$, i.e., $\push{c}\mu(B)=\mu(\{x : cx\in B\})$ for every Borel set $B$.

\subsection{Robust POMDPs}
\label{sec:rpomdp}

\paragraph{RPOMDPs.} A \emph{robust POMDP} (RPOMDP) is a tuple
\[
\rmdp=(S,A,\Unc,s_0,\Oa,\Oe,\Za,\Ze,r),
\]
where $S$ and $A$ are finite sets of \emph{states} and \emph{actions},
$s_0\in S$ is the \emph{initial state}, and $\Oa$ and $\Oe$ are finite sets of
\emph{observations} of the agent and of the environment, with
\emph{observation functions} $\Za\colon S\to\Oa$ and $\Ze\colon S\to\Oe$. The
\emph{reward function} is $r\colon S\times A\times S\to\Q$. The \emph{uncertainty set} is given by \( \Unc \subseteq (S \times A \rightarrow \Dist(S)) \).

A \emph{POMDP} is an RPOMDP in which $\Unc$ is a singleton, specifying for each state-action pair in $S \times A$ a single probability distribution over successor states. An \emph{RMDP} is an
RPOMDP in which $\Za$ and $\Ze$ are injective, i.e., every observation
corresponds to a unique state. An \emph{MDP} is an RPOMDP that satisfies both conditions above.

\paragraph{Assumptions.} As is standard in the algorithmic study of robust Markov models~\cite{Iyengar2005,Wiesemann2013,Chatterjee2024,Latha2026}, we consider \emph{$(s,a)$-rectangular polytopic} uncertainty sets of the form $\Unc = \prod_{(s,a)\in S \times A} \Unc{(s,a)}$, where the environment  chooses transition probabilities independently
across state-action pairs, with each $\Unc{(s,a)} \subseteq \Dist(S)$ being a {\em polytope} with a finite vertex set \(V_{s,a}\).
$\Unc(s,a) = \{\sum_{i=1}^{|V_{s,a}|} \alpha_i v_i \;|\; \alpha_i \in [0,1], \sum_i \alpha_i = 1, v_i \in V_{s,a}\},$  each vertex \(v \in V_{s,a}\) has the form \((x_1,\dots,x_{|S|})\) with  \(x_i \geq 0 \text{ and }\sum_i x_i = 1\).  We assume that the uncertainty set $\Unc$ is given in the vertex set representation.

\paragraph{Semantics.} The RPOMDP starts in $s_0$ and proceeds in steps
$t=0,1,\dots$ to generate a random infinite play. If at time $t$ the RPOMDP is in
state $s_t$, the agent receives the observation $\Za(s_t)$ and chooses an action
$a_t\in A$. The environment receives the observation $\Ze(s_t)$ and action $a_t$ and selects a probabilistic transition function $P_t\in\Unc$; the next state $s_{t+1}$ is drawn from $P_t(s_t,a_t)$,
and the transition earns the reward $r_t=r(s_t,a_t,s_{t+1})$. This follows the RPOMDP game semantics of \cite{Bovy2024} for dynamic uncertainty, in which the
environment may choose a new assignment in every step.

The semantics are formalized via agent and environment strategies. A {\em history} is a finite sequence $h_t = s_0,a_0, s_1, \dots, s_t$ $\in (S \times A)^\star$. An \emph{agent observation history} for history $h_t$ is
a sequence $h=o^a_0o^a_1\cdots o^a_t\in\Oa^+$ of observations of the agent such that $o^a_t = \Za(s_t)$, and let $\Ha$ denote the set of agent observation histories. Then, an
\emph{agent strategy} is a map $\sigma\colon\Ha\to\Dist(A)$. Likewise, an
\emph{environment observation history} of $h_t$ is a sequence $h=o^e_0o^e_1\cdots o^e_t\in\Oe^+$ such that $o^e_t = \Ze(s_t)$ of
observations of the environment, and let $\He$ denote the set of environment observation histories. Then, an \emph{environment strategy} is a map
$\pi\colon\He\times A\to\Unc$. We write $\Sigma_\rmdp$ and $\Pi_\rmdp$ for the sets of
agent and environment strategies. A \emph{play} is an infinite sequence
$\rho=s_0a_0s_1a_1\cdots$, and its \emph{reward sequence} is
$(r_t)_{t\ge0}$ with $r_t=r(s_t,a_t,s_{t+1})$. A strategy profile $(\sigma,\pi)$
induces a unique probability measure $\Prob^{\sigma,\pi}_\rmdp$ on plays
\cite{Puterman1994} under which a prefix $s_0a_0\cdots s_t$ is extended by
$a_ts_{t+1}$ with probability
\[
\sigma(h^{\mathsf a}_t)(a_t)\cdot\pi(h^{\mathsf e}_t)(s_t,a_t)(s_{t+1}),
\]
where $h^{\mathsf a}_t=\Za(s_0)\cdots\Za(s_t)$ is the history of the agent and
$h^{\mathsf e}_t=\Ze(s_0)\cdots\Ze(s_t)$ is the history of the environment.

Without loss of generality, we adopt a standard assumption for RPOMDPs (and, later, for POSGs) that every action in the action alphabet of a decision maker is enabled at every state.
We call an action \emph{effective} at a state if the construction explicitly specifies its transition there, and an \emph{effective edge} is a transition of positive probability under an effective action. In figures, we draw only effective actions and edges. Every omitted state-action pair is implicitly completed by a transition to an absorbing sink that is worst possible for the decision maker who chose the action.

\subsection{Partially Observable Stochastic Games}
\label{sec:posg}

\paragraph{POSGs.} A \emph{turn-based partially observable stochastic game} (POSG) between the players $\pmax$ and $\pmin$ is a tuple
\[
\sg=(\Smax,\Smin,\Amax,\Amin,\delta,s_0,\Omax,\Omin,\Zmax,\Zmin,r)
\]
with disjoint finite sets $\Smax,\Smin$ of $\pmax$- and $\pmin$-states ($S=\Smax\cup\Smin$), an initial state $s_0\in\Smax$, action sets $\Amax,\Amin$ with $A = \Amax\cup \Amin$,
a transition function $\delta: (\Smax \times \Amax) \cup (\Smin \times \Amin) \rightarrow \Dist(S)$, observation functions $\Zmax\colon S\to\Omax$ and $\Zmin\colon S\to\Omin$, and a reward function $r\colon S\times(\Amax\cup\Amin)\times S\to\Q$. The game is \emph{alternating-control} (an \emph{ac-POSG}) if $\delta(s,a)\in\Dist(\Smin)$ for all $s\in\Smax$ and $\delta(s,a)\in\Dist(\Smax)$ for all $s\in\Smin$. It is \emph{fully observable} (a turn-based stochastic game (\emph{TBSG})) if $\Zmax$ and $\Zmin$ are injective, i.e., every observation corresponds to a unique state.

A \emph{play} is an infinite sequence $s_0a_0s_1a_1\cdots$ and its \emph{reward sequence} is $(r_t)_{t\ge0}$
with $r_t=r(s_t,a_t,s_{t+1})$. A {\em history} is a finite sequence $h_t = s_0,a_0,s_1,\cdots , s_t \in (S\times A)^\star$. A \emph{$\pmax$-history} of $h_t$ is a sequence $h=o^1_0\cdots o^1_t\in\Omax^+$ of observations of $\pmax$ such that $s_t$ is a max-state and $o^1_k = \Zmax(s_k),$, and let $H_{max}$ denote the set of max histories. A \emph{$\pmax$-strategy} is a map $\sigma\colon H_{max}\to\Dist(\Amax)$. A \emph{$\pmin$-history} of $h_t$ is a sequence $h = o^2_0\cdots o^2_t\in\Omin^+$ of observations of $\pmin$ such that $s_t$ is a $\pmin$-state and $o^2_k = \Zmin(s_k)$, and let $H_{min}$ denote the set of min histories. A \emph{$\pmin$-strategy} is a map $\pi\colon H_{min}\to\prod_{s\in\Smin}\Dist(\Amin)$, which assigns to every min-history $h$ a \emph{prescription} $\pi(h)$, i.e., a probability distribution over actions $\pi(h)(s)\in\Dist(\Amin)$ for every $\pmin$-state $s$. Then, if the current state is $s$, $\pmin$ samples the next action according to $\pi(h)(s)$. We write $\Sigma_\sg$ and $\Pi_\sg$ for the sets of $\pmax$- and $\pmin$-strategies. A profile $(\sigma,\pi)$ induces a unique probability measure $\Prob^{\sigma,\pi}_\sg$ on plays under which a prefix $s_0a_0\cdots s_k$ is extended by $a_ks_{k+1}$ with probability
\[
\begin{cases}
\sigma(h^{\pmax}_k)(a_k)\cdot\delta(s_k,a_k)(s_{k+1}), & \text{if }s_k\in\Smax,\\
\pi(h^{\pmin}_k)(s_k)(a_k)\cdot\delta(s_k,a_k)(s_{k+1}), & \text{if }s_k\in\Smin,
\end{cases}
\]
where $h^{\pmax}_k=\Zmax(s_0)\cdots\Zmax(s_k)$, $h^{\pmin}_k=\Zmin(s_0)\cdots\Zmin(s_k)$ are observation histories of $\pmax$ and $\pmin$, respectively.

\begin{remark}[Definition of $\pmin$-strategies]
\label{rem:prescriptions}
We let a $\pmin$-strategy assign a probability distribution over actions to every $\pmin$-state, so
that $\pmin$ may play differently in states with the same observation. This
is done in order to match the RPOMDP game semantics of~\cite{Bovy2024}, where the environment chooses a probabilistic transition function for each state-action pair.

While this may appear to give additional power to the $\pmin$-player as opposed to the classical POSGs~\cite{Bovy2024,DBLP:journals/fmsd/Chatterjee0H13} in which $\pmin$-strategies are of type $\pi\colon H_{min}\to\Dist(\Amin)$, we show in Appendix~\ref{app:compile} that a POSG with prescriptions as $\pmin$-strategies can be compiled into an equivalent classical POSG with $\pmin$-strategies of type $\pi\colon H_{min}\to\Dist(\Amin)$, by a constant scaling of reward objective values.
\end{remark}

\subsection{Reward Objectives}

We now define the reward objectives in RPOMDPs and POSGs that we consider in this work. Consider a play $\rho=s_0a_0s_1a_1\cdots$ in an RPOMDP or a POSG, and let $(r_t)_{t \geq 0}$ with $r_t = r(s_t,a_t,s_{t+1})$ be the reward sequence of the play. Note that, since our RPOMDPs and POSGs consist of finitely many states and actions, the reward function $r$ is bounded and all rewards lie in some $[-R,R]$. We define the following four reward objectives:
\begin{align*}
\Tot_N&=\textstyle\sum_{t<N}r_t, & \Tot&=\textstyle\sum_{t\ge0}r_t,\\
\Disc_\gamma&=\textstyle\sum_{t\ge0}\gamma^tr_t, & \LimAvg&=\liminf_{n\to\infty}\tfrac1n\textstyle\sum_{t<n}r_t,
\end{align*}
where $N\in\N$ is a horizon and $\gamma\in(0,1)$ is a rational discount factor. We refer to $\Tot_N$ as the {\em total reward until time $N$}, $\Tot$ as the {\em total reward}, $\Disc_\gamma$ as the {\em discounted-sum reward}, and $\LimAvg$ as the {\em long-run average reward (or mean-payoff)} of the play. The total reward $\Tot$ is restricted to nonnegative rewards in $[0,R]$ so that its value is well defined, whereas the other three reward objectives are defined for all rewards in $[-R,R]$.

We note that for long-run average rewards, while our presentation considers the $\liminf$ variant, our equivalence proof for RPOMDPs and POSGs is analogous for the $\limsup$ variant and the same computational complexity results hold.

\section{Law-Invariant Risk Measures} \label{sec:lawinv}

In this work, we study the problem of solving RPOMDPs under the reward
objectives defined above, where the goal is to calculate the value of a \emph{risk measure of the reward objective}. To that end, we first formally define a class of risk measures that we consider. We then define the problem of solving RPOMDPs (and, analogously, POSGs) under these \emph{risk measures of the above reward objectives}.

Let $\Omega$ be the set of plays in an RPOMDP $\rmdp$ or a POSG $\sg$, and let $X: \Omega \rightarrow \mathbb{R}$ be a reward objective which is either the total reward until time $N$ ($X = \Tot_N$), the total reward ($X = \Tot$), the discounted-sum reward ($X = \Disc_\gamma$), or the long-run average reward ($X = \LimAvg$). For a fixed strategy profile $(\sigma,\pi)$, $X$ induces a random variable defined by the reward objective value of a random play. We denote by $\law^{\sigma,\pi}_X$ the {\em law of $X$} under $(\sigma,\pi)$, i.e.~the probability distribution over $\Rbar=[-\infty,+\infty]$ induced by the value of $X$.

\begin{definition}[Law-invariant risk measure]
\label{def:lawinv}
Let $\rmdp$ be an RPOMDP and $X$ be a reward objective. A {\em law-invariant risk measure} is a function $\Psi\colon\Laws\to\Rbar$ which to every law over $\Rbar$ assigns a value in $\Rbar$. The {\em value} of $\rmdp$ with respect to $X$ and $\Psi$ is defined via
\[
\Val_\rmdp(\Psi,X)=\sup_{\sigma\in\Sigma_\rmdp}\ \inf_{\pi\in\Pi_\rmdp}\ \Psi(\law^{\sigma,\pi}_X).
\]
For a POSG $\sg$ and a reward objective $X$, the value $\Val_\sg(\Psi,X)$ of $\sg$ with respect to $X$ and $\Psi$ is defined analogously, with $\pmax$ maximizing and $\pmin$ minimizing.
\end{definition}

Intuitively, law invariance means that the value of the reward objective under a risk measure only depends on the probability distribution of the reward objective value that a random play induces. Thus, any two strategy profiles that induce the same probability distribution of reward objective values will have the same risk measure. This is thus the standard notion of law invariance of risk measures~\cite{Kusuoka2001,FollmerSchied2016}, applied to the reward objective values.

\paragraph{Examples of law-invariant risk measures.} Many risk measures commonly
studied in Markov models depend only on the probability distribution of the reward objective value, and so are law invariant:
\begin{itemize}
    \item {\em Expected value.} $\E\colon\Laws\to\Rbar$ is defined via $\E(\mu)=\textstyle\int x\,\mathrm d\mu(x)$ for all $\mu \in \Laws$.
    \item {\em Mean-variance.} Given $\lambda \geq 0$, $\MV_\lambda\colon\Laws\to\Rbar$ is defined via $\MV_\lambda(\mu)=\E(\mu)-\lambda\Var(\mu)$, where
    $\Var(\mu)=\int(x-\E(\mu))^2\,\mathrm d\mu(x)$ is the variance, for all $\mu \in \Laws$.
    \item {\em Conditional value-at-risk.} Given $\alpha\in(0,1)$, $\CVaR_\alpha\colon\Laws\to\Rbar$ is defined via $\CVaR_\alpha(\mu)=\textstyle\sup_{\eta\in\R}\bigl\{\eta-\frac1\alpha\int(\eta-x)^+\,\mathrm d\mu(x)\bigr\}$,
    where $y^+=\max\{y,0\}$, for all $\mu \in \Laws$.
    \item {\em Entropic risk.} Given $\theta\ne0$, $\ERisk_\theta\colon\Laws\to\Rbar$ is defined via $\ERisk_\theta(\mu)=\textstyle\frac1\theta\log\int e^{\theta x}\,\mathrm d\mu(x) \}$, for all $\mu \in \Laws$.
\end{itemize}
All four risk measures are defined in terms of integrals of different functions with respect to the probability distribution $\mu$. Hence, they are all of type $\Psi\colon\Laws\to\Rbar$, and thus are {\em law invariant}. While most works on solving R(PO)MDPs focus on the expected value $\E$, we consider $\E$ as well as any other law-invariant risk measure.

\begin{remark}
For laws $\mu$ that may have unbounded support, such as law of the total reward objective $\Tot$, we use the same formulas
for $\E$, $\CVaR_\alpha$, and $\ERisk_\theta$ with $\theta<0$, with the
conventions $e^{-\infty}=0$ and $\log0=-\infty$ (so that values may be
$+\infty$), and we use $\MV_\lambda$ only for returns with finite second
moments.
\end{remark}

\paragraph{Threshold problems.} We now define the threshold problems for RPOMDPs and POSGs with respect to reward objectives and law-invariant risk measures that we consider in this work.

\begin{tcolorbox}[colback=maxcolor!4,
colframe=maxcolor, boxrule=0.6pt, arc=1pt,
  left=4pt, right=4pt, top=3pt, bottom=3pt, breakable,
  title={Threshold problems}, fonttitle=\bfseries\small, coltitle=white, colbacktitle=maxcolor]
\label{def:threshold}
Given a model $\rmdp$ (an RPOMDP or a POSG), a reward objective $X$, a law-invariant risk measure $\Psi$ (with rational parameters), and a rational threshold $c$, we consider the following:
\begin{itemize}
\item \textbf{Threshold problem:} Decide if $\Val_{\rmdp}(\Psi,X)\ge c$.
\item \textbf{Strict threshold problem:} Decide if $\Val_{\rmdp}(\Psi,X)>c$.
\end{itemize}
\end{tcolorbox}

\section{Equivalence of RPOMDPs and POSGs}
\label{sec:equiv}

We now show that RPOMDPs and ac-POSGs are equivalent under the four reward objectives defined above, and any law-invariant risk measure. We prove these by constructing reductions in two directions, one from RPOMDPs to ac-POSGs (the \emph{forward} construction) and one from ac-POSGs to RPOMDPs (the \emph{backward} construction). The key property of both reductions is \emph{law preservation}: each strategy in one model is mapped to a strategy in the other model that induces the same law of the reward objective value, up to rescaling by a constant factor which depends on the reward objective. All proofs are deferred to the Appendix, see the Overview of the Appendix for the pointer to each proof.

\subsection{From RPOMDPs to ac-POSGs}

The key idea behind the forward reduction is as follows. Each step of an RPOMDP is split into two steps of the POSG: first the $\pmax$ player chooses the agent's action, then the $\pmin$ player chooses the environment's distribution. Since $\Unc(s,a)$ is a polytope, choosing a point in $\Unc(s,a)$ amounts to randomizing over its vertices, so the actions of the $\pmin$ player are the vertices of $\Unc(s,a)$. The following definition formalizes this construction.

\begin{definition}[Forward construction]
\label{def:fwd}
Let
\[
\rmdp=(S,A,\Unc,s_0,\Oa,\Oe,\Za,\Ze,r)
\]
be an RPOMDP with $\Unc(s,a)=\conv(V_{s,a})$ for all $s\in S$ and $a\in A$. We define an ac-POSG $\fwd{\rmdp}$ as a tuple
\[
(\Smax,\Smin,\Amax,\Amin,\delta,s_0,\Omax,\Omin,\Zmax,\Zmin,r')
\]
with the following components:
\begin{itemize}
\item \emph{States and actions.} The $\pmax$-states and actions are the states and actions of $\rmdp$, i.e., $\Smax=S$ and $\Amax=A$. The $\pmin$-states are the state-action pairs $(s,a)$ with $s\in S$ and $a\in A$, and $\pmin$-actions are vertices of uncertainty polytopes $V_{s,a}$; thus $\Amin=\bigcup_{s,a}V_{s,a}$. The initial state is $s_0$.
\item \emph{Transitions.} For each $s \in \Smax$ and $a \in \Amax$, we have $\delta(s,a)=\dirac{(s,a)}$. For each $(s,a) \in \Smin$ and $v \in V_{s,a}$, we have $\delta((s,a),v)=v$
\item \emph{Observations.} $\Omax=\Oa$ and $\Omin=\Oe$, with $\Zmax(s)=\Zmax((s,a))=\Za(s)$ and $\Zmin(s)=\Zmin((s,a))=\Ze(s)$.
\item \emph{Rewards.} $r'(s,a,(s,a))=0$ and $r'((s,a),v,s')=r(s,a,s')$.
\end{itemize}
\end{definition}

\begin{figure}[t]
\centering
\fbox{\begin{subfigure}{0.38\textwidth}
\centering
\begin{tikzpicture}[
    >={Stealth[scale=1.3]},
    state/.style={circle, draw=maxcolor, thick, minimum size=0.75cm},
    stochNode/.style={circle, fill=mincolor, inner sep=0pt, minimum size=1.5mm},
    edge/.style={thick, ->, >=Stealth, draw=black!80},
    scale = 0.5
]
    \node[state] (s1) at (0,0) {$s_1$};
    \node[state] (s2) at (5,3) {$s_2$};
    \node[state] (s3) at (5,-3) {$s_3$};
    \node[stochNode] (j1) at (2,0)  {};
    \node[stochNode] (j2) at (5,1)  {};
    \node[stochNode] (j3) at (5,-1) {};
    \draw[edge] (-1.5,0) -- (s1) ;
    \draw[edge] (s1) -- (j1)  node[pos=0.5,above] {\tiny $a$};
    \draw[edge] (j1) to[bend left = 50] node[pos=0.5, below] {\tiny $p_1$} (s1);
    \draw[edge] (j1) to[in=190,out=70] node[pos=0.4, left] {\tiny $q_1$} (s2);
    \draw[edge] (j1) to[in=170,out=-70] node[pos=0.4, left] {\tiny $r_1$} (s3);
    \draw[edge] (s2) -- (j2)  node[pos=0.4, right,inner sep = 0.5pt] {\tiny $a$};
    \draw[edge] (j2) to[bend right=-50] node[pos=0.3,left] {\tiny $q_2$} (s2);
    \draw[edge] (j2) to[bend right=50] node[pos=0.3,left] {\tiny $r_2$} (s3);
    \draw[edge] (s3) -- (j3)  node[pos=0.5,left,inner sep = 0.5pt] {\tiny $a$};
    \draw[edge] (j3) to[bend right=-50] node[pos=0.5, right] {\tiny $r_3$} (s3);
    \draw[edge] (j3) to[bend right=50] node[pos=0.5, right] {\tiny $q_3$} (s2);
    \draw[edge] (s1) edge[loop above] node {\tiny $b,1$} (s1);
\end{tikzpicture}
\end{subfigure}}
\hfill
\fbox{\begin{subfigure}{0.55\textwidth}
\centering
\begin{tikzpicture}[
    ={Stealth[scale=0.9]},
    maxstate/.style={circle, draw=maxcolor, thick, minimum size=0.75cm},
    minstate/.style={rectangle, draw=mincolor, thick, minimum size=0.2cm},
    dn/.style={circle, fill=black, inner sep=1pt, minimum size=0.75mm},
    edge/.style={->, thick},
    scale = 0.5
]
\node[maxstate] (s1) at (0,0) {\tiny $s_1$};
\node[maxstate] (s2) at (8,2) {\tiny $s_2$};
\node[maxstate] (s3) at (8,-2) {\tiny $s_3$};
\node[minstate] (s1a) at (3,0) {\tiny $(s_1,a)$};
\node[minstate] (s2a) at (8,4) {\tiny $(s_2,a)$};
\node[minstate] (s3a) at (8,-4) {\tiny $(s_3,a)$};
\node[dn] (s1av1) at (5,1) {};
\node[dn] (s1av2) at (5,0) {};
\node[dn] (s1av3) at (5,-1) {};
\node[dn] (s2av1) at (10,4.5) {};
\node[dn] (s2av2) at (10,3.5) {};
\node[dn] (s3av1) at (10,-4) {};
\node[minstate] (s1b) at (0,3) {\tiny $(s_1,b)$};
\draw[->,thick] (-2,0) -- (s1);
\draw[->,thick] (s1) -- node[pos=0.5, above] {\tiny $a$} (s1a);
\draw[->,thick] (s1) to[bend right=30] node[pos=0.5, right] {\tiny $b$}  (s1b);
\draw[->,thick] (s2) -- node[pos = 0.5, left] {\tiny $a$} (s2a);
\draw[->,thick] (s3) -- node[pos=0.5,left] {\tiny $a$} (s3a);
\draw[->,thick] (s1a.east) -- node[pos=1,above] {\tiny$v_{11}$} (s1av1);
\draw[->,thick] (s1a.east) -- node[pos=1,right] {\tiny$v_{12}$} (s1av2);
\draw[->,thick] (s1a.east) -- node[pos=1,below] {\tiny$v_{13}$} (s1av3);
\draw[->,thick] (s2a.east) -- node[pos=1,right] {\tiny$v_{21}$} (s2av1);
\draw[->,thick] (s2a.east) -- node[pos=1,right] {\tiny$v_{22}$} (s2av2);
\draw[->,thick] (s3a.east) -- node[pos=1,right] {\tiny$v_{31}$} (s3av1);
\draw[->,dashed] (s1av3) to[bend left=10] (s1.south);
\draw[->,dashed] (s1av3) to[bend right=10] (s2.south);
\draw[->,dashed] (s1av3) -- (s3.north);
\draw[->,dashed] (s2av2) to[bend left=30] (s3.east);
\draw[->,dashed] (s2av2) to[bend left=10] (s2.east);
\draw[->,dashed] (s3av1) to[bend right=10] (s3.east);
\draw[->,dashed] (s3av1) to[bend right=30] (s2.east);
\draw[->,thick] (s1b) to[bend right=30] node[pos=0.5, left] {\tiny $(1,0,0)$} (s1);
\end{tikzpicture}
\end{subfigure}}
\caption{The forward construction. (a) An RPOMDP with states $s_1,s_2,s_3$; action $b$ is available only in $s_1$. The drawn distributions are generic points of the uncertainty polytopes, e.g., $(p_1,q_1,r_1)\in\Unc(s_1,a)=\conv\{v_{11},v_{12},v_{13}\}$, $(0,q_2,r_2)\in\Unc(s_2,a)=\conv\{v_{21},v_{22}\}$, and $\Unc(s_3,a)=\{v_{31}\}$ with $v_{31}=(0,q_3,r_3)$. (b) The ac-POSG $\fwd{\rmdp}$. Circles are $\pmax$-states and rectangles are $\pmin$-states $(s,a)$, whose actions are the vertices of $\Unc(s,a)$. By randomizing over $v_{11},v_{12},v_{13}$ with weights $(\alpha_1,\alpha_2,\alpha_3)$, $\pmin$ realizes the point $\alpha_1v_{11}+\alpha_2v_{12}+\alpha_3v_{13}$ of $\Unc(s_1,a)$. For readability, dashed outcome edges are drawn for one vertex per $\pmin$-state only.}
\label{fig:forward}
\Description{Two boxed panels. Left: an RPOMDP with three states s1, s2, s3; action a from each state leads through an uncertain distribution to several states, and action b from s1 is a self-loop. Right: the constructed game, in which every max-state s is followed by min-states (s,a) whose outgoing actions are the vertices of the uncertainty polytope of (s,a); each vertex leads to a distribution over the original states.}
\end{figure}
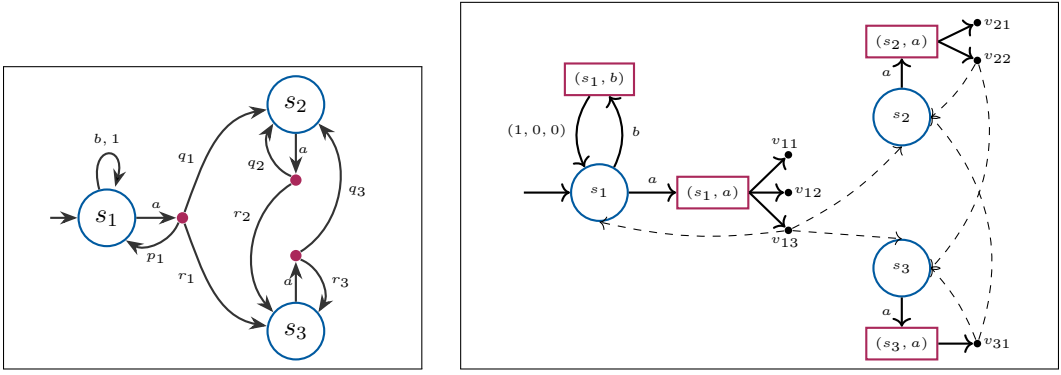

In both the $\pmax$-state $s$ and the $\pmin$-state $(s,a)$, both players receive the same observations as the agent and the environment receive in state $s$ in the RPOMDP $\rmdp$, respectively.
Strategies of the two models correspond as follows:
\begin{itemize}
\item An agent strategy $\sigma$ in the RPOMDP becomes a $\pmax$-strategy that plays $\sigma(o_0o_1\cdots o_t)$ on observation history $o_0o_0o_1o_1\cdots$ $o_{t-1}o_{t-1}o_t$ (in $\fwd{\rmdp}$, the $\pmax$-observations in $\pmax$- and $\pmin$-states are the same so each observation appears twice).
\item An environment strategy $\pi$ becomes a $\pmin$-strategy as follows. In RPOMDP $\rmdp$, after the environment history $h=o_0\cdots o_t$, the environment selects for every pair $(s,a)$ a point $p=\pi(h)(s,a)$ of $\Unc(s,a)=\conv(V_{s,a})$. We write this point as a convex combination $p=\sum_{v\in V_{s,a}}\lambda_v\,v$ of the vertices. Then, in the POSG $\fwd{\rmdp}$, we let $\pmin$, after the $\pmin$-history $o_0o_0o_1o_1\cdots o_to_t$, play in the $\pmin$-state $(s,a)$ the mixed action that chooses each vertex $v$ with probability~$\lambda_v$. The next state is sampled from $\sum_v\lambda_v\,v=p$, exactly as in~$\rmdp$.
\end{itemize}
The following lemma shows that this strategy mapping is law preserving, up to rescaling by a constant factor. Recall the definition of $\push{c}$ from Section~\ref{sec:prelim}.

\begin{lemma}[Forward law preservation]
\label{lem:fwd}
Let $\sg=\fwd{\rmdp}$, and associate with each reward objective $X$ of $\rmdp$ a reward objective $X^\sg$ of $\sg$ and a constant $c>0$ as follows:
\[
\begin{array}{lcccc}
X: & \Tot_N & \Tot & \LimAvg & \Disc_{\beta^2}\\
X^{\sg}: & \Tot_{2N} & \Tot & \LimAvg & \Disc_{\beta}\\
c: & 1 & 1 & 1/2 & \beta
\end{array}
\]
where $\beta\in(0,1)$ is assumed to be rational. Then, there is a bijection $\sigma\mapsto\sigma^\sg$ from agent strategies of $\rmdp$ to $\pmax$-strategies of $\sg$ such that, for every agent strategy $\sigma$,
\[
\bigl\{\law^{\sigma^\sg,\pi'}_{X^\sg}:\pi'\in\Pi_\sg\bigr\}=\bigl\{\push{c}\law^{\sigma,\pi}_{X}:\pi\in\Pi_\rmdp\bigr\}.
\]
More precisely, each $\pmin$-strategy $\pi'$ induces an environment strategy $\pi$ with $\law^{\sigma^\sg,\pi'}_{X^\sg}=\push{c}\law^{\sigma,\pi}_{X}$ for all $\sigma$.
\end{lemma}

The idea of the proof is that erasing the $\pmin$-states and vertex choices from a play of $\sg$ yields a play of $\rmdp$, and that under corresponding strategies this projection maps $\Prob^{\sigma^\sg,\pi'}_{\sg}$ to $\Prob^{\sigma,\pi}_{\rmdp}$; the reward sequence of the game play is that of the projected play, with a zero inserted before every reward. The constants $c$ then follow from a direct computation, e.g., $\sum_i\beta^{2i+1}r_i=\beta\sum_i(\beta^2)^ir_i$ for the discounted return. Thus the law of the return is preserved up to the factor $c$. As $c$ depends only on the return (for $\Disc_{\beta^2}$, on $\beta$), and neither on the strategies nor on the law functional, the robust values agree up to the same rescaling (Theorem~\ref{thm:umbrella}).

\subsection{From ac-POSGs to RPOMDPs}

In the other direction, the environment in the RPOMDP takes the role of the $\pmin$-player in the ac-POSG. Two points need care. First, in the ac-POSG, $\pmin$ moves in a state $t$ of its own, after observing $\Zmin(t)$. In contrast, in the RPOMDP, the environment moves in the same step as the agent, with no intermediate state, and hence no observation, in between. To address this discrepancy, in our reduction we turn every $\pmin$-state $t$ into a state $d_t$ of the RPOMDP in which the agent has a single dummy action~$\anew$ and the environment, after observing $\Zmin(t)$, chooses the next distribution from the polytope spanned by $\pmin$'s options. Second, the reward of a $\pmin$-transition may depend on $\pmin$'s action, whereas the reward of an RPOMDP transition depends only on the agent state, the agent's action, and the transition's target state. We therefore extend every $\pmax$-state $s$ to a pair $(s,b)$ whose \emph{tag} $b$ records the $\pmin$ action that led to $s$, so that the transition from $d_t$ to $(s',b)$ can earn the reward $r(t,b,s')$. Neither player observes the tags.

\begin{definition}[Backward construction]
\label{def:bwd}
Let
\[
\sg=(\Smax,\Smin,\Amax,\Amin,\delta,s_0,\Omax,\Omin,\Zmax,\Zmin,r)
\]
be an ac-POSG, and let $\anew\notin\Amax\cup\Amin$ be a fresh symbol. The RPOMDP $\bwd{\sg}$ is the tuple
\[
(S',A',\Unc,(s_0,\anew),\Omax\times\{0,1\},\Omin,\Za,\Ze,r')
\]
with the following components.
\begin{itemize}
\item \emph{States.} $S'$ consists of initial state $(s_0,\anew)$, a state $d_t$ for every $t\in\Smin$ and of the \emph{tagged states} $(s,b)$ for $s\in\Smax$ and $b\in\Amin$, where the tag $b$ is the $\pmin$ action that led to $s$.
\item \emph{Actions.} We set $A'=\Amax\cup\{\anew\}$. For each tagged state $(s,b)$, the set of effective actions is $A$, and for each state $d_t$, the only effective action is $\anew$.
\item \emph{Uncertainty.} In tagged states, the environment has no choice:
\[
\Unc((s_0,\top),a)=\{\hat\delta(s_0,a)\},\qquad\hat\delta(s_0,a)(d_t)=\delta(s_0,a)(t)
\]
\[
\Unc((s,b),a)=\{\hat\delta(s,a)\},\qquad\hat\delta(s,a)(d_t)=\delta(s,a)(t).
\]
In the states $d_t$, it chooses from the polytope
\[
\Unc(d_t,\anew)=\conv\{\delta'_{t,b}:b\in A(t)\},
\]
where $\delta'_{t,b}((s',b'))=\delta(t,b)(s')$ if $b'=b$ and $0$ otherwise.
\item \emph{Observations.} For the agent, $\Za((s,b))=(\Zmax(s),0)$ and $\Za(d_t)=(\Zmax(t),1)$; for the environment, $\Ze((s,b))=\Zmin(s)$ and $\Ze(d_t)=\Zmin(t)$.
\item \emph{Rewards.} $r'((s,b),a,d_t)=r(s,a,t)$ and $r'(d_t,\anew,(s',b))=r(t,b,s')$; all other rewards are~$0$.
\end{itemize}
\end{definition}

Since the distributions $\delta'_{t,b}$ have pairwise disjoint supports, a point of $\Unc(d_t,\anew)$ determines the mixed action of $\pmin$ that it represents. The additional bit in the agent's observations tells the agent whose turn it is, so that it can distinguish the tagged $\pmax$-states $(s,b)$, where it chooses among the effective actions $A(s)$, from the states $d_t$, where it has only the effective action $\anew$. The bit gives the agent no new information about $\sg$, because turns alternate: it is $0$ after an odd number of observations and $1$ after an even number.

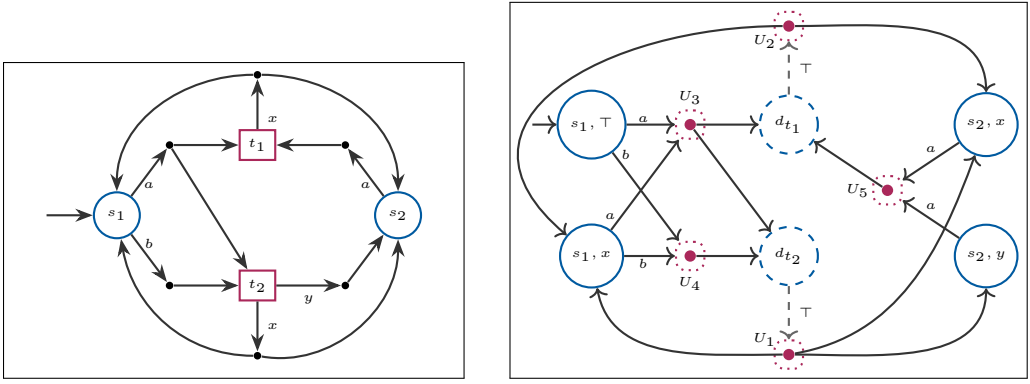
\begin{figure}[t]
\fbox{\begin{subfigure}{0.42\textwidth}
\centering
\begin{tikzpicture}[
    maxNode/.style={circle, draw=maxcolor, thick, minimum size=0.25cm},
    minNode/.style={rectangle, draw=mincolor, thick, minimum size=0.25cm},
    stochNode/.style={circle, fill=black, inner sep=0pt, minimum size=1mm},
    edge/.style={thick, ->, >=Stealth, draw=black!80},
    scale=0.62
]
    \node[maxNode] (s1) at (0, 0)   {\tiny $s_1$};
    \node[maxNode] (s2) at (6, 0)  {\tiny $s_2$};
    \node[minNode] (t1) at (3, 1.5) {\tiny $t_1$};
    \node[minNode] (t2) at (3,-1.5) {\tiny $t_2$};
    \node[stochNode] (ja)  at (1.125,  1.5) {};
    \node[stochNode] (jb)  at (1.125, -1.5) {};
    \node[stochNode] (2ja) at (4.875,  1.5)    {};
    \node[stochNode] (2jy) at (4.875, -1.5) {};
    \node[stochNode] (2jx) at (3, -3) {};
    \node[stochNode] (1jx) at (3, 3) {};
    \draw[edge] (-1.5,0) -- (s1);
    \draw[edge] (s1) -- (ja)  node[pos=0.5,below]  {\tiny $a$};
    \draw[edge] (s1) -- (jb)  node[pos=0.5,above]  {\tiny $b$};
    \draw[edge] (s2) -- (2ja) node[pos=0.5,below]       {\tiny $a$};
    \draw[edge] (t2) -- (2jx) node[pos=0.5,right] {\tiny $x$};
    \draw[edge] (t2) -- (2jy) node[pos=0.5,below]       {\tiny $y$};
    \draw[edge] (t1) -- (1jx) node[pos=0.2,right]       {\tiny $x$};
    \draw[edge] (ja) -- (t1); \draw[edge] (ja) -- (t2);
    \draw[edge] (jb) -- (t2);
    \draw[edge] (2ja) -- (t1);
    \draw[edge] (2jx) to[in = 280, out = 180] (s1);
    \draw[edge] (2jx) to[in = 270, out = 350] (s2);
    \draw[edge] (2jy) -- (s2);
    \draw[edge] (1jx) to[in = 90, out = 0] (s2);
    \draw[edge] (1jx) to[in = 90, out = 180] (s1);
\end{tikzpicture}
\end{subfigure}}
\hfill
\begin{subfigure}{0.55\textwidth}
\centering
\fbox{
\begin{tikzpicture}[
    trim left=-25pt,
    maxNode/.style={circle, draw=maxcolor, thick, minimum size=0.25cm},
    botNode/.style={circle, draw=maxcolor, dashed, thick, minimum size=0.25cm},
    uncSet/.style={rounded corners=3.5pt, dotted, draw=mincolor, thick, inner sep=3pt},
    stochNode/.style={circle, fill=mincolor, inner sep=0pt, minimum size=1.5mm},
    edge/.style={->, thick, draw=black!80},
    botEdge/.style={->, thick, dashed, draw=black!60},
    scale = 0.87
]
    \pgfdeclarelayer{background}
    \pgfsetlayers{background,main}
    \node[maxNode] (s1n)  at (0, 1)    {\tiny $s_1,\anew$};
    \node[maxNode] (s1x)  at (0, -1)    {\tiny $s_1,x$};
    \node[maxNode] (s2x)  at (6, 1)  {\tiny $s_2,x$};
    \node[maxNode] (s2y)  at (6, -1)  {\tiny $s_2,y$};
    \node[botNode] (dt1) at (3, 1)  {\tiny $d_{t_1}$};
    \node[botNode] (dt2) at (3,-1)  {\tiny $d_{t_2}$};
    \node[stochNode] (ja)  at (1.5, 1) {};
    \node[stochNode] (jb)  at (1.5,-1) {};
    \node[stochNode] (2ja) at (4.5, 0)    {};
    \node[stochNode] (1jx) at (3,2.5)  {};
    \node[stochNode] (2jx) at (3,-2.5) {};
    \begin{pgfonlayer}{background}
      \node[uncSet,fit=(ja)]   (uja)  {};
      \node[uncSet,fit=(jb)]   (ujb)  {};
      \node[uncSet,fit=(2ja)]  (u2ja) {};
      \node[uncSet,fit=(1jx)]  (u1jx) {};
      \node[uncSet,fit=(2jx)] (u2jxy) {};
      \node[left=1.2pt of u2jxy.north,align=center]{\tiny $U_1$};
      \node[left=1.2pt of u1jx.south, align=center]{\tiny $U_2$};
      \node[above=2pt of ja.north, align=center]{\tiny $U_3$};
      \node[below=2pt of jb.south, align=center]{\tiny $U_4$};
      \node[left=1pt of 2ja.west, align=center]{\tiny $U_5$};
    \end{pgfonlayer}
    \draw[edge] (-0.9,1) -- (s1n);
    \draw[edge] (s1x) -- (uja)  node[pos=0.1,above left, inner sep=0.5pt]  {\tiny $a$};
    \draw[edge] (s1n) -- (uja)  node[pos=0.5,above left, inner sep=0.5pt]  {\tiny $a$};
    \draw[edge] (s1n) -- (ujb)  node[pos=0.1,above right, inner sep=0.5pt]  {\tiny $b$};
    \draw[edge] (s1x) -- (ujb)  node[pos=0.5,below left, inner sep=0.5pt]  {\tiny $b$};
    \draw[edge] (s2x) -- (u2ja) node[midway,above]       {\tiny $a$};
    \draw[edge] (s2y) -- (u2ja) node[midway,above]       {\tiny $a$};
    \draw[botEdge] (dt1) -- (u1jx)  node[midway,right] {\tiny $\anew$};
    \draw[botEdge] (dt2) -- (u2jxy) node[midway,right]  {\tiny $\anew$};
    \draw[edge] (ja)  -- (dt1); \draw[edge] (ja)  -- (dt2);
    \draw[edge] (jb)  -- (dt2);
    \draw[edge] (2ja) -- (dt1);
    \draw[edge] (1jx)  to[out = 0,in=90] (s2x);
    \draw[edge] (1jx) to[in = 140, out = 180,looseness=1.5] (s1x);
    \draw[edge] (2jx) to[out=180,in=280] (s1x);
    \draw[edge] (2jx) to[out=10,in=250] (s2x);
    \draw[edge] (2jx) to[out=0,in=270] (s2y);
\end{tikzpicture}}
\end{subfigure}
\caption{The backward construction. (a) An ac-POSG $\sg$ with $\pmax$-states $s_1,s_2$ (circles) and $\pmin$-states $t_1,t_2$ (rectangles); $\pmin$ can play $x$ in $t_1$ and $x$ or $y$ in $t_2$. (b) The RPOMDP $\bwd{\sg}$; dotted boxes are uncertainty sets. The sets $U_3=\Unc((s_1,\tau),a)$, $U_4=\Unc((s_1,\tau),b)$, and $U_5=\Unc((s_2,\tau),a)$ are the singletons $\{\hat\delta(s_1,a)\}$, $\{\hat\delta(s_1,b)\}$, and $\{\hat\delta(s_2,a)\}$ for every tag~$\tau$, and $U_2=\Unc(d_{t_1},\anew)=\{\delta'_{t_1,x}\}$ is a singleton as well. The set $U_1=\Unc(d_{t_2},\anew)=\conv\{\delta'_{t_2,x},\delta'_{t_2,y}\}$ is a segment: choosing a point on it is choosing a mixed action of $\pmin$ in $t_2$, and the tag of the next state records which action was realized.}
\label{fig:backward}
\Description{Two boxed panels. Left: a game with max-states s1, s2 and min-states t1, t2, where t1 has one action x and t2 has actions x and y leading to max-states. Right: the constructed RPOMDP with tagged states such as (s1, top), (s1, x), (s2, x), (s2, y), dummy states d_t1 and d_t2 with a single dummy action, and uncertainty sets drawn as dotted boxes; the uncertainty set at d_t2 is the segment between the two distributions corresponding to the min actions x and y.}
\end{figure}

\begin{lemma}[Backward law preservation]
\label{lem:bwd}
Let $\rmdp=\bwd{\sg}$. There is a bijection between the plays of $\sg$ and those of $\rmdp$ under which corresponding plays have identical reward sequences, and a bijection $\sigma\mapsto\sigma^{\rmdp}$ from $\pmax$-strategies to agent strategies such that, for every $\pmax$-strategy $\sigma$ and every reward objective $X$ (with the same horizon or discount factor on both sides),
\[
\bigl\{\law^{\sigma^\rmdp,\pi'}_{X}:\pi'\in\Pi_\rmdp\bigr\}=\bigl\{\law^{\sigma,\pi}_{X}:\pi\in\Pi_\sg\bigr\}.
\]
\end{lemma}

Since both constructions preserve achievable return laws, they preserve the value of every law functional which is our main result.
\subsection{The umbrella theorem}

    \begin{theorem}[Umbrella theorem]
    \label{thm:umbrella}
    Let $\Psi\colon\Laws\to\Rbar$ be a law-invariant risk measure.
    \begin{enumerate}
    \item[(a)] For every RPOMDP $\rmdp$ and reward objective $X$, with $X^\sg$ and $c$ as in Lemma~\ref{lem:fwd},
    \[
    \Val_{\fwd{\rmdp}}(\Psi,X^\sg)=\Val_\rmdp(\Psi\circ\push{c},X).
    \]
    \item[(b)] For every ac-POSG $\sg$ and reward objective $X$,
    \[
    \Val_{\bwd{\sg}}(\Psi,X)=\Val_\sg(\Psi,X).
    \]
    \item[(c)] Both constructions are computable in logarithmic space.
    \end{enumerate}
Consequently, for every law-invariant risk measure and each of the four reward objectives, the threshold problems (strict and non-strict) for ac-POSGs reduce in logarithmic space to those for RPOMDPs with the same risk measure and reward objective. Conversely, the threshold problems for $\Psi$ and $X$ on RPOMDPs reduce in logarithmic space to those for $\Psi\circ\push{1/c}$ and $X^\sg$ on ac-POSGs, with the same threshold. For instance, $c=1$ for finite-horizon total return,  hence the risk measure is unchanged.
Likewise, for discounted-sum rewards, the direction applies to RPOMDPs whose discount factor is the square $\beta^2$ of a rational $\beta$, and the resulting ac-POSG has discount factor $\beta$ with $c=\beta$.
    \end{theorem}

    \begin{proof}
    For a model and a strategy $\sigma$ of the maximizing player, let $\lawset(\sigma)$ be the set of laws of the reward objective that the opponent can induce against $\sigma$; then $\Val(\Psi,X)=\sup_\sigma\inf_{\mu\in\lawset(\sigma)}\Psi(\mu)$. By Lemma~\ref{lem:fwd}, $\sigma\mapsto\sigma^\sg$ is a bijection with $\lawset_{\sg}(\sigma^\sg)=\push{c}\lawset_{\rmdp}(\sigma)$, so
    \[
    \inf_{\mu\in\lawset_\sg(\sigma^\sg)}\Psi(\mu)=\inf_{\mu\in\lawset_\rmdp(\sigma)}\Psi(\push{c}\mu),
    \]
    and taking the supremum over $\sigma$ yields~(a). Part~(b) follows in the same way from Lemma~\ref{lem:bwd}. Part~(c) is shown in the appendix. For the threshold problems, (b) gives $\Val_\sg(\Psi,X)=\Val_{\bwd{\sg}}(\Psi,X)$. Moreover, (a), applied to the law functional $\Psi\circ\push{1/c}$, gives
    \[
    \Val_\rmdp(\Psi,X)=\Val_{\fwd{\rmdp}}(\Psi\circ\push{1/c},X^\sg),
    \]
    because $\push{1/c}\push{c}\mu=\mu$ for every law $\mu$. In both cases, the threshold is unchanged. A detailed proof  of all three parts can be found in the appendix.
    \end{proof}

\begin{corollary}
\label{cor:standard}
In the setting of Theorem~\ref{thm:umbrella}(a), let $\sg=\fwd{\rmdp}$. For all $\lambda\ge0$, $\alpha\in(0,1)$, and $\theta\ne0$,
\begin{align*}
\Val_{\sg}(\E,X^\sg)&=c\,\Val_\rmdp(\E,X),\\
\Val_{\sg}(\MV_\lambda,X^\sg)&=c\,\Val_\rmdp(\MV_{c\lambda},X),\\
\Val_{\sg}(\CVaR_\alpha,X^\sg)&=c\,\Val_\rmdp(\CVaR_\alpha,X),\\
\Val_{\sg}(\ERisk_\theta,X^\sg)&=c\,\Val_\rmdp(\ERisk_{c\theta},X).
\end{align*}
For the backward construction, all values coincide with unchanged parameters.
\end{corollary}

\begin{table}[t]
\caption{\emph{Partial observation}: we obtain lower bounds  for POMDPs which apply also to RPOMDPs and ac-POSGs; upper bounds (decidability and approximation) hold for RPOMDPs and ac-POSGs. ``undec.''\ means undecidable for both threshold
variants, ``undec.$^*$'' for the strict variant only, ``approx.''\ that values can be
approximated to any precision. \emph{Full observation}: polytopic $(s,a)$-rectangular
RMDPs; the RMDP lower bounds also hold for TBSGs. $^\ddagger$Under polynomial-time Turing reductions. New results are in
\ours{green}.}
\label{tab:results}
\centering
\footnotesize
\makebox[\textwidth][c]{%
\begin{tabular}{@{}c@{\hspace{8pt}}l cccc@{}}
\toprule
\multirow{6}{*}{\rotatebox{90}{Partial obs.}}
& Objective
& \shortstack{Finite horizon}
& \shortstack{Total $\Tot$}
& \shortstack{Average $\LimAvg$}
& \shortstack{Discounted $\Disc_\gamma$} \\
&& \shortstack{(Thm.~\ref{thm:finite}, Prop.~\ref{prop:approx})}
& \shortstack{(Thm.~\ref{thm:undec})}
& \shortstack{(Thm.~\ref{thm:undec})}
& \shortstack{(Thm.~\ref{thm:disc-cvar}, Prop.~\ref{prop:approx})} \\
\cmidrule{2-6}
& $\E$ & $\PSPACE$-complete~\cite{PapadimitriouTsitsiklis1987}, \ours{decidable} & undec.$^*$~\cite{Madani2003} & undec.$^*$~\cite{Madani2003} & undec.$^*$~\cite{Madani2003}; \ours{approx.}\\
& $\MV_\lambda$ & \ours{$\PSPACE$-hard, decidable} & \ours{undec.} & \ours{undec.} & open for $\lambda>0$; \ours{approx.}\\
& $\CVaR_\alpha$ & \ours{$\PSPACE$-hard, decidable} & \ours{undec.} & \ours{undec.} & \ours{undec.$^*$; approx.}\\
& $\ERisk_\theta$ & \ours{$\PSPACE$-hard, decidable} & \ours{undec.} & \ours{undec.} & open; \ours{approx.}\\
\midrule
\multirow{6}{*}{\rotatebox{90}{Full obs.}}
& Objective (return) & MDP & \multicolumn{3}{c}{RMDP}\\
\cmidrule{2-6}
& $\E$ (discounted) & $\Pclass$~\cite{Puterman1994} & \multicolumn{3}{c}{\ours{SSG-hard$^\ddagger$ (Thm.~\ref{cor:expected})}, in $\NP\cap\coNP$~\cite{Chatterjee2024}}\\
& $\E$ (total, average) & $\Pclass$~\cite{Puterman1994} & \multicolumn{3}{c}{\ours{SSG-hard (Thm.~\ref{thm:ssg})}; average: in $\NP\cap\coNP$~\cite{Chatterjee2024}}\\
& $\MV_\lambda$ (total, average) & total: $\EXPTIME$-hard, in $\NEXPTIME$~\cite{PiribauerSankurBaier2022} & \multicolumn{3}{c}{\ours{SSG-hard (Thm.~\ref{thm:ssg})}}\\
& $\CVaR_\alpha$ (total, average) & average, non-strict: $\Pclass$~\cite{KretinskyMeggendorfer2018} & \multicolumn{3}{c}{\ours{SSG-hard (Thm.~\ref{thm:ssg})}}\\
& $\ERisk_\theta$ (total, average) & total: $\Pclass$~\cite{BaierEntropicRisk2023} & \multicolumn{3}{c}{\ours{SSG-hard (Thm.~\ref{thm:ssg})}}\\
\bottomrule
\end{tabular}%
}
\end{table}

\section{The Complexity Landscape}
\label{sec:complexity}

By Theorem~\ref{thm:umbrella}, every complexity bound for the threshold problem for each of the reward objectives and any law-invariant risk measure transfers between RPOMDPs and ac-POSGs (for average and discounted returns, with the rescaling by the constant $c$ of Lemma~\ref{lem:fwd}).
We now use this result to derive new complexity bounds for R(PO)MDPs and ac-POSGs under reward objectives; Table~\ref{tab:results} summarizes them. All numbers in the input (probabilities, rewards, discount factors, thresholds, and parameters) are rationals encoded in binary, so that every instance has a finite description. Our lower bounds share one proof pattern, which we first isolate in terms of the law-invariant risk measure $\Psi$ in Section \ref{sec:bern}.

\subsection{Bernoulli profiles}
\label{sec:bern}
Our hardness results are reductions from decision problems about the \emph{probability of an event} in a simpler model: whether a POMDP strategy avoids a bad event with probability~$1$ over a finite horizon~\cite{PapadimitriouTsitsiklis1987} (Theorem~\ref{thm:finite}); whether some POMDP strategy reaches a target with probability above $b$ or all strategies stay below $a$, a gap problem derived from probabilistic automata~\cite{CondonLipton1989} (Theorem~\ref{thm:undec}); and whether a simple stochastic game reaches its target with probability above $\tfrac12$~\cite{Condon1992} (Theorem~\ref{thm:ssg}). In each reduction we turn the instance into a reward model whose return equals $v>0$ if the event occurs and $0$ otherwise. For the finite-horizon total reward we pay $v$ once at the last step if the event has occurred; for the total reward we pay $v$ once when the absorbing target is entered; for the long-run average we pay $v$ in every step spent in the target. This does not work for discounted returns, where the return also depends on \emph{when} the event occurs, so discounted-sum rewards are treated separately.

A strategy under which the event has probability $p$ then yields the return law
$\Ber_v(p)=(1-p)\dirac{0}+p\,\dirac{v}$ for $p\in[0,1]$.
A law-invariant risk measure $\Psi$ sees only this law, so the value of the strategy is $\Psi(\Ber_v(p))$, a function of $p$ alone. If this function is nondecreasing in $p$, the best strategy is the one that maximizes $p$. A threshold for $\Psi$ then becomes a threshold on the optimal probability, which is the question the source problem asks. We therefore only need to know how $\Psi$ behaves on the laws $\Ber_v(p)$, which motivates the following notion.

The \emph{Bernoulli profile} of a law-invariant risk measure $\Psi$ is the function $\beta^\Psi_v(p)=\Psi(\Ber_v(p))$ for $v > 0$ and $p \in [0,1]$. For the
four risk measures that we consider (computations from the definitions in
Section~\ref{sec:lawinv} are provided in Appendix~\ref{app:bernoulli}),
\[
\beta^{\E}_v(p)=pv \qquad \beta^{\CVaR_\alpha}_v(p)=v\max\bigl\{0,\,1-(1-p)/\alpha\bigr\}
\]
\[
\beta^{\MV_\lambda}_v(p)=pv-\lambda v^2p(1-p) \qquad \beta^{\ERisk_\theta}_v(p)=\tfrac1\theta\log\bigl(1-p+pe^{\theta v}\bigr)
\]

\begin{definition}[Monotone and strict at the top Bernoulli profiles]
\label{def:bernoulli}
A law-invariant risk measure $\Psi$ has a \emph{monotone Bernoulli profile} if, for some rational $v>0$, $\beta^\Psi_v$ is nondecreasing on $[0,1]$ and not constant on $(0,1)$. The profile is \emph{strict at the top} if, in addition, $\beta^\Psi_v(p)<\beta^\Psi_v(1)$ for all $p<1$ and $\beta^\Psi_v(1)$ is rational.
\end{definition}

The expectation, $\CVaR_\alpha$, and $\ERisk_\theta$ have monotone Bernoulli profiles that are strict at the top for every $v>0$. $\MV_\lambda$ has monotone Bernoulli profiles that is strict at the top for $v=1/(1+\lambda)$, since then $\lambda v<1$.(Appendix ~\ref{app:bernoulli})

\subsection{Partial observation}
\label{sec:po}

\paragraph{Lower bounds.} Lower bounds for POMDPs apply to RPOMDPs directly, since a POMDP is an RPOMDP with singleton uncertainty sets, and they apply to ac-POSGs by our equivalence in Theorem~\ref{thm:umbrella}.

\begin{theorem}[Finite horizon total reward]
\label{thm:finite}
If the Bernoulli profile of a law-invariant risk measure $\Psi$ is strict at the top, then the finite-horizon total reward threshold problem for $\Psi$ (both strict and non-strict) is $\PSPACE$-hard, already for POMDPs; hence it is $\PSPACE$-hard for RPOMDPs and ac-POSGs.
\end{theorem}

\begin{proof}[Proof sketch]
\cite{PapadimitriouTsitsiklis1987} show that it is $\PSPACE$-hard to decide whether a POMDP with nonnegative costs has a strategy of expected total cost $0$ over the first $N\le|S|$ steps, i.e., one that avoids all pairs with positive cost with probability~$1$. Their result concerns the expected cost, and it is a question about the probability $p^\sigma$ of avoiding the costly pairs. We show that, for every law-invariant risk measure $\Psi$ whose Bernoulli profile is strict at the top, the threshold problem for $\Psi$ encodes exactly this question. To this end, we add a step counter and collect reward $v$ exactly when step $N$ is completed without a costly pair, so that the total return has law $\Ber_v(p^\sigma)$, where $p^\sigma$ is the success probability of $\sigma$. By compactness, $\max_\sigma p^\sigma$ exists, so the value is $\beta^\Psi_v(\max_\sigma p^\sigma)$. Strictness at the top gives $\beta^\Psi_v(p)<\beta^\Psi_v(1)$ for $p<1$, so the value is at least the rational threshold $\beta^\Psi_v(1)$ if and only if some strategy succeeds with probability~$1$. Hence the threshold problem for $\Psi$ is $\PSPACE$-hard for POMDPs, by a reduction computable in logarithmic space. The result transfers to RPOMDPs, of which POMDPs are a special case, and to ac-POSGs by the forward construction of Section~\ref{sec:equiv}, which is computable in logarithmic space and preserves the laws of the finite-horizon total return (with the horizon doubled and no rescaling of the rewards). Thus, the hardness of \cite{PapadimitriouTsitsiklis1987} for expected cost is lifted to all risk objectives with a strict Bernoulli profile, and to both RPOMDPs and ac-POSGs.
\end{proof}

For infinite horizon total reward and long-run average reward, we use the undecidability of probabilistic automata with a gap~\cite{CondonLipton1989}. Lemma~\ref{lem:gap} in Appendix~\ref{app:thegap} translates the result to POMDPs: for rationals $0<a<b<1$, one computes from a two-counter machine a POMDP with a single observation and an absorbing target $T$ such that some strategy reaches $T$ with probability greater than $b$ if the machine halts, while every strategy, randomized or not, reaches $T$ with probability at most $a$ otherwise.

\begin{theorem}[Total and long-run average reward]
\label{thm:undec}
If a law-invariant risk measure $\Psi$ has a monotone Bernoulli profile, then there is a rational $c$ such that, for $X\in\{\Tot,\LimAvg\}$, both $\Val(\Psi,X)\ge c$ and $\Val(\Psi,X)>c$ are undecidable for POMDPs with a single observation; hence the threshold problems for $\Psi$ are undecidable for RPOMDPs and ac-POSGs.
\end{theorem}
\begin{proof}[Proof sketch]
Choose rationals $0{<}a{<}b{<}1$ with $\beta^\Psi_v(a){<}\beta^\Psi_v(b)$ and a rational $c$
strictly between these values; they depend only on $\Psi$.
Given an instance of the probabilistic automaton gap problem for rationals $a$ and $b$, in our proof we translate the automaton into a POMDP which collects reward $v$ upon entering $T$ (for $\Tot$), or in every step spent in $T$ (for $\LimAvg$). Then, the value of the reward objective in the POMDP under strategy $\sigma$ has law $\Ber_v(p^\sigma)$, where $p^\sigma$ is the probability of reaching $T$ under $\sigma$.
If the machine halts, some $\sigma$ has $p^\sigma>b$, so
the value is at least $\beta^\Psi_v(b)>c$. Otherwise $p^\sigma\le a$ for all $\sigma$,
so the value is at most $\beta^\Psi_v(a)<c$. Hence both threshold problems with the
fixed threshold $c$ reduce from the halting problem of two-counter
machines. The result transfers to RPOMDPs and by our equivalence result to ac-POSGs.
\end{proof}

For discounted-sum rewards, the Bernoulli technique fails, since the return depends on the distribution of the hitting time of $T$. For CVaR, however, we show that a nonnegative return has $\CVaR_\alpha>0$ if and only if it is positive with probability greater than $1-\alpha$. Together with Lemma~\ref{lem:gap}, this gives undecidability for discounted-sum rewards and CVaR risk measure.

\begin{theorem}[Discounted-sum reward with CVaR]
\label{thm:disc-cvar}
For all rationals $\alpha,\gamma\in(0,1)$, it is undecidable whether $\Val(\CVaR_\alpha,\Disc_\gamma)>0$, already for POMDPs with a single observation. Hence the strict threshold problem for discounted-sum rewards with CVaR risk measure is undecidable for RPOMDPs and ac-POSGs, with any rational discount factor.
\end{theorem}

\begin{proof}[Proof sketch]
Take the POMDP of Lemma~\ref{lem:gap} with $a<1-\alpha<b$, and collect reward $1$ upon entering $T$. For every discount factor the return is nonnegative, and it is positive exactly on plays reaching $T$. Hence, by Lemma~\ref{lem:cvar-positive}, a strategy has positive $\CVaR_\alpha$ if and only if it reaches $T$ with probability greater than $1-\alpha$. If the machine halts, the value is positive; otherwise every strategy reaches $T$ with probability at most $a<1-\alpha$ and the value is $0$. This transfers to RPOMDPs, and to by our equivalence result to ac-POSGs.
\end{proof}

\paragraph{Decidability and approximability.} The above results establish lower bounds and undecidability of problems. We now present results concerning decidability and approximability. We show that, for finite horizon total rewards and all four risk measures that we consider, the problem is decidable. Then, we use this results to show that, for discounted-sum rewards and all four risk measures, the value can be approximated up to an arbitrary precision.

\begin{proposition}[Decidability and approximation]
\label{prop:approx}
Let $\Psi$ be one of the four risk measures $\E$, $\MV_\lambda$, $\CVaR_\alpha$, or $\ERisk_\theta$, with rational parameters.
\begin{enumerate}
\item[(a)] The finite-horizon total reward threshold problem under any of the four risk measures $\Psi$ is decidable for RPOMDPs and ac-POSGs.
\item[(b)] For discounted-sum rewards under any of the four risk measures $\Psi$, the value can be approximated up to any precision; in particular, the strict threshold problem is recursively enumerable.
\end{enumerate}
\end{proposition}

\begin{proof}[Proof sketch]
(a) With a finite horizon, the threshold question is a first-order sentence over the reals of the form ``there exist agent parameters such that for all environment parameters a polynomial inequality holds'' (CVaR is a maximum of polynomials), which is decidable~\cite{Tarski1951}. For entropic risk, all constants $e^{\theta x}$ are integer powers of one transcendental number $e^{1/D}$~\cite{Baker1975}; we replace it by a variable and compare it with the finitely many algebraic boundary points of the resulting semialgebraic set.
(b) Truncating after $N$ steps changes the discounted return by at most $\epsilon_N=\gamma^N\max|r|/(1-\gamma)$ on every play. All four functionals are Lipschitz under such uniform perturbations, so the value changes by $O(\epsilon_N)$, and the truncated value is computable by~(a).
\end{proof}

\subsection{Full observation}
\label{sec:full}

For the special case of full observability and RMDPs, our equivalence result shows that $(s,a)$-rectangular polytopic RMDPs are equivalent to turn-based stochastic games (TBSGs), for all four reward objectives and under any law-invariant risk measure. Our novel complexity results for RMDPs concern {\em lower bounds}, which to the best of our knowledge present the first lower complexity bounds for RMDPs. In particular, we show that a number of problems are {\em SSG-hard}. The SSG problem is concerned with solving \emph{simple stochastic games} (SSGs) and deciding whether the value of a vertex exceeds $\frac12$. It is known that SSG is in $\NP\cap\coNP$~\cite{Condon1992}, but no polynomial-time algorithm is known. The proofs can be found in the appendix.

\begin{theorem}[Total and long-run average reward]
\label{thm:ssg}
Let $\Psi$ be one of the four risk measures $\E$, $\MV_\lambda$ ($\lambda\ge0$), $\CVaR_\alpha$
($\alpha\in(0,1)$), or $\ERisk_\theta$ ($\theta\ne0$), and let $X$ be $\Tot$ or
$\LimAvg$. Both threshold problems for $\Psi$ and $X$ are SSG-hard for RMDPs
under polynomial-time many-one reductions; for $\E$, $\MV_\lambda$, and $\CVaR_\alpha$,
even under logspace reductions.
\end{theorem}

\begin{theorem}[Discounted-sum reward]
\label{cor:expected}\label{cor:erisk}
Let $\Psi$ be the expected value risk measure $\mathbb{E}$, and let $X$ be $\Disc_\gamma$ with a rational discount factor $\gamma$. Both threshold problems for $\Psi$ and $X$ are SSG-hard under polynomial-time Turing reductions.
\end{theorem}

\section{Related Work}

The study of robust MDPs goes back to \cite{Iyengar2005,Nilim2005}. Efficient algorithms for solving RMDPs under expected discounted-sum rewards were proposed in ~\cite{NilimG03,Iyengar2005,Wiesemann2013,KaufmanS13,TamarMX14,GhavamzadehPC16,HoPW21}, and under expected long-run average rewards in~\cite{TewariB07,GC2023,GoyalG23,WangVAPZ23,Chatterjee2024,MeggendorferWW25,AsadiCGKMP26arxiv}. Robust POMDPs were introduced by \cite{Osogami2015}, and \cite{Bovy2024} proposed game semantics under partial observability.

The relationship between RPOMDPs and POSGs is not a novel observation. \cite{Chatterjee2024} showed that RMDPs under expected discounted-sum and long-run average rewards can be reduced to TBSGs, and \cite{Bovy2024} showed that RPOMDPs can be reduced to POSGs. Most recently, \cite{Latha2026} showed that RPOMDPs and POSGs are equivalent (i.e.~inter-reducible) under omega-regular objectives, a class of logical objectives including reachability, safety, and linear temporal logic objectives, but which does not consider reward objectives. The novelty of our work is that RPOMDPs and POSGs are \emph{equivalent for reward objectives and every law-invariant risk measure}. Our reductions are \emph{law preserving}: they map strategies to strategies that induce the same probability distribution of the reward objective, up to a normalizing constant factor that depends only on the type of the reward objective.

Law invariance is a basic property of risk measures~\cite{Kusuoka2001,FollmerSchied2016}. Variance-based criteria in MDPs are studied in \cite{Sobel1982,FilarKallenbergLee1989,MannorTsitsiklis2013,PiribauerSankurBaier2022}; CVaR in \cite{Rockafellar2000,Chow2015,KretinskyMeggendorfer2018,PiribauerBaier2020,Meggendorfer2022}; and entropic risk in \cite{HowardMatheson1972}. For TBSGs, they are studied in \cite{BaierEntropicRisk2023}. Time-consistent dynamic risk measures are developed in \cite{Ruszczynski2010}.The works \cite{Osogami2012,Chow2015} relate risk sensitivity to robustness in MDPs, however they do not propose complexity results or algorithms for solving RMDPs with reward objectives under these risk measures.

\section{Conclusion}
\label{sec:conclusion}

We showed that robust POMDPs and alternating-control POSGs are equivalent in a strong, distributional sense. We designed logspace reductions between the RPOMDP and ac-POSG models, that map strategies to strategies
and preserve the distribution of the reward objective, up to a constant factor, for any of our four reward objectives and any law-invariant risk measure. This allows us to derive a number of new computational complexity results for R(PO)MDPs.

Several questions remain open: the exact complexity of finite-horizon total reward RPOMDPs beyond $\PSPACE$-hardness; the decidability of the non-strict threshold problems for discounted $\E$ and $\CVaR_\alpha$, and of both threshold problems for discounted mean-variance and entropic risk, under partial observation; and upper bounds for mean-variance and CVaR in robust MDPs, where only SSG-hardness is known. It would also be interesting to extend the equivalence beyond $(s,a)$-rectangular uncertainty.

\bibliographystyle{plainnat}
\bibliography{references}

\appendix

\makeatletter
\newcommand{\appnumber}[1]{%
  \@addtoreset{#1}{section}%
  \expandafter\gdef\csname the#1\endcsname{\thesection.\arabic{#1}}%
  \expandafter\gdef\csname theH#1\endcsname{\theHsection.\arabic{#1}}}
\makeatother
\appnumber{theorem}
\appnumber{lemma}
\appnumber{proposition}
\appnumber{corollary}
\appnumber{definition}
\appnumber{remark}
\appnumber{example}
\newpage
\section*{Overview of the Appendix}
\newcommand{\appitem}[3]{\item[\textbf{\ref{#1}.}]\textbf{#2}\hfill{\small p.~\pageref{#1}}\\[-1pt]{\small #3}}
\newcommand{\appsub}[3]{\item[\textbf{\ref{#1}}]#2\hfill{\small p.~\pageref{#1}}\\[-1pt]{\small #3}}
\newenvironment{appsubs}{\begin{enumerate}[labelwidth=2.2em,leftmargin=2.7em,labelsep=0.5em,align=left,itemsep=1pt,topsep=1pt]}{\end{enumerate}}
\begin{enumerate}[labelwidth=1.8em,leftmargin=2.3em,labelsep=0.5em,align=left,itemsep=3pt,topsep=3pt]
  \appitem{app:compile}{Compiling prescriptions into ordinary strategies}
    {Two compilations of games with $\pmin$-prescriptions into ordinary games: the chain gadget for
     $\Tot_N$, $\Tot$, and $\LimAvg$ (Lemma~\ref{lem:compile}), and the vector-action compilation for
     discounted returns (Proposition~\ref{prop:compile-disc}). Examples~\ref{ex:chain} and~\ref{ex:vector}
     show why the chain does not preserve $\Disc_\gamma$ while vector actions do.}
  \appitem{app:equiv}{Proofs for Section~\ref{sec:equiv}}
    {The two law-preserving constructions and the umbrella theorem.}
  \begin{appsubs}
    \appsub{app:lemfwd}{Forward construction}
      {Lemma~\ref{lem:fwd}: from an RPOMDP to an ac-POSG, with a bijection of strategies and equal
       return laws up to the factor $c$; consequences for ordinary POSGs
       (Corollary~\ref{cor:ordinary} and its discounted version).}
    \appsub{app:lembwd}{Backward construction}
      {Lemma~\ref{lem:bwd}: from an ac-POSG to an RPOMDP, with tagged states and dummy states; the
       return laws are preserved exactly.}
    \appsub{app:thmum}{The umbrella theorem}
      {Theorem~\ref{thm:umbrella}: equality of values for every law-invariant risk measure, and the
       logarithmic-space bound.}
    \appsub{app:scale}{Scaling identities}
      {How $\E$, $\MV_\lambda$, $\CVaR_\alpha$, and $\ERisk_\theta$ react to multiplying the return
       by $c$ (Equation~\eqref{eq:scaling}), used for Corollary~\ref{cor:standard}.}
  \end{appsubs}
  \appitem{app:complexity}{Proofs for Section~\ref{sec:complexity}}
    {Lower bounds, undecidability, and decidability results for the reward objectives.}
  \begin{appsubs}
    \appsub{app:auxfacts}{Auxiliary facts}
      {Monotone maps commute with $\inf$ and $\sup$ (F1); memoryless deterministic saddle points
       persist on restricted strategy classes (F2).}
    \appsub{app:bernoulli}{Derivation of the profiles}
      {The four Bernoulli profiles computed from the definitions, and
       Lemma~\ref{lem:bernoulli-standard}: they are monotone and strict at the top.}
    \appsub{app:pspace-hard}{Finite horizon}
      {Theorem~\ref{thm:finite}: $\PSPACE$-hardness, from~\cite{PapadimitriouTsitsiklis1987}, by adding a
       step counter and rewarding survival.}
    \appsub{app:thegap}{The gap lemma and total and average returns}
      {Lemma~\ref{lem:gap}: gap POMDPs from two-counter machines; Theorem~\ref{thm:undec}:
       undecidability of $\Tot$ and $\LimAvg$ with a fixed rational threshold.}
    \appsub{app:thmdiscvar}{Discounted CVaR}
      {Lemma~\ref{lem:cvar-positive}: positivity of $\CVaR$; Theorem~\ref{thm:disc-cvar}:
       undecidability of the strict threshold problem.}
    \appsub{app:propapprox}{Decidability and approximation}
      {Proposition~\ref{prop:approx}: finite-horizon thresholds are decidable via the first-order
       theory of the reals; discounted values are approximated by truncation.}
    \appsub{app:thmssg}{SSG-hardness for total and average returns}
      {Theorem~\ref{thm:ssg}: stopping games, the threshold shift for $\CVaR$, rewards that make
       every return law Bernoulli, the backward construction, and the thresholds.}
    \appsub{app:thmcorexp}{SSG-hardness for expected discounted reward}
      {Theorem~\ref{cor:expected}: the source problem for discounted TBSGs, the RMDP built from a game,
       its value computed by a verification argument, and the composition of the reductions.}
  \end{appsubs}
\end{enumerate}

\begin{center}
    \Large \bf Appendix
\end{center}
\appendix

Recall that at the end of section~\ref{sec:rpomdp}, we assume all actions are available at all the states of POSG. In this appendix, we
present reductions from one model to the another (RPOMDP to POSG and vice-versa), as well as POSGs with prescriptions
to ordinary POSGs. In some of these proofs, while going from one model to the other, the set of actions
are expanded, and in some transitions we use only the newly defined actions.  Indeed, as seen in the main paper, the remaining actions
are not ``missing'', one can always complete the transitions by directing the transitions on the remaining actions
to a sink which is losing for the owner of the said state. The actions that are depicted on transitions from state $s$ (prior to completion) are the \emph{effective}
ones, in the same spirit as \cite{Latha2026}.

To ease our notations in proofs, we define $A(s)$ to denote the set of effective actions at $s$.

\section{Compiling prescriptions into ordinary strategies}
\label{app:compile}
Let $\sg$ be an ac-POSG, with $\pmin$-strategies as in Section~\ref{sec:posg} (prescriptions). A $\pmin$-strategy is \emph{ordinary} if $\pi(h)(t)=\pi(h)(t')$ whenever $\Zmin(t)=\Zmin(t')$ (so $\pi$ is a map $H_{min}\to\Dist(\Amin)$), and $\pmax$-strategies are always ordinary. Given $\sg$ with prescriptions, we build a game $\sg'$ (the \emph{compiled game}) in which $\pmin$ has ordinary strategies only, and relate the two games. For an ac-POSG $\mathcal H$ with observation functions $\Zmax^{\mathcal H}$ and $\Zmin^{\mathcal H}$ and a prefix $u=s_0a_0s_1\cdots s_k$ of $\mathcal H$ we write $h^{\pmax}_{\mathcal H}(u)=\Zmax^{\mathcal H}(s_0)\cdots\Zmax^{\mathcal H}(s_k)$ and $h^{\pmin}_{\mathcal H}(u)=\Zmin^{\mathcal H}(s_0)\cdots\Zmin^{\mathcal H}(s_k)$ for the observation histories of the two players, as in Section~\ref{sec:posg}.

\paragraph{The compiled game.} For $o\in\Omin$, let $C_o=\{t\in\Smin:\Zmin(t)=o\}=\{t^o_1,\dots,t^o_{m_o}\}$ be a fixed enumeration of the class of $o$, and let $\mathrm{pos}(t^o_i)=i$. Let $k=\max_o m_o$ be the size of the largest class. For $1\le i\le k$ let $A_o(i)=A(t^o_i)$ if $i\le m_o$ and $A_o(i)=A(t^o_1)$ if $m_o<i\le k$. The idea is to replace every $\pmin$-state $t$ by a chain of $k$ \emph{slots}, in which $\pmin$ chooses an action at every slot, but only the choice at the slot $\mathrm{pos}(t)$ is the \emph{real} one. Since $\pmin$ observes the slot but not $t$, a strategy of $\pmin$ in the chain can only depend on the class $C_o$ and the slot $i$, and the choice at slot $i$ plays the role of the prescription for $t^o_i$.

Let $\sg=(\Smax,\Smin,\Amax,\Amin,\delta,s_0,\Omax,\Omin,\Zmax,\Zmin,r)$ be an ac-POSG and let $\bot\notin\Amax\cup\Amin$ and $\star,\circledcirc$ be fresh symbols. For $t\in\Smin$, write $o=\Zmin(t)$ and $p=\mathrm{pos}(t)$. The \emph{compiled game} is the ac-POSG\\ $\sg'=(S'_{\pmax},S'_{\pmin},A'_{\pmax},A'_{\pmin},\delta',s_0,O'_{\pmax},O'_{\pmin},\Zmax',\Zmin',r')$ defined as follows.
\begin{itemize}
\item \emph{States.} $S'_{\pmax}=\Smax\cup D$ and $S'_{\pmin}=X$, where
\begin{align*}
X&=\{x_{t,i}:t\in\Smin,\ 1\le i\le \mathrm{pos}(t)\}\cup\{x_{t,b,i}:t\in\Smin,\ \mathrm{pos}(t)<i\le k,\ b\in A(t)\},\\
D&=\{d_{t,i}:t\in\Smin,\ 1\le i<\mathrm{pos}(t)\}\cup\{d_{t,b,i}:t\in\Smin,\ \mathrm{pos}(t)\le i<k,\ b\in A(t)\}.
\end{align*}
Here, $A(t)$ denotes set of effective actions of state $t$ in $\sg$ and $A'(t)$ denotes set of effective actions of state $t$ in $\sg'$. The state $x_{t,i}$ is the slot $i$ of the chain of $t$ before the real choice has been made (the choice at $i=\mathrm{pos}(t)$ is the real one), and $x_{t,b,i}$ is the slot $i$ after the real choice $b$ has been made and recorded in the state. We call $i$ the \emph{slot index} of the states $x_{t,i}$, $x_{t,b,i}$, $d_{t,i}$ and $d_{t,b,i}$.
\item \emph{Actions.} $A'_{\pmax}=\Amax\cup\{\bot\}$ and $A'_{\pmin}=\Amin$. The available actions are $A'(s)=A(s)$ for $s\in\Smax$, $A'(d)=\{\bot\}$ for $d\in D$, and $A'(x)=A_o(i)$ for every $x\in X$ of the form $x_{t,i}$ or $x_{t,b,i}$, where $o=\Zmin(t)$.
\item \emph{Transitions and rewards.} Every transition has reward $0$, except for those marked with a reward. Let $t\in\Smin$, $o=\Zmin(t)$, and $p=\mathrm{pos}(t)$.
\begin{enumerate}
\item[(E)] \emph{Entry.} For $s\in\Smax$ and $a\in A(s)$: $\delta'(s,a)(x_{t,1})=\delta(s,a)(t)$, with reward $r'(s,a,x_{t,1})=r(s,a,t)$.
\item[(C1)] For $1\le i<p$ and $b\in A_o(i)$: $\delta'(x_{t,i},b)=\dirac{d_{t,i}}$ and $\delta'(d_{t,i},\bot)=\dirac{x_{t,i+1}}$.
\item[(C2)] If $p<k$: for $b\in A(t)$, the real choice leads to $\delta'(x_{t,p},b)=\dirac{d_{t,b,p}}$. For $p<i<k$ and $b'\in A(t)$, $b\in A_o(i)$: $\delta'(x_{t,b',i},b)=\dirac{d_{t,b',i}}$. For $p\le i<k$: $\delta'(d_{t,b',i},\bot)=\dirac{x_{t,b',i+1}}$.
\item[(X)] \emph{Exit.} If $p=k$: for $b\in A(t)$ and $s'\in\Smax$, $\delta'(x_{t,k},b)=\delta(t,b)$ with reward $r'(x_{t,k},b,s')=r(t,b,s')$. If $p<k$: for $b'\in A(t)$ and $b\in A_o(k)$, $\delta'(x_{t,b',k},b)=\delta(t,b')$ with reward $r'(x_{t,b',k},b,s')=r(t,b',s')$.
\end{enumerate}
\item \emph{Observations.} $O'_{\pmax}=\Omax\cup(\Omax\times\{\star\})$ and $O'_{\pmin}=\Omin\cup(\Omin\times[k])\cup(\Omin\times[k]\times\{\circledcirc\})$ where $[k] = \{1,2,\cdots, k\}$. For $s\in\Smax$, $\Zmax'(s)=\Zmax(s)$ and $\Zmin'(s)=\Zmin(s)$. For all chain states of $t$ (that is, all states in $X\cup D$ with the first index $t$), with $o=\Zmin(t)$ and slot index $i$:
\[
\Zmax'=(\Zmax(t),\star),\qquad \Zmin'(x_{\cdot,i})=(o,i),\qquad \Zmin'(d_{\cdot,i})=(o,i,\circledcirc).
\]
\end{itemize}

\paragraph{Properties of $\sg'$.} Every state of $S'_{\pmax}$ moves to a state of $S'_{\pmin}$ and vice versa, so $\sg'$ is an ac-POSG. A transition from min-state $t$ to $s'$ on action $b$ in $\sg$ with reward $r(t,b,s')$ corresponds in $\sg'$ to the path
\[
x_{t,1}\to y_1\to x_{\cdot,2}\to y_2\to\cdots\to y_{k-1}\to x_{\cdot,k}\to s'
\]
with $y_i\in D$, which has $k$ states of $X$, $k-1$ states of $D$, and $2k-1$ transitions, the last of which earns the reward $r(t,b,s')$. Together with the entry transition, a round of $\sg$ (a $\pmax$-transition and a $\pmin$-transition) takes $2k$ transitions in $\sg'$. The slots reveal nothing: the observations of both players along a chain do not depend on $t$ within $C_o$, and the chain has a fixed length.

\paragraph{Size.} The game $\sg'$ has polynomial size in $\sg$. Indeed, let $A_{\mathrm{min}}^{\#}=\max_{t\in\Smin}|A(t)|$. For every $t\in\Smin$, the chain of $t$ has at most $k+(k-1)+2(k-1)A_{\mathrm{min}}^{\#}\le 2k(1+A_{\mathrm{min}}^{\#})$ states, so
\[
|S'|\le|\Smax|+2k\,|\Smin|\,(1+A_{\mathrm{min}}^{\#}),
\]
and $k\le|\Smin|$. The action sets are $\Amax\cup\{\bot\}$ and $\Amin$. Every state of $S'$ has at most $\max\{|\Amax|,|\Amin|\}$ available actions, and every transition distribution is either a Dirac distribution or a copy of a distribution of $\sg$: at the entry, $\delta'(s,a)$ is a copy of $\delta(s,a)$ in which every target $t\in\Smin$ is renamed to $x_{t,1}$, and at the exit, $\delta'(x,b)$ is a copy of $\delta(t,b')$ with unchanged targets, defined at a chain state $x$ of $t$ instead of at $t$ (so the source is renamed). The observations are pairs or triples of an observation of $\sg$ and a number up to $k$, and the rewards are rewards of $\sg$ or $0$. Hence the encoding of $\sg'$ is polynomial in the encoding of $\sg$.

\paragraph{Returns.} Let $\rho'$ be a play of $\sg'$. It consists of rounds: after the real state $s_j$ and an action $a_j$, the play traverses the chain of a state $t_j$, whose real choice is $b_j$, and moves to $s_{j+1}$. Let $\mathrm{pr}(\rho')=s_0a_0t_0b_0s_1a_1t_1b_1\cdots$ be the play of $\sg$ obtained by erasing the chain states and keeping the action $b_j$ chosen at the slot $\mathrm{pos}(t_j)$. The same definition applies to a prefix $v$ of $\sg'$ that ends in a state of $\Smax$, which consists of complete rounds, and gives a prefix $\mathrm{pr}(v)$ of $\sg$ that ends in a state of $\Smax$. Let $(r_n)_{n\ge0}$ be the reward sequence of $\mathrm{pr}(\rho')$, so that $r_{2j}=r(s_j,a_j,t_j)$ and $r_{2j+1}=r(t_j,b_j,s_{j+1})$. By the definition of $\sg'$, the $n$-th transition of $\rho'$ earns $r_{2j}$ if $n=2kj$, earns $r_{2j+1}$ if $n=2kj+2k-1$, and earns $0$ otherwise, so the reward sequence of $\rho'$ is
\[
r_0,\,0^{2k-2},\,r_1,\;r_2,\,0^{2k-2},\,r_3,\;\dots
\]
For a return $X$ of $\sg$, let $X'$ be the return of $\sg'$ and $c$ the scaling factor given by
\[
\begin{array}{l|ccc}
X & \Tot_N & \Tot & \LimAvg\\\hline
X' & \Tot_{N'} & \Tot & \LimAvg\\
c & 1 & 1 & 1/k
\end{array}
\qquad
N'=\begin{cases}kN & \text{if } N \text{ is even},\\ k(N-1)+1 & \text{if } N \text{ is odd}.\end{cases}
\]
The factor $c$ is defined by the requirement $X'(\rho')=c\cdot X(\mathrm{pr}(\rho'))$ for every play $\rho'$ of $\sg'$ (proved in the proof of Lemma~\ref{lem:compile} below). For $\Tot_N$ and $\Tot$, the plays $\rho'$ and $\mathrm{pr}(\rho')$ contain the same rewards, in the same order, and $N'$ is the number of transitions of $\rho'$ that contain the first $N$ rewards of $\mathrm{pr}(\rho')$ (each full round has $2k$ transitions, and an odd $N$ has one extra $\pmax$-transition, which is a single transition of $\sg'$), so $c=1$. For $\LimAvg$, the rewards are also the same, but $2$ rewards of $\sg$ are spread over $2k$ transitions of $\sg'$, so the average per transition of $\sg'$ is $1/k$ times the average per transition of $\sg$, and $c=1/k$.

\begin{lemma}[Compilation]
\label{lem:compile}
Let $X\in\{\Tot_N,\Tot,\LimAvg\}$ and let $X'$ and $c$ be as above. There is a bijection $\sigma\mapsto\sigma'$ from $\Sigma_\sg$ to $\Sigma_{\sg'}$ and a map $\mathrm{dec}$ from the ordinary $\pmin$-strategies of $\sg'$ \emph{onto} the prescriptions $\Pi_\sg$, such that for all $\sigma\in\Sigma_\sg$ and all ordinary $\pi'$,
\[
\law^{\sigma',\pi'}_{X'}=\push{c}\,\law^{\sigma,\mathrm{dec}(\pi')}_X .
\]
Recall, $\push{c}$ from section~\ref{sec:prelim}. Consequently, for every law functional $\Psi$, the value of $\sg'$ with ordinary strategies satisfies $\Val_{\sg'}(\Psi,X')=\Val_\sg(\Psi\circ\push{c},X)$.
\end{lemma}

\begin{proof}
\emph{Histories.} Let $W$ be the set of \emph{finite} observation histories $w=o_0o_1\cdots o_{2n}$ ($n\ge0$) of $\sg$ that end in a $\pmax$-state, so that the letters with even index belong to $\pmax$-states and those with odd index belong to $\pmin$-states. For $o\in\Omin$, let $\xi_o=(o,1)(o,1,\circledcirc)(o,2)(o,2,\circledcirc)\cdots(o,k-1,\circledcirc)(o,k)$ be the word of the $\pmin$-observations of a chain, which has length $2k-1$ and depends on $o$ only, and let $\xi_o^{\le i}$ be its prefix that ends in $(o,i)$. For $w\in W$, let $\Theta^{\pmax}(w)$ and $\Theta^{\pmin}(w)$ be the finite words obtained from $w$ by replacing every letter $o$ with odd index by the block $(o,\star)^{2k-1}$ and by the block $\xi_o$, respectively. Both maps are injective, because all blocks have the same length $2k-1$. Let $u$ be a prefix of $\sg$ that ends in $s\in\Smax$, and let $v$ be a prefix of $\sg'$ with $\mathrm{pr}(v)=u$ that ends in $s$ (here $\mathrm{pr}$ erases the chain states of complete rounds, as above). Then
\[
h^{\pmax}_{\sg'}(v)=\Theta^{\pmax}(h^{\pmax}_{\sg}(u)),\qquad h^{\pmin}_{\sg'}(v)=\Theta^{\pmin}(h^{\pmin}_{\sg}(u)).
\]
Suppose now that $v$ is extended by the entry transition to $x_{t,1}$ and by the following chain up to the slot $i$, where $o=\Zmin(t)$. The map $\Theta^{\pmin}$ is applied to the history $h^{\pmin}_{\sg}(u)$ \emph{before} the visit of $t$, which ends in the observation of the $\pmax$-state $s$, and the block $\xi_o^{\le i}$ of the current visit is appended. Hence the $\pmin$-history of $\sg'$ at the slot $i$ is $\Theta^{\pmin}(h^{\pmin}_{\sg}(u))\,\xi_o^{\le i}$, which depends on $h^{\pmin}_{\sg}(u)$, $o$ and $i$, but \emph{not} on $t\in C_o$. The $\pmax$-histories of $\sg'$ at the states of $\Smax$ are exactly the words $\Theta^{\pmax}(w)$, and the symbol $\star$ separates them from the histories at $d$-states, where $\pmax$ has the single action $\bot$.

\emph{Example} ($k=2$). Let $u=s_0a_0t_0b_0s_1$. Its observation histories are
$h^{\pmin}_{\sg}(u)=o_0o_1o_2$ and $h^{\pmax}_{\sg}(u)=q_0q_1q_2$, where
$o_0=\Zmin(s_0)$, $o_1=\Zmin(t_0)$, $o_2=\Zmin(s_1)$ and
$q_0=\Zmax(s_0)$, $q_1=\Zmax(t_0)$, $q_2=\Zmax(s_1)$. The letters $o_0,o_2$ and $q_0,q_2$
are observations of the $\pmax$-states $s_0$ and $s_1$ (even indices), so $\Theta^{\pmax}$ and
$\Theta^{\pmin}$ copy them unchanged. Only the letters $o_1$ and $q_1$ of the $\pmin$-state
$t_0$ (odd index) are replaced by blocks of $2k-1=3$ letters. Let the next state be $t$ with
$\Zmin(t)=o$. In $\sg$, $\pmin$ has the history $o_0o_1o_2\,o$ at $t$. In $\sg'$, after the
round of $t_0$ and the entry transition to $x_{t,1}$, we have
\begin{align*}
h^{\pmax}_{\sg'}(v)&=\Theta^{\pmax}(q_0q_1q_2)=q_0\,(q_1,\star)(q_1,\star)(q_1,\star)\,q_2,\\
h^{\pmin}_{\sg'}(v)&=\Theta^{\pmin}(o_0o_1o_2)=o_0\,(o_1,1)(o_1,1,\circledcirc)(o_1,2)\,o_2,
\end{align*}
and the $\pmin$-histories at the slots $1$ and $2$ of the chain of $t$ are $\Theta^{\pmin}(o_0o_1o_2)\,(o,1)$ and $\Theta^{\pmin}(o_0o_1o_2)\,(o,1)(o,1,\circledcirc)(o,2)$, respectively.

\emph{Strategies.} Let $\sigma\in\Sigma_\sg$. In $\sg'$, $\pmax$ moves at the states of $\Smax$ and at the states of $D$. At a $\pmax$-history $\Theta^{\pmax}(w)$ ends in a state of $\Smax$, we let $\sigma'(\Theta^{\pmax}(w))=\sigma(w)$. At a $\pmax$-history $h$ ending in a state $d\in D$ the only available action is $\bot$, so there is no choice, and $\sigma'(h)=\dirac{\bot}$. Since $\Theta^{\pmax}$ is injective, $\sigma'$ is well defined, and $\sigma$ can be recovered from $\sigma'$ by $\sigma(w)=\sigma'(\Theta^{\pmax}(w))$. Conversely, every $\pmax$-strategy of $\sg'$ plays $\bot$ at the histories of states of $D$, so it is determined by its values at the histories $\Theta^{\pmax}(w)$. Hence $\sigma\mapsto\sigma'$ is a bijection between $\Sigma_\sg$ and $\Sigma_{\sg'}$.

For an ordinary $\pmin$-strategy $\pi'$ of $\sg'$, define the prescription $\mathrm{dec}(\pi')$ by
\[
\mathrm{dec}(\pi')(w^-o)(t^o_i)=\pi'\bigl(\Theta^{\pmin}(w^-)\,\xi_o^{\le i}\bigr)\qquad(1\le i\le m_o),
\]
where $w^-o$ is a $\pmin$-history of $\sg$ that ends in $o$ (for states $t\in\Smin$ with $\Zmin(t)\ne o$, the value $\mathrm{dec}(\pi')(w^-o)(t)$ is never used and is chosen arbitrarily). This is a mixed action in $\Dist(A_o(i))=\Dist(A(t^o_i))$. Conversely, for a prescription $\pi$ let $\mathrm{enc}(\pi)$ be the ordinary strategy of $\sg'$ with
\[
\mathrm{enc}(\pi)\bigl(\Theta^{\pmin}(w^-)\,\xi_o^{\le i}\bigr)=\pi(w^-o)(t^o_i)\quad(1\le i\le m_o),
\]
and an arbitrary element of $\Dist(A_o(i))$ for $i>m_o$ (the choice at such a slot is never real). It is ordinary, because its argument depends on the observation word only, and it is legal, because $A_o(i)=A(t^o_i)$.
By definition, $\mathrm{dec}(\mathrm{enc}(\pi))=\pi$, so $\mathrm{dec}$ is onto $\Pi_\sg$. Conversely, $\mathrm{enc}(\mathrm{dec}(\pi'))$ agrees with $\pi'$ at all slots $i\le m_o$ and can differ from $\pi'$ only at slots $i>m_o$, where the choice is never real. We identify strategies that differ only at histories that cannot occur, and ordinary $\pmin$-strategies of $\sg'$ that differ only at such slots.

\emph{Measures.} Fix $\sigma\in\Sigma_\sg$ and an ordinary $\pi'$ of $\sg'$, let $\pi=\mathrm{dec}(\pi')$, and write $P=\Prob^{\sigma,\pi}_\sg$ and $P'=\Prob^{\sigma',\pi'}_{\sg'}$. For a prefix $u=s_0a_0t_0b_0s_1\cdots s_j$ of $\sg$ that ends in a state of $\Smax$ (that is, $u$ consists of $j$ rounds), let
\[
V(u)=\{v:\ v\text{ is a prefix of }\sg'\text{ that ends in a state of }\Smax\text{ and }\mathrm{pr}(v)=u\}.
\]
The set $V(u)$ is finite, and every $v\in V(u)$ consists of $j$ rounds of $\sg'$, so it has exactly $2kj$ transitions. Hence no element of $V(u)$ is a proper prefix of another, and the cylinders $\Cyl(v)$ of $\sg'$ for $v\in V(u)$ are pairwise disjoint. We first compare the probabilities of cylinders: we show that for every prefix $u$ of $\sg$ that ends in a state of $\Smax$,
\[
P\bigl(\Cyl(u)\bigr)=\sum_{v\in V(u)}P'\bigl(\Cyl(v)\bigr).\tag{$*$}
\]
The proof is by induction on the number $j$ of rounds of $u$.

\smallskip
\noindent\emph{Base case.} For $u=s_0$ we have $V(u)=\{s_0\}$, and both sides of $(*)$ are $1$.

\smallskip
\noindent\emph{Induction step.} Let $u$ end in $s\in\Smax$, and let $u^+=u\,a\,t\,b\,s'$ be an extension of $u$ by one round of $\sg$: $\pmax$ plays $a\in A(s)$, the game moves to $t\in\Smin$, $\pmin$ plays $b\in A(t)$, and the game moves to $s'\in\Smax$. Let $o=\Zmin(t)$, $p=\mathrm{pos}(t)$ (so $t=t^o_p$), and $w^-=h^{\pmin}_{\sg}(u)$, so that $\pmin$ has the history $w^-o$ at $t$. By the definition of $P$,
\[
P\bigl(\Cyl(u^+)\bigr)=P\bigl(\Cyl(u)\bigr)\cdot f,\qquad f=\sigma\bigl(h^{\pmax}_{\sg}(u)\bigr)(a)\,\delta(s,a)(t)\,\pi(w^-o)(t)(b)\,\delta(t,b)(s').
\]
Let $B=\{\bar b=(b_1,\dots,b_k)\in A_o(1)\times\cdots\times A_o(k):b_p=b\}$ be the set of tuples of slot actions whose real component is $b$. For $v\in V(u)$ and $\bar b\in B$, let
\[
v_{\bar b}=v\,a\,x_{t,1}\,b_1\,y_1\,\bot\,y_2\,b_2\cdots y_{k-1}\,\bot\,x_k\,b_k\,s',
\]
where the chain states $y_i\in D$ and $x_{i}\in X$ are determined by $t$, $b$, and the position $i$, by the definition of $\sg'$. Then $v_{\bar b}\in V(u^+)$, and the map $(v,\bar b)\mapsto v_{\bar b}$ from $V(u)\times B$ to $V(u^+)$ is a bijection. Indeed, it is injective because $v_{\bar b}$ contains $v$ and the actions $b_1,\dots,b_k$. For surjectivity, let $v'\in V(u^+)$. Truncating $v'$ after its first $j$ rounds gives a prefix $v$ with $\mathrm{pr}(v)=u$, that is, $v\in V(u)$, and the last round of $v'$ traverses the chain of some state $\tilde t$ with real choice $\tilde b$ and ends in $s'$. Since $\mathrm{pr}(v')=u\,a\,\tilde t\,\tilde b\,s'$ must equal $u^+$, we get $\tilde t=t$ and $\tilde b=b$. Let $\bar b$ be the tuple of the actions chosen at the $k$ slots in this round. Then $\bar b\in B$, because its component at the slot $p=\mathrm{pos}(t)$ is the real choice $b$, and $v'=v_{\bar b}$.

We compute $P'(\Cyl(v_{\bar b}))$ by the definition of $P'$, as $P'(\Cyl(v))$ times the product of the probabilities of the transitions of the last round. The entry transition has probability $\sigma'(h^{\pmax}_{\sg'}(v))(a)\,\delta'(s,a)(x_{t,1})=\sigma(h^{\pmax}_{\sg}(u))(a)\,\delta(s,a)(t)$, because $h^{\pmax}_{\sg'}(v)=\Theta^{\pmax}(h^{\pmax}_{\sg}(u))$ and $\sigma'(\Theta^{\pmax}(w))=\sigma(w)$. At the slot $i$, $\pmin$ plays $b_i$ with probability $\pi'(\Theta^{\pmin}(w^-)\xi_o^{\le i})(b_i)$, since this is its history there. All other transitions of the chain are deterministic: $\pmax$ plays $\bot$ at the $d$-states with probability $1$ and the slot transitions are Dirac distributions. The exit transition has probability $\delta'(x_k,b_k)(s')=\delta(t,b)(s')$, which does not depend on $b_k$ if $p<k$ (the real choice $b$ is recorded in the state), and equals $\delta(t,b_p)(s')=\delta(t,b)(s')$ if $p=k$. Hence
\[
P'\bigl(\Cyl(v_{\bar b})\bigr)=P'\bigl(\Cyl(v)\bigr)\cdot\sigma\bigl(h^{\pmax}_{\sg}(u)\bigr)(a)\,\delta(s,a)(t)\cdot\prod_{i=1}^k\pi'\bigl(\Theta^{\pmin}(w^-)\xi_o^{\le i}\bigr)(b_i)\cdot\delta(t,b)(s').
\]
We sum over $\bar b\in B$. The product over $i$ splits as $\pi'(\Theta^{\pmin}(w^-)\xi_o^{\le p})(b)$ for $i=p$, which is $\pi(w^-o)(t)(b)$ by the definition of $\mathrm{dec}$, times the factors for $i\ne p$. For each $i\ne p$, $\pi'(\Theta^{\pmin}(w^-)\xi_o^{\le i})$ is a probability distribution on $A_o(i)$, so summing over $b_i\in A_o(i)$ gives $1$. Therefore
\[
\sum_{\bar b\in B}P'\bigl(\Cyl(v_{\bar b})\bigr)=P'\bigl(\Cyl(v)\bigr)\cdot f .
\]
Using the bijection above, the induction hypothesis $(*)$ for $u$, and the formula for $P(\Cyl(u^+))$,
\[
\sum_{v'\in V(u^+)}P'\bigl(\Cyl(v')\bigr)=\sum_{v\in V(u)}\ \sum_{\bar b\in B}P'\bigl(\Cyl(v_{\bar b})\bigr)=f\cdot\sum_{v\in V(u)}P'\bigl(\Cyl(v)\bigr)=f\cdot P\bigl(\Cyl(u)\bigr)=P\bigl(\Cyl(u^+)\bigr).
\]
This proves $(*)$.

\smallskip
\noindent\emph{From $(*)$ to the equality of measures.} Every infinite play $\rho'$ of $\sg'$ consists of complete rounds (entry, chain, exit), so $\mathrm{pr}(\rho')$ is a well-defined play of $\sg$, and $\mathrm{pr}$ is defined on all plays of $\sg'$. Fix a finite prefix $u$ of $\sg$ that ends in $\Smax$ and has $j$ rounds. A play $\rho'$ satisfies $\mathrm{pr}(\rho')\in\Cyl(u)$ if and only if the prefix $v$ of $\rho'$ consisting of its first $j$ rounds satisfies $\mathrm{pr}(v)=u$, that is, $v\in V(u)$. Hence
\[
\mathrm{pr}^{-1}\bigl(\Cyl(u)\bigr)=\biguplus_{v\in V(u)}\Cyl(v).
\]
The union is disjoint because all $v\in V(u)$ have the same length ($2kj$ transitions), so two distinct such prefixes are incomparable and their cylinders do not intersect. It is finite because $\sg'$ is finite, so $V(u)$ is finite. Therefore, by $(*)$,
\[
P'\bigl(\mathrm{pr}^{-1}(\Cyl(u))\bigr)=\sum_{v\in V(u)}P'\bigl(\Cyl(v)\bigr)=P\bigl(\Cyl(u)\bigr).
\]
This holds for every cylinder $\Cyl(u)$ of $\sg$ (the cylinder of a prefix that ends in $\Smin$ is the disjoint union of the cylinders of its one-step extensions, so it is covered as well).

In particular, $\mathrm{pr}$ is measurable. The cylinders of $\sg$ generate its $\sigma$-algebra of plays, and the preimage of each generator is a finite union of cylinders of $\sg'$, hence measurable. So the push-forward $\mathrm{pr}_\#P'$, defined by $\mathrm{pr}_\#P'(E)=P'(\mathrm{pr}^{-1}(E))$, is a probability measure on the plays of $\sg$. Hence, $\mathrm{pr}_\#P'$ and $P$ agree on all cylinders. Any two cylinders are either nested or disjoint, so the cylinders together with the empty set are closed under finite intersections, and they generate the $\sigma$-algebra of plays. By the uniqueness theorem for probability measures (two probability measures that agree on an intersection-closed generator agree on the generated $\sigma$-algebra~\cite[Thm.~3.3]{Billingsley1995}), $\mathrm{pr}_\#P'=P$.

\emph{Returns.} We show $X'(\rho')=c\cdot X(\mathrm{pr}(\rho'))$ on every play $\rho'$ of $\sg'$. Let $(r_n)$ be the reward sequence of $\mathrm{pr}(\rho')$. The rewards $r_{2j}$ and $r_{2j+1}$ are earned at the transitions with indices $2kj$ and $2kj+2k-1$ of $\rho'$. For $\Tot$ the two sums coincide. For $\Tot_N$, the first $N'$ transitions of $\rho'$ contain exactly the rewards $r_0,\dots,r_{N-1}$: if $N=2n$, the last of them, $r_{2n-1}$, is earned at the transition $2kn-1$, and $N'=kN=2kn$; if $N=2n+1$, the last one, $r_{2n}$, is earned at the transition $2kn$, and $N'=k(N-1)+1=2kn+1$.

For $\LimAvg$, let $\rho'$ be a play of $\sg'$ and $\rho=\mathrm{pr}(\rho')$ the corresponding play of $\sg$, with rewards $r_0,r_1,\dots$ (all transitions of $\sg$ that carry a reward are indexed from $0$). Let $R=\max|r|$, which is finite since the game is finite. Write $S_m=\sum_{i<m}r_i$ and $A_m=S_m/m$ for $m\ge1$, so that $\LimAvg(\rho)=\liminf_mA_m$ (the case of $\limsup$ is identical). In $\sg'$ the reward $r_{2j}$ is earned at transition $2kj$, the reward $r_{2j+1}$ at transition $2kj+2k-1$, and all other transitions have reward $0$. Let $B_n$ be the average of the first $n$ rewards of $\rho'$, that is, of the rewards earned at the transitions $0,\dots,n-1$. Then $\LimAvg(\rho')=\liminf_nB_n$, and we claim
\[
\liminf_nB_n=\tfrac1k\liminf_mA_m,\qquad \limsup_nB_n=\tfrac1k\limsup_mA_m .
\]
\emph{Step 1: $B_n$ is close to $\frac1kA_{2j}$.} Every $n\ge2k$ lies in $[2kj,2k(j+1))$ for a unique $j=j(n)\ge1$, namely $j(n)=\lfloor n/(2k)\rfloor$, and we distinguish two cases.
\begin{itemize}
\item \emph{Case $n=2kj$.} The first $2kj$ transitions of $\rho'$ are $j$ complete rounds. They contain the rewards $r_0,\dots,r_{2j-1}$ (the rewards $r_{2i}$ and $r_{2i+1}$ of the round $i<j$ are earned at the transitions $2ki$ and $2ki+2k-1$, both below $2kj$) and no others. Hence $B_{2kj}=S_{2j}/(2kj)=\frac1kA_{2j}$.
\item \emph{Case $2kj<n<2k(j+1)$.} The reward $r_{2j}$ is earned at the transition $2kj\le n-1$, so it is among the first $n$ rewards. The reward $r_{2j+1}$ is earned at the transition $2kj+2k-1\ge n$, so it is not. Together with the rewards of the first $j$ rounds, the sum of the first $n$ rewards is therefore $S_{2j}+r_{2j}$, and $B_n=(S_{2j}+r_{2j})/n$. Recall that $|r_{2j}|\le R$, $|S_{2j}|\le 2jR$, and $2kj<n<2k(j+1)$. Put $\Delta=n-2kj$, so $0<\Delta<2k$. Subtracting and regrouping gives
\[
B_n-\tfrac1kA_{2j}
=\frac{S_{2j}+r_{2j}}{n}-\frac{S_{2j}}{2kj}
=\frac{r_{2j}}{n}+S_{2j}\Bigl(\frac1n-\frac1{2kj}\Bigr)
=\frac{r_{2j}}{n}-S_{2j}\cdot\frac{\Delta}{2kj\,n}.
\]
We bound the two terms separately.
\begin{align*}
\Bigl|\frac{r_{2j}}{n}\Bigr|&\le\frac{R}{n}<\frac{R}{2kj}
&&\text{since }|r_{2j}|\le R\text{ and }n>2kj,\\[2pt]
\Bigl|S_{2j}\cdot\frac{\Delta}{2kj\,n}\Bigr|&\le 2jR\cdot\frac{\Delta}{2kj\,n}=\frac{R\,\Delta}{k\,n}<\frac{2R}{n}<\frac{R}{kj}
&&\text{since }|S_{2j}|\le2jR,\ \Delta<2k,\text{ and }n>2kj.
\end{align*}
Adding the two bounds,
\[
\bigl|B_n-\tfrac1kA_{2j}\bigr|<\frac{R}{2kj}+\frac{R}{kj}=\frac{3R}{2kj}.
\]
\end{itemize}
In both cases $|B_n-\frac1kA_{2j(n)}|\le\frac{3R}{2kj(n)}$, and the right-hand side tends to $0$ as $n\to\infty$, because $j(n)\to\infty$. Hence the sequences $(B_n)_n$ and $(\frac1kA_{2j(n)})_n$ have the same $\liminf$ and the same $\limsup$.

\emph{Step 2: from $(A_{2j(n)})_n$ to $(A_{2j})_j$.} The map $n\mapsto j(n)$ is non-decreasing and takes every value $j\ge1$ exactly $2k$ times. So $(A_{2j(n)})_n$ is the sequence $(A_{2j})_{j\ge1}$ with every term repeated $2k$ times, which does not change its $\liminf$ or $\limsup$. Together with Step 1, $\liminf_nB_n=\frac1k\liminf_jA_{2j}$ and $\limsup_nB_n=\frac1k\limsup_jA_{2j}$.

\emph{Step 3: from even indices to all indices.} For every $m\ge1$,
\[
A_{m+1}-A_m=\frac{S_m+r_m}{m+1}-\frac{S_m}{m}=\frac{r_m}{m+1}-\frac{S_m}{m(m+1)},
\qquad\text{so}\qquad |A_{m+1}-A_m|\le\frac{R}{m+1}+\frac{mR}{m(m+1)}=\frac{2R}{m+1},
\]
using $|S_m|\le mR$. Thus every $A_m$ with $m$ odd is within $2R/m$ of the even-indexed term $A_{m-1}$, and $2R/m\to0$. Consequently the sequences $(A_m)_m$ and $(A_{2\lfloor m/2\rfloor})_m$ have the same $\liminf$ and $\limsup$, and by the repetition argument of Step 2 so do $(A_{2\lfloor m/2\rfloor})_m$ and $(A_{2j})_j$. Therefore $\liminf_jA_{2j}=\liminf_mA_m$ and $\limsup_jA_{2j}=\limsup_mA_m$.

Combining Steps 2 and 3 gives $\LimAvg(\rho')=\frac1k\LimAvg(\rho)$ (for $\liminf$, and likewise for $\limsup$), which is the scaling $c=\frac1k$ for $\LimAvg$.

\emph{Conclusion.} Let $\sigma$ be a strategy of $\pmax$ in $\sg$, and let $\sigma'$ be its image under the bijection $\sigma\mapsto\sigma'$ defined above: $\sigma'(\Theta^{\pmax}(w))=\sigma(w)$ at the histories of prefixes of $\sg'$ that end in $\Smax$ (and $\sigma'=\dirac{\bot}$ at the histories of prefixes that end in $D$). Let $\pi'$ be a strategy of $\pmin$ in $\sg'$, and let $\mathrm{dec}(\pi')$ be the strategy of $\pmin$ in $\sg$ that $\pi'$ induces.

Write $P=P^{\sigma,\mathrm{dec}(\pi')}_{\sg}$ and $P'=P^{\sigma',\pi'}_{\sg'}$. Recall that the law of a return is the distribution of the return under the play measure, $\law^{\sigma,\pi}_X=P^{\sigma,\pi}\circ X^{-1}$, and that $\push{c}(\mu)=\mu\circ(x\mapsto cx)^{-1}$ is the push-forward of $\mu$ under multiplication by $c$. We have $X'=c\cdot X\circ\mathrm{pr}$, that is, $X'(\rho')=c\,X(\mathrm{pr}(\rho'))$ for every play $\rho'$ of $\sg'$, with $c>0$ ($c=1$ or $c=\frac1k$). Let $E\subseteq\mathbb R$ be a Borel set. Then
\begin{align*}
\law^{\sigma',\pi'}_{X'}(E)
&=P'\bigl(\{\rho':c\,X(\mathrm{pr}(\rho'))\in E\}\bigr)
&&\text{definition of the law}\\
&=P'\bigl(\mathrm{pr}^{-1}\bigl(X^{-1}(c^{-1}E)\bigr)\bigr)
&&\text{where }c^{-1}E=\{x:cx\in E\}\\
&=(\mathrm{pr}_\#P')\bigl(X^{-1}(c^{-1}E)\bigr)
&&\text{definition of the push-forward }\mathrm{pr}_\#P'\\
&=P\bigl(X^{-1}(c^{-1}E)\bigr)
&&\mathrm{pr}_\#P'=P\text{ (from }(*)\text{)}\\
&=\law^{\sigma,\mathrm{dec}(\pi')}_X(c^{-1}E)
&&\text{definition of the law}\\
&=\push{c}\bigl(\law^{\sigma,\mathrm{dec}(\pi')}_X\bigr)(E)
&&\text{since }\push{c}(\mu)(E)=\mu(\{x:cx\in E\})=\mu(c^{-1}E)\text{ for every }\mu.
\end{align*}
The sets in the first rows are measurable because $\mathrm{pr}$ and $X$ are measurable and $c^{-1}E$ is Borel. Since $E$ was arbitrary, the first and last entries  give
\[
\law^{\sigma',\pi'}_{X'}=\push{c}\,\law^{\sigma,\mathrm{dec}(\pi')}_X .
\]
Since $\sigma\mapsto\sigma'$ is a bijection and $\mathrm{dec}$ is onto $\Pi_\sg$, the sets of laws $\{\law^{\sigma',\pi'}_{X'}\}$ and $\push{c}\{\law^{\sigma,\pi}_X\}$ coincide for every $\sigma$, and the statement on values follows.
\end{proof}

\paragraph{Discounted returns.} Lemma~\ref{lem:compile} covers $\Tot_N$, $\Tot$ and $\LimAvg$ because these returns do not depend on exactly when a reward is earned, apart from the horizon, which we adjust, and a constant factor for the average. A discounted return does: the reward of the $t$-th transition has weight $\gamma^t$. The chain gadget delays the exit reward of every $\pmin$-visit and makes one round take $2k$ instead of $2$ transitions. To preserve discounted returns for every $\gamma$, we need a compilation that keeps the timing of $\sg$. The idea is to give $\pmin$ the whole prescription as a single action: the available actions at all states of an observation class are the \emph{tuples} that contain one action for each state of the class, and the state that is actually visited uses its own component. This costs an exponential blow-up of the action sets, but changes neither the states nor the timing.

\begin{proposition}[Discounted returns]
\label{prop:compile-disc}
Let $\sg$ be an ac-POSG with $\pmin$-prescriptions and let $\gamma\in(0,1)$ be rational. There is an ordinary ac-POSG $\sg^\times$ with the same states, the same timing and the same rewards on corresponding plays, such that for every $\pmax$-strategy $\sigma$,
\[
\{\law^{\sigma,\pi'}_{\Disc_\gamma}:\pi'\text{ ordinary in }\sg^\times\}=\{\law^{\sigma,\pi}_{\Disc_\gamma}:\pi\in\Pi_\sg\}.
\]
Hence $\Val_{\sg^\times}(\Psi,\Disc_\gamma)=\Val_\sg(\Psi,\Disc_\gamma)$ for all law functionals $\Psi$, with the same discount factor $\gamma$ and no rescaling. The action sets of $\sg^\times$ may be exponentially larger than those of $\sg$.
\end{proposition}
\begin{proof}
\emph{Idea.} A prescription assigns an independent mixed action to every state of an observation class, but only the action of the state that is actually visited is used. An ordinary strategy cannot name the state, but it can choose a whole \emph{tuple} of actions, one for each state of the class. The visited state then reads off its own component. Playing the product of the mixed actions on tuples reproduces the prescription, and the marginals of any distribution on tuples form a prescription. Nothing else in the game changes, so the timing and the rewards are the same and every return, in particular $\Disc_\gamma$, is preserved.

\medskip
\noindent\emph{Step 1: the game $\sg^\times$.} Let $C_o=\{t_1,\dots,t_m\}\subseteq\Smin$ be an observation class of $\pmin$. In $\sg^\times$, every state of $C_o$ gets the same set of actions, namely the tuples
\[
A_o^\times=A(t_1)\times\cdots\times A(t_m).
\]
For a tuple $\beta=(\beta_1,\dots,\beta_m)\in A_o^\times$, the state $t_i$ uses only the $i$-th component:
\[
\delta^\times(t_i,\beta)=\delta(t_i,\beta_i),\qquad r^\times(t_i,\beta,s')=r(t_i,\beta_i,s').
\]
All other components of $\beta$ are ignored at $t_i$. States, observations, the $\pmax$-states with their actions, the transitions of $\pmax$ and their rewards are copied from $\sg$. All states of $C_o$ now have the same actions and $\sg^\times$ is an ac-POSG. The observation functions are unchanged, so \emph{the observation histories of both players are the same words in $\sg$ and in $\sg^\times$}. In particular, observation histories do not contain actions, so the tuples chosen by $\pmin$ do not appear in them. Hence $\pmax$-strategies of the two games are the same objects, and $\sigma$ can be used in both games.

\smallskip
\noindent\emph{Step 2: strategies.} Let $w$ be a $\pmin$-history that ends in the observation $o$. A prescription $\pi\in\Pi_\sg$ assigns to $w$ a mixed action $\pi(w)(t_i)\in\Dist(A(t_i))$ for every $t_i\in C_o$. An ordinary strategy $\pi'$ of $\sg^\times$ assigns to $w$ a single distribution $\pi'(w)\in\Dist(A_o^\times)$ on tuples. We relate the two as follows.
\begin{itemize}
\item \emph{From tuples to prescriptions.} Let $\mathrm{mar}(\pi')$ be the prescription with
\[
\mathrm{mar}(\pi')(w)(t_i)(b)=\sum_{\beta\in A_o^\times:\ \beta_i=b}\pi'(w)(\beta)\qquad(b\in A(t_i)),
\]
that is, $\mathrm{mar}(\pi')(w)(t_i)$ is the $i$-th marginal of $\pi'(w)$. It is a distribution on $A(t_i)$, so $\mathrm{mar}(\pi')\in\Pi_\sg$.
\item \emph{From prescriptions to tuples.} For $\pi\in\Pi_\sg$, let $\pi^\times$ be the ordinary strategy that plays the product distribution
\[
\pi^\times(w)(\beta)=\prod_{i=1}^m\pi(w)(t_i)(\beta_i).
\]
Its $i$-th marginal is $\pi(w)(t_i)$, since the other factors sum to $1$. Hence $\mathrm{mar}(\pi^\times)=\pi$.
\end{itemize}
Thus $\mathrm{mar}$ maps the ordinary strategies of $\sg^\times$ \emph{onto} $\Pi_\sg$. It is in general not injective: different distributions on tuples, for instance correlated ones, can have the same marginals. This does no harm, because in each step only one coordinate of the tuple is used, and a fresh tuple is drawn in the next step.

\smallskip
\noindent\emph{Step 3: the projection of plays.} A play of $\sg^\times$ is $\rho^\times=s_0\,c_0\,s_1\,c_1\cdots$, where $c_j$ is an action of $\pmax$ if $s_j\in\Smax$, and a tuple if $s_j\in\Smin$. Let $\mathrm{pr}(\rho^\times)$ be the play of $\sg$ in which every tuple $c_j=\beta$ chosen at $s_j=t_i$ is replaced by its component $\beta_i$. This is a play of $\sg$, because $\delta^\times(t_i,\beta)=\delta(t_i,\beta_i)$. The map $\mathrm{pr}$ is measurable and keeps the sequence of states, so for every prefix $u$ of $\sg$, the set $\mathrm{pr}^{-1}(\Cyl(u))$ is the disjoint union of the cylinders $\Cyl(u^\times)$ over all prefixes $u^\times$ of $\sg^\times$ with $\mathrm{pr}(u^\times)=u$.

\smallskip
\noindent\emph{Step 4: the measures agree.} Fix $\sigma$ and an ordinary $\pi'$ of $\sg^\times$, and write $\pi=\mathrm{mar}(\pi')$. We show
\[
\Prob_{\sg^\times}^{\sigma,\pi'}\bigl(\mathrm{pr}^{-1}(\Cyl(u))\bigr)=\Prob_\sg^{\sigma,\pi}\bigl(\Cyl(u)\bigr)\quad\text{for every prefix $u$ of $\sg$,}\tag{$*$}
\]
by induction on the length of $u$. For $u=s_0$ both sides are $1$. For the induction step, let $u$ be a prefix that ends in a state $s$, let $w$ be the observation-history of the player who moves from $s$ after $u$ (it is the same in both games, since it only depends on the states of $u$), and let $u\,b\,s'$ be an extension of $u$.
\begin{itemize}
\item \emph{$s\in\Smax$.} The extension has the same probability $\sigma(w)(b)\,\delta(s,b)(s')$ in both games, because $\pmax$, its actions and its transitions are not changed. Since this factor does not depend on how $u$ was reached, $(*)$ for $u$ gives $(*)$ for $u\,b\,s'$.
\item \emph{$s=t_i\in\Smin$.} Let $u^\times$ be a prefix of $\sg^\times$ with $\mathrm{pr}(u^\times)=u$. The cylinders of $\sg^\times$ that project to $u\,b\,s'$ are those of $u^\times\,\beta\,s'$ with $\beta_i=b$, and
\[
\Prob_{\sg^\times}\bigl(\Cyl(u^\times\beta s')\bigr)=\Prob_{\sg^\times}\bigl(\Cyl(u^\times)\bigr)\cdot\pi'(w)(\beta)\cdot\delta(t_i,b)(s').
\]
We sum over all $\beta$ with $\beta_i=b$. By the definition of $\mathrm{mar}$, the sum of the factors $\pi'(w)(\beta)$ is $\pi(w)(t_i)(b)$. This value depends only on $w$ and $t_i$, and not on $u^\times$. Summing then over all $u^\times$ with $\mathrm{pr}(u^\times)=u$ gives
\[
\Prob_{\sg^\times}\bigl(\mathrm{pr}^{-1}(\Cyl(u\,b\,s'))\bigr)=\Prob_{\sg^\times}\bigl(\mathrm{pr}^{-1}(\Cyl(u))\bigr)\cdot\pi(w)(t_i)(b)\,\delta(t_i,b)(s'),
\]
and by $(*)$ for $u$ and the definition of $\Prob_\sg$, the right-hand side equals $\Prob_\sg^{\sigma,\pi}(\Cyl(u\,b\,s'))$.
\end{itemize}

We now pass from cylinders to all measurable sets of plays. Since $\mathrm{pr}$ is measurable (Step~3), $\mu_1=\mathrm{pr}_\#\Prob_{\sg^\times}^{\sigma,\pi'}$, defined by $\mu_1(B)=\Prob_{\sg^\times}^{\sigma,\pi'}(\mathrm{pr}^{-1}(B))$, is a probability measure on the plays of $\sg$. Let $\mu_2=\Prob_\sg^{\sigma,\mathrm{mar}(\pi')}$, also a probability measure on the plays of $\sg$. By the definition of $\mu_1$, equation $(*)$ says exactly that
\[
\mu_1(\Cyl(u))=\mu_2(\Cyl(u))\quad\text{for every prefix $u$ of $\sg$.}
\]
Two cylinder sets are either disjoint or nested, so their intersection is again a cylinder set or empty: the cylinders form a $\pi$-system. It generates the $\sigma$-algebra on plays, which is by definition the one generated by the cylinders. The family of sets on which two probability measures agree is closed under complements and countable disjoint unions, so it is a $\lambda$-system. By Dynkin's $\pi$-$\lambda$ theorem, it contains the $\sigma$-algebra generated by any $\pi$-system it contains. Hence $\mu_1=\mu_2$, that is,
\[
\mathrm{pr}_\#\Prob_{\sg^\times}^{\sigma,\pi'}=\Prob_\sg^{\sigma,\mathrm{mar}(\pi')}.
\]

\smallskip
\noindent\emph{Step 5: the returns agree.} By the choice of $r^\times$, every transition of $\rho^\times$ earns exactly the reward of the corresponding transition of $\mathrm{pr}(\rho^\times)$, and both plays have their transitions at the same positions. So both plays have the same reward sequence $(r_t)_{t\ge0}$, and for every return $X$ that is a function of the reward sequence, in particular for
\[
X = \Disc_\gamma=\sum_{t\ge0}\gamma^t r_t
\]
we have $X(\rho^\times)=X(\mathrm{pr}(\rho^\times))$ for every play. We stress that no discount factor is changed: the $t$-th reward carries the weight $\gamma^t$ in both games, because $\sg^\times$ has the same timing. Combining this with Step 4,
\[
\law^{\sigma,\pi'}_{X}=\Prob_{\sg^\times}^{\sigma,\pi'}\circ(X\circ\mathrm{pr})^{-1}
=\bigl(\mathrm{pr}_\#\Prob_{\sg^\times}^{\sigma,\pi'}\bigr)\circ X^{-1}
=\Prob_\sg^{\sigma,\mathrm{mar}(\pi')}\circ X^{-1}=\law^{\sigma,\mathrm{mar}(\pi')}_{X}.
\]

\smallskip
\noindent\emph{Conclusion.} For every $\pmax$-strategy $\sigma$, Step 5 shows that every law of an ordinary strategy $\pi'$ of $\sg^\times$ is a law of the prescription $\mathrm{mar}(\pi')$ of $\sg$. Conversely, Step 2 shows that every prescription $\pi$ is $\mathrm{mar}(\pi^\times)$, so its law $\law^{\sigma,\pi}_X=\law^{\sigma,\pi^\times}_X$ is a law of $\sg^\times$ under an ordinary strategy. Hence the two sets of laws coincide. The value of a game is determined by these sets (it is the same expression in $\Psi$ of the laws, whatever the order of the optimization over $\sigma$ and over the strategy of $\pmin$), and thus $\Val_{\sg^\times}(\Psi,X)=\Val_\sg(\Psi,X)$ for all $\Psi$. Finally, the only blow-up is in the action sets: $|A_o^\times|=\prod_{i\le m}|A(t_i)|$ for a class of $m$ states.
\end{proof}

\begin{figure*}[t]
\centering
\hspace*{\dimexpr -1in-\hoffset-\oddsidemargin\relax}%
\makebox[\paperwidth][c]{%
\begin{subfigure}[t]{0.39\paperwidth}
\centering
\begin{tikzpicture}[
    maxNode/.style={circle, draw=maxcolor, thick, minimum size=0.55cm, font=\scriptsize},
    sinkNode/.style={circle, draw=maxcolor, thick, double, double distance=0.6pt, minimum size=0.55cm, font=\scriptsize},
    minNode/.style={rectangle, draw=mincolor, thick, minimum size=0.55cm, font=\scriptsize},
    stochNode/.style={circle, fill=black, inner sep=0pt, minimum size=1mm},
    cls/.style={rounded corners=4pt, dotted, draw=mincolor, thick, inner sep=5pt},
    edge/.style={thick, ->, >=Stealth, draw=black!80},
    lab/.style={font=\scriptsize, inner sep=1pt, fill=white, fill opacity=0.85, text opacity=1},
    scale=0.95, transform shape
]
    \pgfdeclarelayer{background}
    \pgfsetlayers{background,main}
    \node[maxNode] (s0) at (0,0) {$s_0$};
    \node[stochNode] (j) at (1.1,0) {};
    \node[minNode] (t1) at (2.7, 1.3) {$t_1$};
    \node[minNode] (t2) at (2.7,-1.3) {$t_2$};
    \node[sinkNode] (z) at (5.4,0) {$z$};
    \begin{pgfonlayer}{background}
      \node[cls, fit=(t1)(t2)] (C) {};
    \end{pgfonlayer}
    \draw[edge] (-0.9,0) -- (s0);
    \draw[edge] (s0) -- node[lab, above] {$a$} (j);
    \draw[edge] (j) -- node[lab, above left, pos=0.55] {$\frac12$} (t1);
    \draw[edge] (j) -- node[lab, below left, pos=0.55] {$\frac12$} (t2);
    \draw[edge] (t1) to[bend left=28]  node[lab, above right, pos=0.5] {$l/0$} (z);
    \draw[edge] (t1) to[bend right=12] node[lab, below, pos=0.38] {$r/1$} (z);
    \draw[edge] (t2) to[bend left=12]  node[lab, above, pos=0.38] {$l/1$} (z);
    \draw[edge] (t2) to[bend right=28] node[lab, below right, pos=0.5] {$r/0$} (z);
    \node[font=\scriptsize, color=mincolor] at (2.7,-2.5) {$C_o$: same observation};
\end{tikzpicture}
\caption{$\sg$ with prescriptions}
\end{subfigure}
\hfill
\begin{subfigure}[t]{0.6\paperwidth}
\centering
\begin{tikzpicture}[
    maxNode/.style={circle, draw=maxcolor, thick, minimum size=0.55cm, font=\scriptsize},
    dNode/.style={circle, draw=maxcolor, dashed, thick, inner sep=0pt, minimum size=1.15cm, font=\tiny},
    sinkNode/.style={circle, draw=maxcolor, thick, double, double distance=0.6pt, minimum size=0.55cm, font=\scriptsize},
    minNode/.style={rectangle, draw=mincolor, thick, minimum height=0.55cm, minimum width=1.15cm, inner sep=1pt, font=\scriptsize},
    realNode/.style={minNode, fill=mincolor!22},
    stochNode/.style={circle, fill=black, inner sep=0pt, minimum size=1mm},
    band/.style={rounded corners=3pt, dotted, draw=mincolor, thick, fill=mincolor!5},
    gband/.style={rounded corners=3pt, dotted, draw=black!50, thick, fill=black!4},
    edge/.style={thick, ->, >=Stealth, draw=black!80},
    dedge/.style={thick, ->, >=Stealth, dashed, draw=black!60},
    lab/.style={font=\scriptsize, inner sep=1pt, fill=white, fill opacity=0.85, text opacity=1},
    scale=0.85, transform shape
]
    \pgfdeclarelayer{background}
    \pgfsetlayers{background,main}
    \begin{pgfonlayer}{background}
      \draw[band]  (1.85,-2.45) rectangle (4.15, 3.2);
      \draw[gband] (5.25,-2.45) rectangle (7.15, 3.2);
      \draw[band]  (8.25,-2.45) rectangle (10.55,3.2);
    \end{pgfonlayer}
    \node[font=\scriptsize, color=mincolor] at (3.0,3.5) {$\pmin$ observes $(o,1)$};
    \node[font=\scriptsize, color=black!70] at (6.2,3.5) {$\pmin$ observes $(o,1,\circ)$};
    \node[font=\scriptsize, color=mincolor] at (9.4,3.5) {$\pmin$ observes $(o,2)$};
    \node[maxNode] (s0) at (-0.9,0) {$s_0$};
    \node[stochNode] (j) at (0.5,0) {};
    \node[realNode] (x11) at (3.0, 1.6) {$x_{t_1,1}$};
    \node[dNode] (dl) at (6.2, 2.5) {$d_{t_1,l,1}$};
    \node[dNode] (dr) at (6.2, 0.7) {$d_{t_1,r,1}$};
    \node[minNode] (xl2) at (9.4, 2.5) {$x_{t_1,l,2}$};
    \node[minNode] (xr2) at (9.4, 0.7) {$x_{t_1,r,2}$};
    \node[minNode] (x21) at (3.0,-1.6) {$x_{t_2,1}$};
    \node[dNode] (d21) at (6.2,-1.6) {$d_{t_2,1}$};
    \node[realNode] (x22) at (9.4,-1.6) {$x_{t_2,2}$};
    \node[sinkNode] (z) at (12.6, 0.4) {$z$};
    \draw[edge] (s0) -- node[lab, above] {$a$} (j);
    \draw[edge] (j) -- node[lab, above left, pos=0.5] {$\frac12$} (x11);
    \draw[edge] (j) -- node[lab, below left, pos=0.5] {$\frac12$} (x21);
    \draw[edge] (x11) -- node[lab, above, pos=0.45] {$l$} (dl);
    \draw[edge] (x11) -- node[lab, below, pos=0.45] {$r$} (dr);
    \draw[dedge] (dl) -- node[lab, above] {$\bot$} (xl2);
    \draw[dedge] (dr) -- node[lab, above] {$\bot$} (xr2);
    \draw[edge] (x21) -- node[lab, above] {$*$} (d21);
    \draw[dedge] (d21) -- node[lab, above] {$\bot$} (x22);
    \draw[edge] (xl2) -- node[lab, above right, pos=0.35] {$*/0$} (z);
    \draw[edge] (xr2) -- node[lab, below right, pos=0.35] {$*/1$} (z);
    \draw[edge] (x22) to[bend right=18] node[lab, below right, pos=0.4] {$r/0$} (z);
    \draw[edge] (x22) to[bend left=18]  node[lab, above left, pos=0.4] {$l/1$} (z);
\end{tikzpicture}
\caption{The chain game $\sg'$ ($k=2$)}
\end{subfigure}
}

\medskip
\begin{subfigure}[t]{0.98\textwidth}
\centering
\begin{tikzpicture}[
    ev/.style={circle, inner sep=0pt, minimum size=2.2mm},
    tk/.style={circle, draw=black!45, fill=white, inner sep=0pt, minimum size=1.6mm},
    axis/.style={thick, ->, >=Stealth, draw=black!70},
    entry/.style={draw=maxcolor, thick, densely dashed},
    exit/.style={draw=mincolor, thick, densely dashed},
    lab/.style={font=\scriptsize},
    scale=0.95, transform shape
]
    \draw[axis] (-0.4,2.2) -- (11.7,2.2);
    \draw[axis] (-0.4,0)   -- (11.7,0);
    \node[lab, anchor=east] at (-0.55,2.2) {$\sg$};
    \node[lab, anchor=east] at (-0.55,0)   {$\sg'$};
    \foreach \i/\w in {0/1,1/\gamma,2/\gamma^2,3/\gamma^3}{
      \node[tk] at (\i*1.5+0.0,2.2) {};
    }
    \node[ev, fill=maxcolor] (a0) at (0,2.2) {};
    \node[ev, fill=mincolor] (a1) at (1.5,2.2) {};
    \node[ev, fill=maxcolor] (a2) at (3.0,2.2) {};
    \node[ev, fill=mincolor] (a3) at (4.5,2.2) {};
    \node[lab, above=3pt] at (a0) {$r_0$};
    \node[lab, above=3pt] at (a1) {$r_1$};
    \node[lab, above=3pt] at (a2) {$r_2$};
    \node[lab, above=3pt] at (a3) {$r_3$};
    \node[lab, below=3pt] at (a0) {$1$};
    \node[lab, below=3pt] at (a1) {$\gamma$};
    \node[lab, below=3pt] at (a2) {$\gamma^2$};
    \node[lab, below=3pt] at (a3) {$\gamma^3$};
    \foreach \j in {0,...,7}{
      \node[tk] at (\j*1.5,0) {};
    }
    \node[ev, fill=maxcolor] (b0) at (0,0)   {};
    \node[ev, fill=mincolor] (b3) at (4.5,0) {};
    \node[ev, fill=maxcolor] (b4) at (6.0,0) {};
    \node[ev, fill=mincolor] (b7) at (10.5,0) {};
    \node[lab, below=3pt] at (b0) {$1$};
    \node[lab, below=3pt] at (b3) {$\gamma'^{3}$};
    \node[lab, below=3pt] at (b4) {$\gamma'^{4}$};
    \node[lab, below=3pt] at (b7) {$\gamma'^{7}$};
    \foreach \j in {1,2,5,6}{
      \node[lab, below=3pt, color=black!50] at (\j*1.5,0) {$\gamma'^{\j}$};
    }
    \draw[entry] (a0) -- (b0);
    \draw[exit]  (a1) -- (b3);
    \draw[entry] (a2) -- (b4);
    \draw[exit]  (a3) -- (b7);
    \node[lab, anchor=west] at (9.0,1.6) {\textcolor{maxcolor}{$\bullet$} entry reward - reward of $\pmax$-transition};
    \node[lab, anchor=west] at (9.0,1.2) {\textcolor{mincolor}{$\bullet$} exit reward - reward of $\pmin$-transition};
\end{tikzpicture}
\caption{Position and discount weight of the rewards $r_0,\dots,r_3$ in $\sg$ and in $\sg'$.}
\end{subfigure}
\caption{Example~\ref{ex:chain}. (a) A game $\sg$ with two $\pmin$-states with the same observation. (b) Its chain compilation for $k=2$. Both players observe the same along the two chains. The shaded rectangles are the slots where the real choice of $\pmin$ is made, the choice at the other slot is ignored (or recorded in the state). (c) The $i$-th reward of $\sg$ is earned at the transition $2kj$ or $2kj+2k-1$ of $\sg'$ ($i=2j$ or $2j+1$).}
\label{fig:chain-example}
\end{figure*}
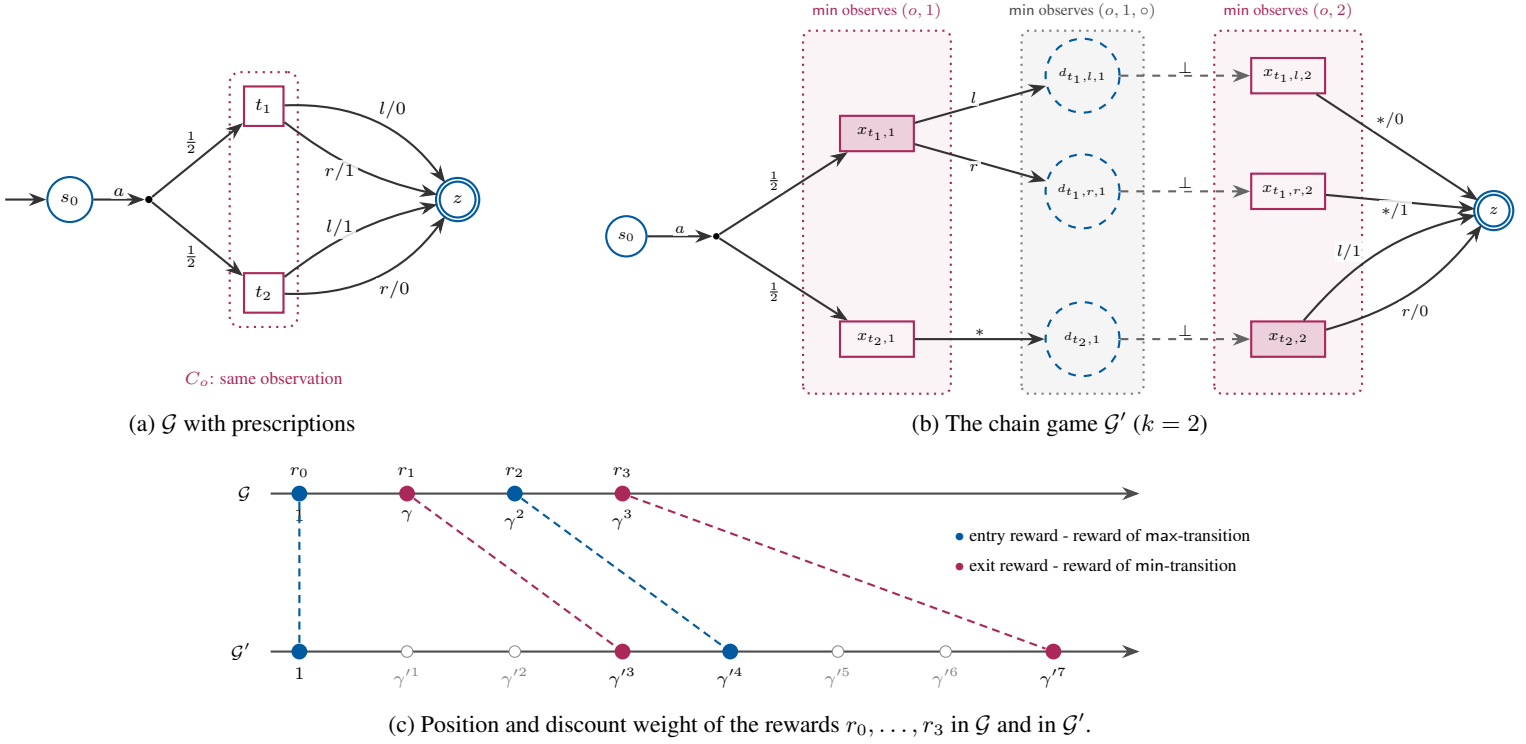
\begin{figure*}[t]
\centering
\begin{subfigure}[t]{0.50\textwidth}
\centering
\begin{tikzpicture}[
    maxNode/.style={circle, draw=maxcolor, thick, minimum size=0.55cm, font=\scriptsize},
    sinkNode/.style={circle, draw=maxcolor, thick, double, double distance=0.6pt, minimum size=0.55cm, font=\scriptsize},
    minNode/.style={rectangle, draw=mincolor, thick, minimum size=0.55cm, font=\scriptsize},
    stochNode/.style={circle, fill=black, inner sep=0pt, minimum size=1mm},
    cls/.style={rounded corners=4pt, dotted, draw=mincolor, thick, inner sep=5pt},
    edge/.style={thick, ->, >=Stealth, draw=black!80},
    lab/.style={font=\scriptsize, inner sep=0.8pt, fill=white, fill opacity=0.9, text opacity=1},
    scale=0.95, transform shape
]
    \pgfdeclarelayer{background}
    \pgfsetlayers{background,main}
    \node[maxNode] (s0) at (0,0) {$s_0$};
    \node[stochNode] (j) at (1.1,0) {};
    \node[minNode] (t1) at (2.7, 1.3) {$t_1$};
    \node[minNode] (t2) at (2.7,-1.3) {$t_2$};
    \node[sinkNode] (z) at (5.6,0) {$z$};
    \begin{pgfonlayer}{background}
      \node[cls, fit=(t1)(t2)] (C) {};
    \end{pgfonlayer}
    \draw[edge] (-0.9,0) -- (s0);
    \draw[edge] (s0) -- node[lab, above] {$a$} (j);
    \draw[edge] (j) -- node[lab, above left, pos=0.55] {$\frac12$} (t1);
    \draw[edge] (j) -- node[lab, below left, pos=0.55] {$\frac12$} (t2);
    \draw[edge] (t1) to[bend left=28]  node[lab, above right, pos=0.5] {$(l,*)/0$} (z);
    \draw[edge] (t1) to[bend right=12] node[lab, below, pos=0.36] {$(r,*)/1$} (z);
    \draw[edge] (t2) to[bend left=12]  node[lab, above, pos=0.36] {$(*,l)/1$} (z);
    \draw[edge] (t2) to[bend right=28] node[lab, below right, pos=0.5] {$(*,r)/0$} (z);
    \node[font=\scriptsize, color=mincolor] at (2.7,-2.5) {$C_o$: actions $\{l,r\}^2$};
\end{tikzpicture}
\caption{The game $\sg^\times$}
\end{subfigure}
\hfill
\begin{subfigure}[t]{0.47\textwidth}
\centering
\begin{tikzpicture}[
    cell/.style={minimum width=1.25cm, minimum height=0.8cm, draw=black!35, font=\scriptsize, inner sep=1pt},
    hd/.style={font=\scriptsize, inner sep=2pt},
    mg/.style={font=\scriptsize, inner sep=2pt, color=mincolor},
    scale=1.0, transform shape
]
    \node[hd] at (3.65,2.6) {$\beta_2$};
    \node[hd] at (3.0,2.15) {$l$}; \node[hd] at (4.3,2.15) {$r$};
    \node[hd] at (1.15,1.15) {$\beta_1$};
    \node[hd] at (2.1,1.55) {$l$}; \node[hd] at (2.1,0.75) {$r$};
    \node[cell] at (3.0,1.55) {$\tfrac14\cdot\tfrac12=\tfrac18$};
    \node[cell] at (4.3,1.55) {$\tfrac14\cdot\tfrac12=\tfrac18$};
    \node[cell] at (3.0,0.75) {$\tfrac34\cdot\tfrac12=\tfrac38$};
    \node[cell] at (4.3,0.75) {$\tfrac34\cdot\tfrac12=\tfrac38$};
    \node[mg] at (5.4,1.55) {$\tfrac14$}; \node[mg] at (5.4,0.75) {$\tfrac34$};
    \node[mg] at (3.0,0.18) {$\tfrac12$}; \node[mg] at (4.3,0.18) {$\tfrac12$};
    \node[mg, anchor=west] at (5.75,1.15) {$=\pi(t_1)$};
    \node[mg] at (3.65,-0.2) {$=\pi(t_2)$};
    \node[font=\scriptsize, anchor=south, align=center] at (3.65,2.95) {$\pi'=\pi(t_1)\otimes\pi(t_2)$};
\end{tikzpicture}
\caption{An ordinary strategy of $\sg^\times$}
\end{subfigure}

\medskip
\begin{subfigure}[t]{0.98\textwidth}
\centering
\begin{tikzpicture}[
    ev/.style={circle, inner sep=0pt, minimum size=2.2mm},
    axis/.style={thick, ->, >=Stealth, draw=black!70},
    same/.style={draw=black!55, thick, densely dashed},
    lab/.style={font=\scriptsize},
    scale=0.95, transform shape
]
    \draw[axis] (-0.4,1.9) -- (6.4,1.9);
    \draw[axis] (-0.4,0)   -- (6.4,0);
    \node[lab, anchor=east] at (-0.55,1.9) {$\sg$};
    \node[lab, anchor=east] at (-0.55,0)   {$\sg^\times$};
    \foreach \i/\w/\c in {0/1/maxcolor,1/\gamma/mincolor,2/\gamma^2/maxcolor,3/\gamma^3/mincolor}{
      \node[ev, fill=\c] at (\i*1.5,1.9) {};
      \node[ev, fill=\c] at (\i*1.5,0)   {};
      \node[lab, above=3pt] at (\i*1.5,1.9) {$r_\i$};
      \node[lab, below=3pt] at (\i*1.5,0)   {$\w$};
      \draw[same] (\i*1.5,1.9) -- (\i*1.5,0);
    }
    \node[lab, anchor=west] at (7.0,1.2) {\textcolor{maxcolor}{$\bullet$} reward of a $\pmax$-transition};
    \node[lab, anchor=west] at (7.0,0.7) {\textcolor{mincolor}{$\bullet$} reward of a $\pmin$-transition};
\end{tikzpicture}
\caption{Every reward is earned at the same transition in $\sg$ and in $\sg^\times$, so the discount weights $\gamma^i$ agree.}
\end{subfigure}
\caption{Example~\ref{ex:vector}. (a) The game $\sg^\times$ for the game $\sg$ of Figure~\ref{fig:chain-example}(a): both states of $C_o$ have the four actions $(\beta_1,\beta_2)\in\{l,r\}^2$, and $t_i$ uses only $\beta_i$ (an edge labelled $(l,*)$ stands for the two tuples with $\beta_1=l$). (b) The ordinary strategy $\pi'$ that plays the product of $\pi(t_1)=(\frac14,\frac34)$ and $\pi(t_2)=(\frac12,\frac12)$ on tuples; its marginals are the prescription $\pi$. (c) The timing is unchanged.}
\label{fig:vector-example}
\end{figure*}
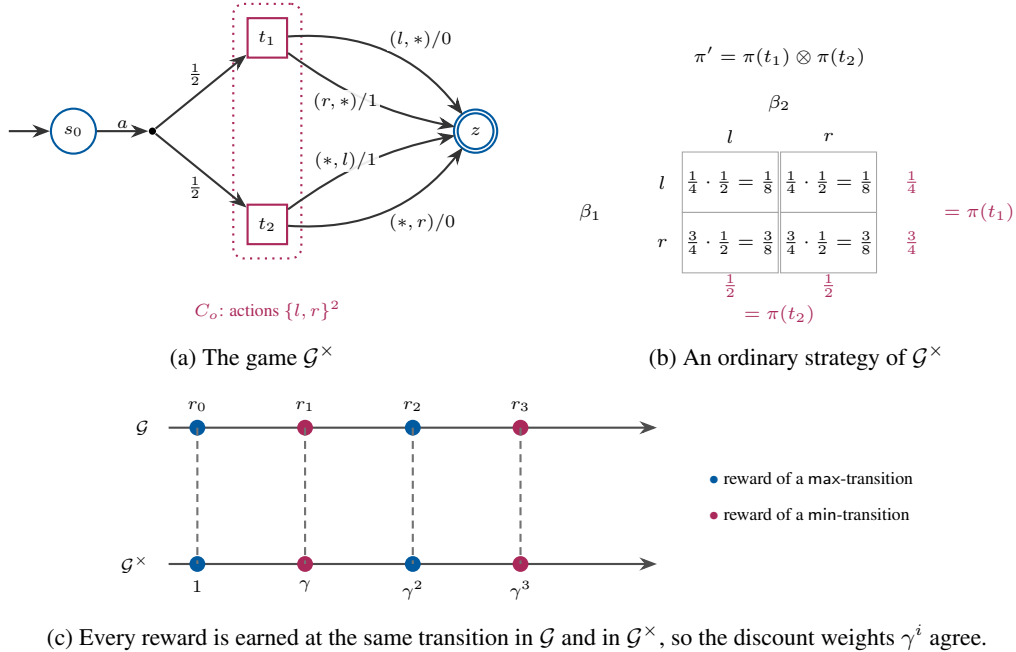

\begin{example}[Chain compilation, and why it does not preserve $\Disc_\gamma$]
\label{ex:chain}
Let $\sg$ be the game of Figure~\ref{fig:chain-example}(a). From $s_0$, the game moves to $t_1$ or $t_2$ with probability $\tfrac12$ each. The two $\pmin$-states have the same observation $o$, so they form one class $C_o$ with $k=2$. In $t_1$ the action $r$ earns $1$ and $l$ earns $0$, and in $t_2$ the rewards are the other way round. The play then stays in the sink $z$.
\medskip

\emph{Why prescriptions matter.} With a prescription, $\pmin$ chooses an action for each state of $C_o$ separately. The prescription $(r,l)$, which plays $r$ in $t_1$ and $l$ in $t_2$, earns $1$ with probability $1$. An ordinary strategy cannot do this: the histories at $t_1$ and $t_2$ look the same, so it plays $r$ with the same probability $p$ in both states, and it earns $1$ with probability $\tfrac12p+\tfrac12(1-p)=\tfrac12$ for every $p$. In general, a prescription with $\pi(t_1)(r)=p_1$ and $\pi(t_2)(r)=p_2$ earns $1$ with probability $\tfrac12p_1+\tfrac12(1-p_2)$, which takes every value in $[0,1]$. The compilation must therefore let an ordinary strategy of the new game reproduce all such prescriptions.

\medskip

\emph{The chain.} Figure~\ref{fig:chain-example}(b) shows $\sg'$. From $t_1$ and from $t_2$, the game continues in a chain of $2k-1=3$ transitions, in which $\pmin$ moves at two slots. In the chain of $t_1$, the first slot is the real choice, so $\pmin$ plays the real action $b\in\{l,r\}$ there, and the state $d_{t_1,b,1}$ records it; the second slot is ignored. In the chain of $t_2$, the second slot is the real choice and the first is ignored. Both chains look the same to $\pmin$: it observes $(o,1)$, then $(o,1,\circledcirc)$, then $(o,2)$, whichever state was visited. An ordinary strategy of $\sg'$ is therefore a pair of mixed actions, one for each slot. It realizes the prescription that uses the first mixed action in $t_1$ and the second in $t_2$. For instance, playing $r$ at the first slot and $l$ at the second realizes $(r,l)$.

\medskip

\emph{Strategies of $\sg'$ and prescriptions coincide.} Let $w$ be a $\pmin$-history of $\sg$ that ends in the observation $o$. When the play enters the chain, the histories of $\pmin$ in $\sg'$ continue as $w\cdot(o,1)\cdot(o,1,\circledcirc)\cdot(o,2)$, whichever of $t_1,t_2$ was visited, because the observations along the two chains are identical. An ordinary strategy $\pi'$ of $\sg'$ therefore fixes, after $w$, one mixed action $\mu_1$ for the first slot and one mixed action $\mu_2$ for the second slot, and the second cannot depend on the action played at the first, since strategies only see observations. In the chain of $t_i$, the action of the $i$-th slot is the real action of $\pmin$ in $t_i$ and the other slot is ignored. Hence $\pi'$ induces the prescription $\pi(w)(t_i)=\mu_i$ for $i=1,2$. Conversely, given a prescription $\pi$, play $\mu_i=\pi(w)(t_i)$ at the $i$-th slot. In the chain of $t_i$, the real action is then drawn from $\pi(w)(t_i)$ at the $i$-th slot, exactly as in $\sg$. So every prescription arises from some ordinary strategy of $\sg'$ (for general $k$, from $k$ mixed actions, one per slot).

\medskip
\emph{The laws agree.} The objectives of corresponding plays of $\sg'$ and $\sg$ are related by $X'=c\cdot X$ with $c$ as in Lemma~\ref{lem:compile}, since both plays have the same rewards in the same order, and the strategies correspond as above. Hence the sets of laws of $\sg'$ and $\sg$ correspond under $\push{c}$.

\medskip

\emph{Discounted returns.} For $\Tot_{N'}$, $\Tot$, and $\LimAvg$, only the order of the rewards matters, so the identity $X'=c\cdot X$ on corresponding plays holds. For discounted returns, the position of a reward matters. A round of $\sg$ takes $2$ transitions, while a round of $\sg'$ takes $2k=4$, and Figure~\ref{fig:chain-example}(c) shows where the first four rewards of a play land: $r_0,r_1,r_2,r_3$ are earned at the transitions $0,3,4,7$ of $\sg'$ instead of $0,1,2,3$. With unchanged rewards, the discounted return of a play of $\sg'$ is $r_0+\gamma'^3r_1+\gamma'^4r_2+\gamma'^7r_3+\cdots$, while that of the corresponding play of $\sg$ is $r_0+\gamma r_1+\gamma^2r_2+\gamma^3r_3+\cdots$. These cannot be proportional for all reward sequences: comparing the weights of $r_0$ forces the constant to be $1$, and then $r_1$ and $r_2$ would need $\gamma'^3=\gamma$ and $\gamma'^4=\gamma^2$ simultaneously, which is impossible for $\gamma<1$. So no discount factor $\gamma'$ and no constant $c$ make the chain gadget preserve $\Disc_\gamma$ with the same rewards.

\end{example}

\begin{example}[Vector actions preserve $\Disc_\gamma$]
\label{ex:vector}
Take again the game $\sg$ of Figure~\ref{fig:chain-example}(a). The game $\sg^\times$ of Proposition~\ref{prop:compile-disc} has the same states, transitions and rewards as $\sg$, but both states of $C_o$ have the four actions $\beta=(\beta_1,\beta_2)\in\{l,r\}^2$ (Figure~\ref{fig:vector-example}(a)). The state $t_i$ plays $\beta_i$ and ignores the other component. For the prescription $\pi$ with $\pi(t_1)=(\tfrac14,\tfrac34)$ and $\pi(t_2)=(\tfrac12,\tfrac12)$, the ordinary strategy $\pi^\times$ plays the product distribution on tuples (Figure~\ref{fig:vector-example}(b)), and its marginals are $\pi(t_1)$ and $\pi(t_2)$.

\emph{Same laws in $\sg$ and $\sg^\times$.} A play of $\sg$ has two steps before it reaches the sink $z$, where all further rewards are $0$. Counting steps from $0$, step $0$ is the move $s_0\to t_i$, and step $1$ is the move of $\pmin$ from $t_i$ to $z$. The only nonzero reward is the reward $1$ of that second step, so $\Disc_\gamma$ takes only the values $0$ and $\gamma^1=\gamma$, and the law of $\Disc_\gamma$ is determined by the probability that $\pmin$ earns $1$.

In $\sg$ under the prescription $\pi$, with $\pi(t_1)=(\tfrac14,\tfrac34)$ and $\pi(t_2)=(\tfrac12,\tfrac12)$, $\pmin$ earns $1$ in $t_1$ by playing $r$ (probability $\tfrac34$) and in $t_2$ by playing $l$ (probability $\tfrac12$). Each state is reached with probability $\tfrac12$, so $\pmin$ earns $1$ with probability $\tfrac12\cdot\tfrac34+\tfrac12\cdot\tfrac12=\tfrac58$.

In $\sg^\times$ under the product strategy $\pi^\times=\pi(t_1)\otimes\pi(t_2)$, the tuples $(l,l),(l,r),(r,l),(r,r)$ have probabilities $\tfrac18,\tfrac18,\tfrac38,\tfrac38$. The state $t_1$ uses the first component, which is $r$ with probability $\tfrac38+\tfrac38=\tfrac34$, and $t_2$ uses the second, which is $l$ with probability $\tfrac18+\tfrac38=\tfrac12$. So $\pmin$ earns $1$ with probability $\tfrac58$ as well.

In both games, the earning happens in step $1$, so the timing is unchanged, and $\Disc_\gamma$ has the law $\tfrac38\,\dirac{0}+\tfrac58\,\dirac{\gamma}$ for every $\gamma\in(0,1)$, with no rescaling (Figure~\ref{fig:vector-example}(c)).
\end{example}

The price of $\sg^\times$ is its action sets, whose size is $\prod_i|A(t_i)|$ for a class of $k$ states. The chain gadget above has polynomial size, but it changes the timing and therefore preserves $\Tot_N$, $\Tot$ and $\LimAvg$ only. Both constructions show that ordinary ac-POSGs and ac-POSGs with prescriptions are equally expressive for all four returns.
\section{Proofs for Section~\ref{sec:equiv}}
\label{app:equiv}

Throughout, the set of plays of a model carries the $\sigma$-algebra generated by the cylinder sets $\Cyl(u)$ of finite prefixes~$u$.
Let $\mathcal{C}$ be the family of cylinder sets, that is, sets of plays with a prescribed finite prefix. The intersection of two cylinder sets is again a cylinder set or empty, so $\mathcal{C}$ is closed under finite intersections (a $\pi$-system). It generates the $\sigma$-algebra on plays. By the uniqueness theorem for probability measures (a consequence of Dynkin's $\pi$-$\lambda$ theorem), two probability measures that agree on $\mathcal{C}$ are therefore equal.
For a measurable map $f$ and a measure $\Prob$, we write $f_\#\Prob=\Prob\circ f^{-1}$. Strategies only matter on histories that can occur and at which the player moves; we identify strategies that differ only on other histories, and we call an entry of an assignment \emph{used} if it can influence the play (for the environment of an RPOMDP, this is the entry $(s_t,a_t)$ of the assignment chosen in step~$t$).

\subsection{Proof of Lemma~\ref{lem:fwd}} \label{app:lemfwd}

Let $\sg=\fwd{\rmdp}$. Every play of $\sg$ has the form
\[
\rho'=s_0\,a_0\,(s_0,a_0)\,v_0\,s_1\,a_1\,(s_1,a_1)\,v_1\,s_2\cdots
\qquad(a_i\in A(s_i),\ v_i\in V_{s_i,a_i}),
\]
and $\proj(\rho')=s_0a_0s_1a_1\cdots$ denotes the play of $\rmdp$ obtained by erasing the $\pmin$-states and the vertex choices.

\paragraph{Histories.} Let $\proj$ be the map that erases the $\pmin$-states and the vertex choices, defined on prefixes and on plays of $\sg$, so that it sends a play $\rho'$ of $\sg$ to the play $\proj(\rho')=s_0a_0s_1a_1\cdots$ of $\rmdp$. For a prefix $u=s_0a_0\cdots s_t$ of $\rmdp$, let
\[
h^{\mathsf a}(u)=\Za(s_0)\cdots\Za(s_t)\quad\text{and}\quad h^{\mathsf e}(u)=\Ze(s_0)\cdots\Ze(s_t).
\]
For a word $w=o_0\cdots o_t$, let $\eta(w)=o_0o_0o_1o_1\cdots o_to_t$ be the word in which every letter is doubled, and let $\eta^-(w)=o_0o_0\cdots o_{t-1}o_{t-1}o_t$ be the word in which every letter except the last one is doubled. The agent observes $\Za(s_i)$ both in $s_i$ and in $(s_i,a_i)$, and the environment observes $\Ze(s_i)$ at both, so for every prefix $u'$ of $\sg$ with $\proj(u')=u$ the histories are as follows. If $u'$ ends in the $\pmax$-state $s_t$, the $\pmax$-history is $\eta^-(h^{\mathsf a}(u))$. If $u'$ is extended to the $\pmin$-state $(s_t,a_t)$, the $\pmin$-history is $\eta(h^{\mathsf e}(u))$. The maps $\eta$ and $\eta^-$ are injective, and neither history depends on the vertex choices.

\paragraph{Strategy maps.} For $\sigma\in\Sigma_\rmdp$, let $\sigma^\sg(\eta^-(h))=\sigma(h)$; at all other observation histories, $\pmax$ does not move. This is a bijection between agent strategies and $\pmax$-strategies. For an observation history $h$, write $\mathrm{last}(h)$ for its last observation. An environment strategy $\pi$ assigns to every history $h$ and every pair $(s,a)$ with $\Ze(s)=\mathrm{last}(h)$ and $a\in A(s)$ a distribution $\pi(h)(s,a)\in\Unc(s,a)$.

For a $\pmin$-strategy $\pi'$, let $\Gamma(\pi')$ be the environment strategy with
\[
\Gamma(\pi')(h)(s,a)=\sum_{v\in V_{s,a}}\pi'(\eta(h))((s,a))(v)\cdot v
\]
for all $h$ and all $(s,a)$ with $\Ze(s)=\mathrm{last}(h)$ and $a\in A(s)$. This is a convex combination of the vertices of $\Unc(s,a)$, so $\Gamma(\pi')(h)(s,a)\in\Unc(s,a)$.

Conversely, since $\Unc(s,a)=\conv(V_{s,a})$, we fix for every pair $(s,a)$ and every $p\in\Unc(s,a)$ a distribution $\mu_{s,a}(p)$ over $V_{s,a}$ with \\ $\sum_{v}\mu_{s,a}(p)(v)\cdot v=p$. Given an environment strategy $\pi$, we define the $\pmin$-strategy $\pi'$ by
\[
\pi'(\eta(h))((s,a))=\mu_{s,a}\bigl(\pi(h)(s,a)\bigr)
\]
for all $h$ and all $(s,a)$ with $\Ze(s)=\mathrm{last}(h)$. This is well defined because $\eta$ is injective and the $\pmin$-history at $(s,a)$ is $\eta(h^{\mathsf e}(u))$, whose last letter is $\Ze(s)$. Then, for all such $h$, $s$, and $a$,
\[
\Gamma(\pi')(h)(s,a)=\sum_{v\in V_{s,a}}\mu_{s,a}\bigl(\pi(h)(s,a)\bigr)(v)\cdot v=\pi(h)(s,a),
\]
so $\Gamma(\pi')=\pi$.

\paragraph{Projection of measures} Let $\proj_\#\mu(B)=\mu(\proj^{-1}(B))$, where $\proj^{-1}(B)=\{\rho'\mid\proj(\rho')\in B\}$ is the set of plays of $\sg$ that project into $B$, for a measure $\mu$ on plays of $\sg$ and a measurable set $B$ of plays of $\rmdp$.  We show $\proj_\#\Prob^{\sigma^\sg,\pi'}_\sg=\Prob^{\sigma,\Gamma(\pi')}_\rmdp$ for all $\sigma$ and $\pi'$. Cylinder sets form a $\pi$-system that generates the $\sigma$-algebra on plays, so two probability measures that agree on all cylinder sets are equal. It therefore suffices to show $\Prob^{\sigma,\Gamma(\pi')}_\rmdp(\Cyl(u))=\Prob^{\sigma^\sg,\pi'}_\sg(\proj^{-1}(\Cyl(u)))$ for every prefix $u$ of $\rmdp$, which we prove by induction on the length of $u=s_0a_0\cdots s_t$.
For $u=s_0$, both sides are~$1$. For the induction step, consider $u\,a\,s'$. The set $\proj^{-1}(\Cyl(u\,a\,s'))$ is the disjoint union of the cylinders $\Cyl(u'\,a\,(s_t,a)\,v\,s')$ over all prefixes $u'$ of $\sg$ with $\proj(u')=u$ and all $v\in V_{s_t,a}$, and
\[
\Prob_\sg\bigl(\Cyl(u'\,a\,(s_t,a)\,v\,s')\bigr)=\Prob_\sg(\Cyl(u'))\cdot\sigma(h^{\mathsf a}(u))(a)
\cdot\pi'(\eta(h^{\mathsf e}(u)))((s_t,a))(v)\cdot v(s')
\]
Summing over $v$ turns the last two factors into $p(s')$ for the point
\[
p=\Gamma(\pi')(h^{\mathsf e}(u))(s_t,a)\in\Unc(s_t,a),
\]
and summing over $u'$ replaces $\Prob_\sg(\Cyl(u'))$ by $\Prob_\sg(\proj^{-1}(\Cyl(u)))$, which equals $\Prob_\rmdp(\Cyl(u))$ by the induction hypothesis. The result is exactly $\Prob_\rmdp(\Cyl(u\,a\,s'))$.

\paragraph{Returns.} Let $(r_i)_{i\ge0}$ be the reward sequence of $\proj(\rho')$. The reward sequence of $\rho'$ is $(0,r_0,0,r_1,\dots)$, because $\pmax$-transitions earn $0$ and the $\pmin$-transition $((s_i,a_i),v_i,s_{i+1})$ earns $r(s_i,a_i,s_{i+1})$. Hence, on every play,
\begin{align*}
\Tot_{2N}(\rho')&=\Tot_N(\proj(\rho')),\qquad\Tot(\rho')=\Tot(\proj(\rho')),\\
\Disc_\beta(\rho')&=\textstyle\sum_i\beta^{2i+1}r_i=\beta\,\Disc_{\beta^2}(\proj(\rho')).
\end{align*}
For the average, let $\rho'$ be a play of $\sg$, let $\rho=\proj(\rho')$ with rewards $r_0,r_1,\dots$, and let $R=\max|r|$. Write $A_m=\frac1m\sum_{i<m}r_i$ and let $B_n$ be the average of the first $n$ rewards of $\rho'$. Since the rewards of $\rho'$ are $0,r_0,0,r_1,\dots$, the first $2m$ rewards sum to $\sum_{i<m}r_i=mA_m$, and the $(2m+1)$-st reward is $0$. Hence
\[
B_{2m}=\frac{mA_m}{2m}=\frac12A_m
\quad\text{and}\quad
B_{2m+1}=\frac{mA_m}{2m+1}=\frac{2m}{2m+1}B_{2m}.
\]
As $|A_m|\le R$, we get $|B_{2m+1}-B_{2m}|=\frac{|B_{2m}|}{2m+1}\le\frac{R}{2(2m+1)}\to0$.

It follows that the sequences $(B_n)_n$ and $(B_{2m})_m$ have the same limit points. Indeed, every subsequence of $(B_n)_n$ converging to some $L$ contains infinitely many even indices or infinitely many odd indices; replacing each odd index $2m+1$ by $2m$ changes the values by a vanishing amount, so $L$ is also a limit point of $(B_{2m})_m$, and the converse inclusion is trivial. Therefore
\[
\liminf_nB_n=\liminf_mB_{2m}=\tfrac12\liminf_mA_m
\]
and likewise for $\limsup$. Thus $X^\sg(\rho')=c\cdot X(\proj(\rho'))$ for $c=\frac12$, whichever of $\liminf$ or $\limsup$ defines $\LimAvg$. For total and discounted return the same identity $X^\sg=c\cdot(X\circ\proj)$ holds with the $c$ of Lemma~\ref{lem:fwd}, by the corresponding computation.

Since $X^\sg=c\cdot(X\circ\proj)$ holds on every play of $\sg$, for every measurable set $B$ of reals we have
\begin{align*}
\law^{\sigma^\sg,\pi'}_{X^\sg}(B)
&=\Prob_\sg\bigl(\{\rho' : c\cdot X(\proj(\rho'))\in B\}\bigr)\\
&=\Prob_\sg\bigl(\proj^{-1}\bigl(X^{-1}(S_c^{-1}(B))\bigr)\bigr)\\
&=\bigl(\proj_\#\Prob_\sg\bigr)\bigl(X^{-1}(S_c^{-1}(B))\bigr)\\
&=\Prob_\rmdp\bigl(X^{-1}(S_c^{-1}(B))\bigr)\\
&=\law^{\sigma,\Gamma(\pi')}_X\bigl(S_c^{-1}(B)\bigr)
=\push{c}\,\law^{\sigma,\Gamma(\pi')}_X(B),
\end{align*}
where $S_c^{-1}(B)=\{x : cx\in B\}$, the fourth equality is the projection of measures shown above (with $\sigma^\sg$ and $\Gamma(\pi')$ in the roles of the strategies), and the last equality is the definition of $\push{c}$. Hence $\law^{\sigma^\sg,\pi'}_{X^\sg}=\push{c}\,\law^{\sigma,\Gamma(\pi')}_X$.
Since every environment strategy is of the form $\Gamma(\pi')$, the two sets of laws in the lemma coincide. \qed

\begin{corollary}[RPOMDPs and ordinary POSGs]
\label{cor:ordinary}
Let $X\in\{\Tot_N,\Tot,\LimAvg\}$. For every RPOMDP $\rmdp$ there is an ordinary ac-POSG $\sg'$, of polynomial size, and a return $X'$ of $\sg'$, such that $\Val_{\sg'}(\Psi,X')=\Val_\rmdp(\Psi\circ\push{c},X)$ for all law functionals $\Psi$, where $\sg'=(\fwd{\rmdp})'$ and $c=c_1c_2$ with $c_1=1$ (for $\Tot_N,\Tot$) or $\tfrac12$ (for $\LimAvg$) from Lemma~\ref{lem:fwd}, and $c_2=1,1,1/k$ from Lemma~\ref{lem:compile}. Here $k$ is the maximal number of states $(s,a)$ with $\Ze(s)$ equal.
\end{corollary}
\begin{proof}
Compose Lemmas~\ref{lem:fwd} and~\ref{lem:compile}, using $\push{c_2}\circ\push{c_1}=\push{c_1c_2}$ and the fact that $\pmin$-observations in $\fwd{\rmdp}$ depend on $\Ze(s)$ only.
\end{proof}

\begin{corollary}[RPOMDPs and ordinary POSGs, discounted]
For every RPOMDP $\rmdp$ and rational $\beta\in(0,1)$ there is an ordinary ac-POSG $\sg'=(\fwd{\rmdp})^\times$ with $\Val_{\sg'}(\Psi,\Disc_\beta)=\Val_\rmdp(\Psi\circ\push{\beta},\Disc_{\beta^2})$ for all law functionals $\Psi$.
\end{corollary}
\begin{proof}
Combine Lemma~\ref{lem:fwd} (with $X=\Disc_{\beta^2}$, $X^\sg=\Disc_\beta$ and $c=\beta$) with Proposition~\ref{prop:compile-disc}, which introduces no further scaling.
\end{proof}

\subsection{Proof of Lemma~\ref{lem:bwd}} \label{app:lembwd}

Let $\rmdp=\bwd{\sg}$. A play of the ac-POSG $\sg$ has the form $\rho=s_0\,a_0\,t_0\,b_0\,s_1\,a_1\,t_1\,b_1\cdots$ with $s_i\in\Smax$ and $t_i\in\Smin$. Let
\[
\Lambda(\rho)=(s_0,\anew)\,a_0\,d_{t_0}\,\anew\,(s_1,b_0)\,a_1\,d_{t_1}\,\anew\,(s_2,b_1)\cdots.
\]
Every play of $\rmdp$ has this form, because tagged states lead to states $d_t$ and vice versa, and $\Lambda$ is a bijection whose inverse reads $b_i$ off the tag of the state $(s_{i+1},b_i)$. By the definition of the rewards of $\rmdp$, the transition $(s_i,b_{i-1})\xrightarrow{a_i} d_{t_i}$ earns $r(s_i,a_i,t_i)$ and the transition $d_{t_i}\to(s_{i+1},b_i)$ earns $r(t_i,b_i,s_{i+1})$, so $\rho$ and $\Lambda(\rho)$ have identical reward sequences.

\paragraph{Observation  Histories and strategies.} At $s_k$, the $\pmax$-history is
\[
\Zmax(s_0)\,\Zmax(t_0)\,\Zmax(s_1)\cdots\Zmax(s_k),
\]
while the agent history at $(s_k,b_{k-1})$ is
\[
(\Zmax(s_0),0)\,(\Zmax(t_0),1)\cdots(\Zmax(s_k),0).
\]
Attaching the alternating bits $0,1,0,1,\dots$ is a bijection $\vartheta$ between $\pmax$ observation  histories and the agent observation  histories that can occur. The agent observation histories that end in an observation with bit~$1$ are exactly those at states $d_t$, where the only action is $\anew$. Hence $\sigma^\rmdp(\vartheta(h))=\sigma(h)$ for observation histories $h$ of odd length, and $\sigma^\rmdp=\dirac{\anew}$ at observation  histories that end with bit~$1$, defines a bijection between $\pmax$-strategies and agent strategies. The environment of $\rmdp$ observes exactly the sequence $\Zmin(s_0)\Zmin(t_0)\Zmin(s_1)\cdots$ that $\pmin$ observes in $\sg$, so in $d_{t_k}$ its observation history coincides with the $\pmin$-history in $t_k$.

For a $\pmin$-strategy $\pi$, let $\Omega(\pi)$ be the environment strategy whose assignment after observation-history $h$ has the entries
\[
\Omega(\pi)(h)(d_t,\anew)=\sum_{b\in A(t)}\pi(h)(t)(b)\cdot\delta'_{t,b}\qquad(t\in\Smin);
\]

the entries at tagged states are the unique elements of the singleton uncertainty sets.
Conversely, for an environment strategy $\pi''$ of $\rmdp$, let
\[
\Xi(\pi'')(h)(t)(b)=\sum_{s'}\pi''(h)(d_t,\anew)((s',b))
\]
be the weight that the point chosen at $d_t$ puts on states with tag $b$. Recall that $\delta'_{t,b}((s',b'))=\delta(t,b)(s')$ if $b'=b$ and $0$ otherwise, and that the uncertainty set at $(d_t,\anew)$ is $\conv\{\delta'_{t,b}:b\in A(t)\}$. The supports of the $\delta'_{t,b}$ lie in the pairwise disjoint sets $\{(s',b):s'\}$, which is the property used below.

\emph{Claim.} Let $p=\sum_{b'}\lambda_{b'}\delta'_{t,b'}$ with $\lambda$ a distribution over $A(t)$. Then $\sum_{s'}p((s',b))=\lambda_b$ for every $b\in A(t)$.

Indeed, $p((s',b))=\lambda_b\,\delta(t,b)(s')$, because for $b'\ne b$ the distribution $\delta'_{t,b'}$ gives $(s',b)$ weight $0$. Summing over $s'$ and using $\sum_{s'}\delta(t,b)(s')=1$ gives $\lambda_b$.

\emph{$\Omega\circ\Xi=\mathrm{id}$.} Let $p=\pi''(h)(d_t,\anew)$. It lies in the convex hull, so $p=\sum_b\lambda_b\delta'_{t,b}$ for some distribution $\lambda$ over $A(t)$. By the claim, $\Xi(\pi'')(h)(t)(b)=\lambda_b$. In particular, $\Xi(\pi'')(h)(t)$ is a distribution over $A(t)$, and since the environment history in $d_t$ coincides with the $\pmin$-history in $t$, $\Xi(\pi'')$ is a $\pmin$-strategy. Moreover,
\[
\Omega(\Xi(\pi''))(h)(d_t,\anew)=\sum_b\Xi(\pi'')(h)(t)(b)\cdot\delta'_{t,b}=\sum_b\lambda_b\,\delta'_{t,b}=p=\pi''(h)(d_t,\anew).
\]
The entries at tagged states are the unique elements of singleton sets, so they agree. Hence $\Omega(\Xi(\pi''))=\pi''$.

\emph{$\Xi\circ\Omega=\mathrm{id}$.} Let $\pi$ be a $\pmin$-strategy. The point $\Omega(\pi)(h)(d_t,\anew)=\sum_{b'}\pi(h)(t)(b')\,\delta'_{t,b'}$ lies in the uncertainty set, so $\Omega(\pi)$ is an environment strategy. By the claim with $\lambda=\pi(h)(t)$,
\[
\Xi(\Omega(\pi))(h)(t)(b)=\sum_{s'}\Omega(\pi)(h)(d_t,\anew)((s',b))=\pi(h)(t)(b)
\]
for all $h$, $t$, and $b$. Hence $\Xi(\Omega(\pi))=\pi$.

\paragraph{Measures.} We show $\Prob^{\sigma^\rmdp,\Omega(\pi)}_\rmdp(\Cyl(\Lambda(u)))=\Prob^{\sigma,\pi}_\sg(\Cyl(u))$ for every prefix $u$ of $\sg$, by induction on the length of $u$. The base case is trivial. If $u$ ends in a $\pmax$-state $s_k$ and is extended by $a$ and $t$, the factor in $\sg$ is $\sigma(h)(a)\,\delta(s_k,a)(t)$, and in $\rmdp$ it is $\sigma^\rmdp(\vartheta(h))(a)\,\hat\delta(s_k,a)(d_t)$, which is the same.
If $u$ ends in a $\pmin$-state $t$ and is extended by $b$ and $s'$, the factor in $\sg$ is $\pi(h)(t)(b)\,\delta(t,b)(s')$. In $\rmdp$, the play passes through $d_t$, where the agent plays $\anew$ with probability~$1$ (its only action, as $\sigma^\rmdp=\dirac{\anew}$ at histories ending with bit~$1$), and the environment then moves to the tagged state $(s',b)$ with probability
\begin{align*}
\Omega(\pi)(h)(d_t,\anew)((s',b))&=\sum_{b'}\pi(h)(t)(b')\,\delta'_{t,b'}((s',b))\\
&=\pi(h)(t)(b)\,\delta(t,b)(s'),
\end{align*}
since $\delta'_{t,b'}((s',b))=0$ for $b'\ne b$. The factors agree.

Thus the identity holds in both cases, which completes the induction. Extend $\Lambda$ to prefixes in the obvious way. Every prefix of $\rmdp$ is $\Lambda(u)$ for a unique prefix $u$ of $\sg$, since $\Lambda$ is a bijection on plays whose inverse reads $b_i$ off the tag of $(s_{i+1},b_i)$, and $\Lambda(\Cyl(u))=\Cyl(\Lambda(u))$. The induction therefore gives, for every cylinder $\Cyl(v)$ of $\rmdp$,
\[
\Lambda_\#\Prob^{\sigma,\pi}_\sg(\Cyl(v))=\Prob^{\sigma,\pi}_\sg\bigl(\Lambda^{-1}(\Cyl(v))\bigr)=\Prob^{\sigma,\pi}_\sg(\Cyl(\Lambda^{-1}v))=\Prob^{\sigma^\rmdp,\Omega(\pi)}_\rmdp(\Cyl(v)).
\]
Cylinder sets form a $\pi$-system that generates the $\sigma$-algebra on plays, so two probability measures that agree on all cylinder sets are equal. Hence $\Lambda_\#\Prob^{\sigma,\pi}_\sg=\Prob^{\sigma^\rmdp,\Omega(\pi)}_\rmdp$.

Every return $X$ is a function of the reward sequence, and $\Lambda$ preserves reward sequences, so $X\circ\Lambda=X$ and
\[
\law^{\sigma^\rmdp,\Omega(\pi)}_X(B)=\Prob^{\sigma^\rmdp,\Omega(\pi)}_\rmdp(X^{-1}(B))=\Prob^{\sigma,\pi}_\sg(\Lambda^{-1}X^{-1}(B))=\Prob^{\sigma,\pi}_\sg(X^{-1}(B))=\law^{\sigma,\pi}_X(B).
\]
So every law in $\sg$ is a law in $\rmdp$. Conversely, let $(\sigma',\pi'')$ be a pair of strategies of $\rmdp$. Then $\sigma'=\sigma^\rmdp$ for a $\pmax$-strategy $\sigma$ (the map $\sigma\mapsto\sigma^\rmdp$ is a bijection), and $\pi''=\Omega(\Xi(\pi''))$, so the pair gives the same law as $(\sigma,\Xi(\pi''))$ in $\sg$. Hence the two sets of laws coincide. \qed

\subsection{Proof of Theorem~\ref{thm:umbrella}} \label{app:thmum}
\emph{Notation.} For a model, a strategy $\sigma$ of the maximizing player, and a
strategy $\pi$ of the opponent, let $\law^{\sigma,\pi}_X$ be the law of the return $X$.
Let $\lawset(\sigma)=\{\law^{\sigma,\pi}_X:\pi\text{ a strategy of the opponent}\}$.
This set is nonempty, and by definition of the robust value,
\begin{equation}
\label{eq:val-lawset-proof}
\Val(\Psi,X)=\sup_\sigma\,\inf_\pi\Psi(\law^{\sigma,\pi}_X)
=\sup_\sigma\,\inf_{\mu\in\lawset(\sigma)}\Psi(\mu),
\end{equation}
because the infimum over $\pi$ of a function of $\law^{\sigma,\pi}_X$ is the infimum of
that function over the set of all values $\law^{\sigma,\pi}_X$ takes. All suprema and
infima are taken in $\Rbar$, where they always exist.

\emph{A fact about images.} Let $f\colon A\to B$ be a map between sets and
$g\colon B\to\Rbar$ a function. Then
\begin{equation}
\label{eq:image}
\inf_{y\in f(A)}g(y)=\inf_{x\in A}g(f(x)).
\end{equation}
Indeed, every $g(f(x))$ with $x\in A$ is a value of $g$ on $f(A)$, which gives
``$\ge$''; and every $y\in f(A)$ is $f(x)$ for some $x\in A$, so
$g(y)=g(f(x))\ge\inf_{x'\in A}g(f(x'))$, which gives ``$\le$''. We apply it with
$f=\push{c}$, restricted to the set $A=\lawset_\rmdp(\sigma)$ of laws, and $g=\Psi$.
It uses nothing about $\Psi$.
\medskip

\emph{Part~(a).} Let $\sg=\fwd{\rmdp}$. By Lemma~\ref{lem:fwd}, the map
$\sigma\mapsto\sigma^\sg$ is a bijection from the strategies of the agent in $\rmdp$ onto
the strategies of $\pmax$ in $\sg$, and for every such $\sigma$,
\[
\lawset_\sg(\sigma^\sg)=\{\push{c}\nu:\ \nu\in\lawset_\rmdp(\sigma)\}.
\]
Here $\push{c}\nu$ is the law of $cY$ when $Y$ has law $\nu$. By~\eqref{eq:image} with
$A=\lawset_\rmdp(\sigma)$, $f=\push{c}$, and $g=\Psi$,
\[
\inf_{\mu\in\lawset_\sg(\sigma^\sg)}\Psi(\mu)
=\inf_{\nu\in\lawset_\rmdp(\sigma)}\Psi(\push{c}\nu)
=\inf_{\nu\in\lawset_\rmdp(\sigma)}(\Psi\circ\push{c})(\nu).
\]
The right-hand side is the inner term of~\eqref{eq:val-lawset-proof} for the law
functional $\Psi\circ\push{c}$ in $\rmdp$. Since $\sigma\mapsto\sigma^\sg$ is a
bijection, the supremum over all strategies $\sigma^\sg$ of $\pmax$ in $\sg$ equals the
supremum over all strategies $\sigma$ of the agent in $\rmdp$. Hence
\[
\Val_\sg(\Psi,X^\sg)=\sup_\sigma\inf_{\mu\in\lawset_\sg(\sigma^\sg)}\Psi(\mu)
=\sup_\sigma\inf_{\nu\in\lawset_\rmdp(\sigma)}(\Psi\circ\push{c})(\nu)
=\Val_\rmdp(\Psi\circ\push{c},X).
\]
\medskip

\emph{Part~(b).} Let $\rmdp=\bwd{\sg}$. By Lemma~\ref{lem:bwd}, $\sigma\mapsto\sigma^\rmdp$ is
a bijection from the strategies of $\pmax$ in $\sg$ onto the strategies of the agent in
$\rmdp$, and $\lawset_\rmdp(\sigma^\rmdp)=\lawset_\sg(\sigma)$, without rescaling. The
computation of part~(a) with $c=1$ (that is, with $\push{1}$ the identity) gives
$\Val_{\bwd{\sg}}(\Psi,X)=\Val_\sg(\Psi,X)$.

 \emph{Threshold problems.} Let $t$ be a rational and let $\bowtie\ \in\{\ge,>\}$.
 \begin{itemize}
 \item From ac-POSGs to RPOMDPs: map $(\sg,\Psi,X,t)$ to $(\bwd{\sg},\Psi,X,t)$. By~(b),
 $\Val_{\bwd{\sg}}(\Psi,X)\bowtie t$ if and only if $\Val_\sg(\Psi,X)\bowtie t$.
 \item From RPOMDPs to ac-POSGs: map $(\rmdp,\Psi,X,t)$ to
 $(\fwd{\rmdp},\Psi\circ\push{1/c},X^\sg,t)$. The functional $\Psi\circ\push{1/c}$ is a
 law functional, since $\push{1/c}$ maps laws to laws. Applying~(a) to it gives
 \[
 \Val_{\fwd{\rmdp}}(\Psi\circ\push{1/c},X^\sg)
 =\Val_\rmdp(\Psi\circ\push{1/c}\circ\push{c},X)=\Val_\rmdp(\Psi,X),
 \]
 because $\push{1/c}\push{c}\nu=\nu$ for every law $\nu$ (here $c>0$). So the answer to the
 threshold question is the same, and the threshold $t$ is unchanged.
 \end{itemize}
 Both maps are computable in logarithmic space by part~(c), since the law functional and
 the threshold are copied or, for $\Psi\circ\push{1/c}$ in the standard cases,
 changed by the parameter rescaling of Corollary~\ref{cor:standard}.

\medskip
\emph{Part~(c).}
We use the standard explicit encoding: lists of states and actions, the vertex lists $V_{s,a}$ (or the transition distributions), observation tables, and rewards, with all numbers in binary. For the forward construction, a logspace transducer outputs the states $s$ and pairs $(s,a)$, labels each vertex by its position in the input, copies the coordinates of every vertex verbatim, copies the observation tables, and copies $r(s,a,s')$ to every $\pmin$-transition $((s,a),v,s')$. For the backward construction, the transducer outputs the tagged states $(s,b)$ with $\delta(t,b)(s)>0$ for some $t$ (found by a scan of the input) together with $(s_0,\anew)$, the states $d_t$, the singleton uncertainty sets (copies of $\delta(s,a)$ with renamed targets), the vertices $\delta'_{t,b}$ (copies of $\delta(t,b)$ with tagged targets), the observation tables (with the bits for the agent), and the rewards. Each output symbol depends on a constant number of positions in the input, which can be stored with $O(\log n)$ bits, and the output size is polynomial. Since logspace reductions compose, the threshold problems are logspace-interreducible in the sense stated in Theorem~\ref{thm:umbrella}. \qed

\subsection{Scaling identities for corollary~\ref{cor:standard}}
\label{app:scale}

 We will use how these functionals react to rescaling: for every $c>0$,
\begin{equation}
\label{eq:scaling}
\begin{aligned}
\E(\push{c}\mu)&=c\,\E(\mu), & \MV_\lambda(\push{c}\mu)&=c\,\MV_{c\lambda}(\mu),\\
\CVaR_\alpha(\push{c}\mu)&=c\,\CVaR_\alpha(\mu), & \ERisk_\theta(\push{c}\mu)&=c\,\ERisk_{c\theta}(\mu).
\end{aligned}
\end{equation}

Let $X$ have law $\mu$ and let $c>0$. Clearly $\E(cX)=c\,\E(X)$, and $\E(cX)-\lambda\Var(cX)=c\bigl(\E(X)-c\lambda\Var(X)\bigr)$. Substituting $\eta=c\eta'$ gives
\[
\CVaR_\alpha(cX)=\sup_{\eta'}\Bigl\{c\eta'-\tfrac1\alpha\E[(c\eta'-cX)^+]\Bigr\}=c\,\CVaR_\alpha(X),
\]
and
\[
\ERisk_\theta(cX)=\tfrac1\theta\log\E[e^{(c\theta)X}]=c\cdot\tfrac1{c\theta}\log\E[e^{(c\theta)X}]=c\,\ERisk_{c\theta}(X).
\]

\section{Proofs for Section~\ref{sec:complexity}}
\label{app:complexity}

\subsection{Auxiliary facts} \label{app:auxfacts}

\begin{enumerate}
\item[(F1)] \emph{Monotone maps.} Let $\beta\colon[0,1]\to\Rbar$ be nondecreasing and continuous, and let $x_{\sigma,\pi}\in[0,1]$. Then $\beta(\inf B)=\inf\beta(B)$ and $\beta(\sup B)=\sup\beta(B)$ for every nonempty $B\subseteq[0,1]$, and therefore $\sup_\sigma\inf_\pi\beta(x_{\sigma,\pi})=\beta(\sup_\sigma\inf_\pi x_{\sigma,\pi})$.

\item[(F2)] \emph{Memoryless deterministic saddle points.} Let $f$ be an objective on the profiles of a fully informed game (in which players may use arbitrary history-dependent strategies), and let $\sigma^*,\pi^*$ be memoryless deterministic (MD) strategies with $\inf_\pi f(\sigma^*,\pi)=\sup_\sigma f(\sigma,\pi^*)=v$, where $\sigma$ and $\pi$ range over all history-dependent strategies. If $\Sigma'\ni\sigma^*$ and $\Pi'\ni\pi^*$ are classes of history-dependent strategies, then
\[
\sup_{\sigma\in\Sigma'}\inf_{\pi\in\Pi'}f(\sigma,\pi)=\inf_{\pi\in\Pi'}\sup_{\sigma\in\Sigma'}f(\sigma,\pi)=v.
\]
Indeed, $\inf_{\pi\in\Pi'}f(\sigma^*,\pi)\ge v$, since the infimum over the smaller class $\Pi'$ is at least the infimum over all strategies. Moreover, for every $\sigma\in\Sigma'$, we have $\inf_{\pi\in\Pi'}f(\sigma,\pi)\le f(\sigma,\pi^*)\le v$. Hence $\sup_{\sigma\in\Sigma'}\inf_{\pi\in\Pi'}f(\sigma,\pi)=v$. Symmetrically, $\sup_{\sigma\in\Sigma'}f(\sigma,\pi)\ge f(\sigma^*,\pi)\ge v$ for every $\pi\in\Pi'$, and $\sup_{\sigma\in\Sigma'}f(\sigma,\pi^*)\le v$, so $\inf_{\pi\in\Pi'}\sup_{\sigma\in\Sigma'}f(\sigma,\pi)=v$.
\end{enumerate}
We apply (F2) to TBSGs and to games $\fwd{\rmdp}$ obtained from RMDPs. In both, the strategies of Section~\ref{sec:prelim} are special history-dependent strategies. They include all MD strategies, because each player can base its choice on the current state: $\pmax$ observes it, and $\pmin$ may use a stationary assignment. Objectives with MD saddle points include reachability~\cite{Condon1992}, reward objectives~\cite{Shapley1953}, and the entropic risk of the total reward with nonnegative rewards~\cite{BaierEntropicRisk2023}. For these objectives, the values under our conventions coincide with the classical values.

\subsection{Derivation of the profiles}
\label{app:bernoulli}
Let $\mu=\Ber_v(p)$, so that $\int f\,\mathrm d\mu=(1-p)f(0)+p\,f(v)$ for every function $f$. Then:
\begin{itemize}
\item $\E(\mu)=pv$ and $\int x^2\,\mathrm d\mu=pv^2$, so $\Var(\mu)=pv^2-p^2v^2=v^2p(1-p)$. Hence $\beta^{\MV_\lambda}_v(p)=pv-\lambda v^2p(1-p)$.
\item $\int e^{\theta x}\,\mathrm d\mu=1-p+pe^{\theta v}$, so $\beta^{\ERisk_\theta}_v(p)=\frac1\theta\log(1-p+pe^{\theta v})$.
\item For $\CVaR_\alpha$, the function inside the supremum is
\[
g(\eta)=\eta-\tfrac1\alpha\bigl[(1-p)\,\eta^+ +p\,(\eta-v)^+\bigr].
\]
It is concave and piecewise linear with breakpoints at $0$ and $v$. Its slope is $1>0$ on $(-\infty,0)$, it is linear on $[0,v]$, and its slope is $1-\frac1\alpha<0$ on $(v,\infty)$ because $\alpha<1$. So the supremum is attained at $0$ or at $v$, that is,
\[
\beta^{\CVaR_\alpha}_v(p)=\max\{g(0),g(v)\}=\max\bigl\{0,\;v\bigl(1-\tfrac{1-p}\alpha\bigr)\bigr\}.
\]
\end{itemize}

\begin{lemma}
$\E$, $\CVaR_\alpha$ ($\alpha\in(0,1)$), and $\ERisk_\theta$ ($\theta\ne0$) have monotone Bernoulli profiles that are strict at the top, for every rational $v>0$. $\MV_\lambda$ ($\lambda\ge0$ rational) has such a profile for $v=1/(1+\lambda)$.
\label{lem:bernoulli-standard}
\end{lemma}
\begin{proof}
Fix a rational $v>0$. For each of the four functionals $\Psi$ we have $\beta^\Psi_v(1)=\Psi(\dirac{v})=v$, which is rational. It remains to show that $\beta^\Psi_v$ is nondecreasing, not constant on $(0,1)$, and satisfies $\beta^\Psi_v(p)<v$ for $p<1$.
\begin{itemize}
\item $\E$: $\beta(p)=pv$ is strictly increasing because $v>0$.
\item $\ERisk_\theta$: the function $h(p)=1-p+pe^{\theta v}$ is affine with slope $e^{\theta v}-1$, which is positive if $\theta>0$ and negative if $\theta<0$. So $\log h$ is strictly increasing if $\theta>0$ and strictly decreasing if $\theta<0$. In both cases $\beta(p)=\frac1\theta\log h(p)$ is strictly increasing, since $\frac1\theta$ has the sign of $\theta$.
\item $\CVaR_\alpha$: $\beta(p)=v\max\{0,1-(1-p)/\alpha\}$ is nondecreasing, because $1-(1-p)/\alpha$ is increasing in $p$. It equals $0$ for $p\le1-\alpha$ and is positive for $p>1-\alpha$, and $1-\alpha\in(0,1)$, so $\beta$ is not constant on $(0,1)$. For $p<1$ we have $(1-p)/\alpha>0$, hence $\beta(p)<v=\beta(1)$.
\item $\MV_\lambda$, with $v=1/(1+\lambda)$: then $\lambda v=\frac{\lambda}{1+\lambda}<1$. The derivative of $\beta(p)=pv-\lambda v^2p(1-p)$ is
$\beta'(p)=v-\lambda v^2(1-2p)\ge v-\lambda v^2=v(1-\lambda v)>0$ on $[0,1]$, since $1-2p\le1$. So $\beta$ is strictly increasing.
\end{itemize}
Whenever $\beta$ is strictly increasing, it is nonconstant on $(0,1)$ and $\beta(p)<\beta(1)$ for $p<1$, so the profile is strict at the top.

\end{proof}

\subsection{Proof of Theorem~\ref{thm:finite}}
\label{app:pspace-hard}
\cite{PapadimitriouTsitsiklis1987} show that the following problem is $\PSPACE$-hard: given a POMDP $\rmdp$ with nonnegative costs $c(s,a)$ and a horizon $N\le|S|$, decide whether some strategy has expected total cost $0$ over the first $N$ steps. Since the costs are nonnegative, the expected total cost of a strategy is $0$ if and only if the strategy plays, with probability $0$, a \emph{forbidden} pair $(s,a)$ with $c(s,a)>0$ during the first $N$ steps. So the problem asks, equivalently, for a set $B$ of forbidden state-action pairs (here $B=\{(s,a):c(s,a)>0\}$), whether some strategy avoids $B$ during the first $N$ steps with probability~$1$.

Given such an instance and the value $v$ from Definition~\ref{def:bernoulli}, we construct a POMDP $\rmdp'$ with the states $(s,k)$ for $s\in S$ and $k\in\{0,\dots,N-1\}$, where $k$ counts the steps, and two absorbing states $\bot$ and $\top$, which indicate failure and success. The initial state is $(s_0,0)$, the observations are $\Za((s,k))=(\Za(s),k)$, $\Za(\bot)=\bot$, and $\Za(\top)=\top$, and the available actions are $A((s,k))=A(s)$, with a single action in $\bot$ and in $\top$. The transitions are as follows:
\begin{itemize}
\item from $(s,k)$, a forbidden action $a$, i.e., one with $(s,a)\in B$, leads to $\bot$ with probability~$1$;
\item from $(s,k)$ with $k<N-1$, an allowed action $a$ leads to $(s',k+1)$ with the probability with which $a$ leads from $s$ to $s'$ in $\rmdp$;
\item from $(s,N-1)$, an allowed action leads to $\top$ with probability~$1$.
\end{itemize}
The reward function $r'$ of $\rmdp'$, which is defined on transitions $(x,a,y)$ like every reward function in this paper, assigns the reward $v$ to every transition $((s,N-1),a,\top)$ that enters $\top$, and the reward $0$ to all other transitions. Thus the $N$-th transition of a play earns $v$ if the first $N$ steps avoid $B$, and every other transition earns $0$.

Strategies of $\rmdp$ and $\rmdp'$ correspond: the observations of $\rmdp'$ additionally reveal the step counter $k$, which the agent knows anyway, as it is the length of its history minus one. Under corresponding strategies, the plays of $\rmdp'$ that reach $\top$ correspond to the plays of $\rmdp$ that avoid $B$ during the first $N$ steps. Hence the return $\Tot_N$ of $\rmdp'$ equals $v$ on these plays and $0$ on all others, so it has law $\Ber_v(p^\sigma)$, where $p^\sigma$ is the probability that $\sigma$ avoids $B$ during the first $N$ steps.

We claim that $\max_\sigma p^\sigma$ exists. The probability $p^\sigma$ depends only on the distributions $\sigma(w)\in\Dist(A)$ at the finitely many histories $w$ of length at most $N$, and it is a polynomial in the probabilities $\sigma(w)(a)$: it is a sum, over the finitely many prefixes of length $N$ that avoid $B$, of products of fixed transition probabilities and of factors $\sigma(w)(a)$. The set of these parameters is a product of finitely many simplices, which is compact, and a polynomial is continuous. So $p^\sigma$ attains its maximum $p^*=\max_\sigma p^\sigma$.

Since the environment of a POMDP has no choice and $\beta^\Psi_v$ is nondecreasing, $\Val_{\rmdp'}(\Psi,\Tot_N)=\sup_\sigma\beta^\Psi_v(p^\sigma)=\beta^\Psi_v(p^*)$: indeed $\beta^\Psi_v(p^\sigma)\le\beta^\Psi_v(p^*)$ for every $\sigma$, with equality for a strategy that attains $p^*$. As the profile is strict at the top, $\beta^\Psi_v(p^*)\ge\beta^\Psi_v(1)$ if and only if $p^*=1$, i.e., if and only if the answer to the original instance is positive. Here $\beta^\Psi_v(1)$ is a rational number, so this is a threshold test. The construction is computable in logarithmic space. Since every POMDP is an RPOMDP, the problem is $\PSPACE$-hard for RPOMDPs. For ac-POSGs, Theorem~\ref{thm:umbrella} applies with rescaling factor $1$ for finite-horizon total returns (and horizon $2N$ in the game), so the problem is $\PSPACE$-hard for ac-POSGs as well.

    \subsection{The gap lemma and the proof of Theorem~\ref{thm:undec}}\label{app:thegap}

\begin{lemma}
\label{lem:gap}
For all rationals $0<a<b<1$, there is a computable map from two-counter machines $\mathcal C$ to POMDPs $\rmdp_{\mathcal C}$ with a single observation and an absorbing target set $T$ such that: if $\mathcal C$ halts, then $\Prob^\sigma(\Diamond T)>b$ for some pure strategy $\sigma$; if $\mathcal C$ does not halt, then $\Prob^\sigma(\Diamond T)\le a$ for all strategies~$\sigma$.
\end{lemma}

\begin{proof}

\emph{The construction.} \cite{CondonLipton1989} (see also the account by \cite{Madani2003}) construct from $\mathcal C$, a rational $\epsilon>0$, and an integer $K\ge1$ a probabilistic finite automaton $\mathcal A$ with a finite alphabet $\Sigma$, a finite set of states $Q$, an initial state $q_0$, an absorbing accepting state $q_{\mathrm{acc}}$, and an absorbing rejecting state $q_{\mathrm{rej}}$, such that, writing $\Pr(w\text{ accepted})$ for the probability that $\mathcal A$ is in $q_{\mathrm{acc}}$ after reading the word $w$ from $q_0$:
\begin{itemize}
\item if $\mathcal C$ does not halt, then $\Pr(w\text{ accepted})\le\epsilon$ for every word $w\in\Sigma^*$;
\item if $\mathcal C$ halts, then there is a word $w$ with $\lim_{n\to\infty}\Pr(w^n\text{ accepted})=1-2^{-K}$.
\end{itemize}
In particular, $q_0\ne q_{\mathrm{acc}}$, since otherwise the empty word would be accepted with probability $1>\epsilon$ for non-halting machines. We use these properties of $\mathcal A$, and construct a POMDP $\rmdp_{\mathcal C}$ from $\mathcal A$ and then perform the  analysis of arbitrary (randomized) strategies of this POMDP as below. Given the rationals $a<b$, choose $\epsilon=a$ and an integer $K$ with $1-2^{-K}>b$. Let $\rmdp_{\mathcal C}$ be the POMDP with the state set $Q$, the initial state $q_0$, the action set $\Sigma$ at every state, and the transition probabilities $\delta(q,\alpha)(q')$ equal to the probability that $\mathcal A$ moves from $q$ to $q'$ when it reads the letter $\alpha$. All states have the same observation, so the POMDP has a single observation. Let $T=\{q_{\mathrm{acc}}\}$; it is absorbing in $\rmdp_{\mathcal C}$ because $q_{\mathrm{acc}}$ is absorbing in $\mathcal A$.

\emph{Strategies of $\rmdp_{\mathcal C}$.} A strategy depends only on the history of observations. Since all states have the same observation, the history of length $i$ is determined by $i$. A strategy $\sigma$ is therefore a sequence $(\sigma_i)_{i\ge0}$ of distributions over $\Sigma$, where $\sigma_i$ is used at step $i$. It is pure if every $\sigma_i$ is a Dirac distribution, that is, $\sigma$ is an infinite word. In particular, the letters played by $\sigma$ do not depend on the states visited.

\emph{Reaching $T$ within $n$ steps.} Let $E_n$ be the event that the state after $n$ steps lies in $T$. Since $T$ is absorbing, once a play is in $T$ it stays there, so $E_n$ is the event of reaching $T$ within the first $n$ steps. Hence $E_1\subseteq E_2\subseteq\cdots$ and $\bigcup_nE_n=\Diamond T$.

Fix a strategy $\sigma=(\sigma_i)_{i\ge0}$ and $n\ge1$. The event $E_n$ depends only on the first $n$ steps, so $\Prob^\sigma(E_n)$ is the sum of the probabilities of the finitely many prefixes $q_0w_0q_1w_1\cdots w_{n-1}q_n$ with $q_n\in T$. At step $i$, the strategy plays the letter $w_i$ with probability $\sigma_i(w_i)$, independently of the previous states and letters, because it has no information other than the step number $i$; then the state moves from $q_i$ to $q_{i+1}$ with probability $\delta(q_i,w_i)(q_{i+1})$. So such a prefix has probability
\[
\prod_{i<n}\sigma_i(w_i)\,\delta(q_i,w_i)(q_{i+1})=\Bigl(\prod_{i<n}\sigma_i(w_i)\Bigr)\cdot\prod_{i<n}\delta(q_i,w_i)(q_{i+1}).
\]

Group the prefixes by the word $w=w_0\cdots w_{n-1}\in\Sigma^n$ that they contain. The prefixes that contain a fixed word $w$ are those with an arbitrary sequence of states $q_1,\dots,q_n$ with $q_n\in T$, that is, $q_n=q_{\mathrm{acc}}$. Their total probability is
\[
\sum_{\substack{q_1,\dots,q_n\in Q\\ q_n=q_{\mathrm{acc}}}}\Bigl(\prod_{i<n}\sigma_i(w_i)\Bigr)\prod_{i<n}\delta(q_i,w_i)(q_{i+1})
=\Bigl(\prod_{i<n}\sigma_i(w_i)\Bigr)\cdot\sum_{\substack{q_1,\dots,q_n\in Q\\ q_n=q_{\mathrm{acc}}}}\prod_{i<n}\delta(q_i,w_i)(q_{i+1}),
\]
where we have taken the factor $\prod_{i<n}\sigma_i(w_i)$ out of the sum, because it does not depend on the states. The remaining sum is the probability that $\mathcal A$ is in $q_{\mathrm{acc}}$ after reading $w$, that is, $\Pr(w\text{ accepted})$. To see this, let $\mu_k$ be the distribution over $Q$ of the state of $\mathcal A$ after reading the first $k$ letters of $w$. Then $\mu_0=\dirac{q_0}$ and $\mu_{k+1}(q')=\sum_{q}\mu_k(q)\,\delta(q,w_k)(q')$, and unfolding this recursion gives
\[
\mu_n(q_{\mathrm{acc}})=\sum_{\substack{q_1,\dots,q_n\in Q\\ q_n=q_{\mathrm{acc}}}}\prod_{i<n}\delta(q_i,w_i)(q_{i+1}),
\]
which is exactly the remaining sum. Summing over all words $w\in\Sigma^n$, we obtain
\[
\Prob^\sigma(E_n)=\sum_{w\in\Sigma^n}\Bigl(\prod_{i<n}\sigma_i(w_i)\Bigr)\Pr(w\text{ accepted}).\tag{$\dagger$}
\]

The weights $\lambda(w)=\prod_{i<n}\sigma_i(w_i)$, for $w\in\Sigma^n$, are nonnegative and sum to $1$:
\[
\sum_{w\in\Sigma^n}\lambda(w)=\prod_{i<n}\sum_{\alpha\in\Sigma}\sigma_i(\alpha)=1 .
\]
So the right-hand side of $(\dagger)$ is a weighted average of the numbers $\Pr(w\text{ accepted})$, $w\in\Sigma^n$, and therefore
\[
\min_{w\in\Sigma^n}\Pr(w\text{ accepted})\le\Prob^\sigma(E_n)\le\max_{w\in\Sigma^n}\Pr(w\text{ accepted}).
\]
For a pure strategy, $\lambda$ is the Dirac distribution on the first $n$ letters of the infinite word $\sigma$, and $(\dagger)$ says that $\Prob^\sigma(E_n)$ is the acceptance probability of this prefix.

\emph{If $\mathcal C$ halts.} Let $w$ be as above. Since $\lim_{k\to\infty}\Pr(w^k\text{ accepted})=1-2^{-K}>b$, there is $k$ with $\Pr(w^k\text{ accepted})>b$. Let $L=k|w|$ be the length of $w^k$, and let $\sigma$ be the pure strategy that plays the letters of $w^k$ in its first $L$ steps and an arbitrary fixed letter afterwards. For $i<L$, the distribution $\sigma_i$ is the Dirac distribution on the $i$-th letter of $w^k$. Hence, in $(\dagger)$ for $n=L$, the weight $\prod_{i<L}\sigma_i(u_i)$ of a word $u\in\Sigma^L$ is $1$ if $u=w^k$ and $0$ otherwise, so $(\dagger)$ reduces to $\Prob^\sigma(E_L)=\Pr(w^k\text{ accepted})>b$. Since $E_L\subseteq\Diamond T$, we get $\Prob^\sigma(\Diamond T)>b$.

\emph{If $\mathcal C$ does not halt.} Let $\sigma$ be an arbitrary strategy, randomized or not, and let $n\ge1$. Every word is accepted with probability at most $\epsilon=a$, so by $(\dagger)$,
\[
\Prob^\sigma(E_n)\le\max_{w\in\Sigma^n}\Pr(w\text{ accepted})\le a .
\]
Since the events $E_n$ increase to $\Diamond T$, we get $\Prob^\sigma(\Diamond T)=\lim_{n\to\infty}\Prob^\sigma(E_n)\le a$.
\end{proof}

\begin{proof}[Proof of Theorem~\ref{thm:undec}]
\emph{Choice of the threshold.} Let $v$ be as in Definition~\ref{def:bernoulli} and $\beta=\beta^\Psi_v$. Since $\beta$ is nondecreasing and not constant on $(0,1)$, there are $a'<b'$ in $(0,1)$ with $\beta(a')<\beta(b')$. Choose rationals $a$ and $b$ with $0<a\le a'$ and $b'\le b<1$; then, since $\beta$ is nondecreasing,
\[
\beta(a)\le\beta(a')<\beta(b')\le\beta(b).
\]
Choose a rational $c$ with $\beta(a)<c<\beta(b)$. The numbers $v$, $a$, $b$, and $c$ depend only on $\Psi$ and not on any machine.

\emph{The POMDPs.} Given a two-counter machine $\mathcal C$, let $\rmdp_{\mathcal C}$ and $T$ be as in Lemma~\ref{lem:gap} for these $a$ and $b$. We define the rewards for the two kinds of returns:
\begin{itemize}
\item for $\Tot$, the reward is $v$ on the transitions that enter $T$ (that is, from a state outside $T$ to a state in $T$) and $0$ on all other transitions; since $T$ is absorbing, at most one transition of a play enters $T$;
\item for $\LimAvg$, the reward is $v$ on the transitions inside $T$ (from a state in $T$ to a state in $T$) and $0$ on all other transitions.
\end{itemize}
The rewards are nonnegative, as required for $\Tot$. In both cases, $X=v\cdot\mathbf 1[\Diamond T]$ on every play. For $\Tot$, this holds because the play collects $v$ once if it reaches $T$ and $0$ otherwise. For $\LimAvg$, if the play reaches $T$ at time $\tau$, then all transitions after $\tau$ lie inside $T$ and earn $v$, so the averages converge to $v$; if the play never reaches $T$, all rewards are $0$.

\emph{The value.} For a strategy $\sigma$ of the agent, let $p^\sigma=\Prob^\sigma(\Diamond T)$. Then $X$ has the law $\Ber_v(p^\sigma)$, and, since the environment of a POMDP has no choice,
\[
\Val(\Psi,X)=\sup_\sigma\Psi\bigl(\Ber_v(p^\sigma)\bigr)=\sup_\sigma\beta(p^\sigma).
\]

\emph{The two cases.}
\begin{itemize}
\item If $\mathcal C$ halts, some strategy $\sigma$ has $p^\sigma>b$ by Lemma~\ref{lem:gap}. As $\beta$ is nondecreasing, $\Val\ge\beta(p^\sigma)\ge\beta(b)>c$.
\item If $\mathcal C$ does not halt, then $p^\sigma\le a$ for \emph{every} strategy $\sigma$, randomized ones included, by Lemma~\ref{lem:gap}. As $\beta$ is nondecreasing, $\beta(p^\sigma)\le\beta(a)$ for all $\sigma$, so $\Val\le\beta(a)<c$.
\end{itemize}
Hence $\Val>c$ (and therefore also $\Val\ge c$) holds if and only if $\mathcal C$ halts. Since $\mathcal C\mapsto\rmdp_{\mathcal C}$ is computable, the halting problem for two-counter machines reduces to both the threshold problem and the strict threshold problem for POMDPs with a single observation, with the fixed rational threshold $c$. Both are therefore undecidable. They are also undecidable for RPOMDPs, of which POMDPs are a special case.

\emph{ac-POSGs.} We apply the forward construction of Definition~\ref{def:fwd} to $\rmdp_{\mathcal C}$, regarded as an RPOMDP. For $\Tot$, the rescaling factor of Lemma~\ref{lem:fwd} is $1$, so the game has the same achievable return laws as $\rmdp_{\mathcal C}$, and the same threshold $c$ separates the two cases. For $\LimAvg$, Lemma~\ref{lem:fwd} halves the average return. We therefore use the reward $2v$, instead of $v$, on the transitions inside $T$ before applying the construction. Then the average return of the game has the law $\Ber_v(p^\sigma)$ again, and the same argument applies. Hence both threshold problems are also undecidable for ac-POSGs.
\end{proof}

\subsection{Proof of Theorem~\ref{thm:disc-cvar}} \label{app:thmdiscvar}

\begin{lemma}
\label{lem:cvar-positive}
For every nonnegative random variable $Z$ and every $\alpha\in(0,1)$: $\CVaR_\alpha(Z)>0$ if and only if $\Prob(Z>0)>1-\alpha$.
\end{lemma}

\begin{proof}
Let $Z\ge0$ and $\alpha\in(0,1)$, let $m=\Prob(Z=0)$, and let
\[
g(\eta)=\eta-\tfrac1\alpha\E\bigl[(\eta-Z)^+\bigr],\qquad\text{so that}\qquad\CVaR_\alpha(Z)=\sup_{\eta\in\R}g(\eta).
\]
Since $Z\ge0$, we have $0\le(\eta-Z)^+\le\eta^+$, so the expectation is finite and $g$ is well defined. Because $Z\ge0$ we also have $\Prob(Z>0)=1-m$, so the claim is equivalent to
\[
\CVaR_\alpha(Z)>0\iff m<\alpha .
\]

\emph{The values $\eta\le0$.} If $\eta\le0$, then $\eta-Z\le0$, so $(\eta-Z)^+=0$ and $g(\eta)=\eta\le0$. In particular $g(0)=0$, so $\CVaR_\alpha(Z)\ge0$ always, and positivity can only come from some $\eta>0$.

\emph{Case $m\ge\alpha$.}
Let $\eta>0$. Writing $1=\mathbf 1[Z=0]+\mathbf 1[Z>0]$, we have
\[
(\eta-Z)^+=(\eta-Z)^+\,\mathbf 1[Z=0]+(\eta-Z)^+\,\mathbf 1[Z>0]
=\eta\,\mathbf 1[Z=0]+(\eta-Z)^+\,\mathbf 1[Z>0]\ \ge\ \eta\,\mathbf 1[Z=0],
\]
where the second equality holds because $(\eta-0)^+=\eta$ (as $\eta>0$), and the inequality holds because $(\eta-Z)^+\mathbf 1[Z>0]\ge0$. Taking expectations gives
\[
\E\bigl[(\eta-Z)^+\bigr]\ge\eta\,\Prob(Z=0)=\eta m\ge\eta\alpha,
\]
where the last inequality uses $m\ge\alpha$.

 Hence
\[
g(\eta)\le\eta-\tfrac1\alpha\,\eta\alpha=0 .
\]
Together with the values $\eta\le0$, we get $g\le0$ on $\R$, and $g(0)=0$, so $\CVaR_\alpha(Z)=\sup_\eta g(\eta)=0$. Thus $\CVaR_\alpha(Z)>0$ fails.

\emph{Case $m<\alpha$.} Let $\eta>0$. Since $Z\ge0$, we have $0\le(\eta-Z)^+\le\eta$, and $(\eta-Z)^+=0$ if $Z\ge\eta$. Hence $(\eta-Z)^+\le\eta\cdot\mathbf 1[Z<\eta]$, so $\E[(\eta-Z)^+]\le\eta\,\Prob(Z<\eta)$ and
\[
g(\eta)\ge\eta-\tfrac1\alpha\,\eta\,\Prob(Z<\eta)=\eta\Bigl(1-\frac{\Prob(Z<\eta)}{\alpha}\Bigr).
\]
As $\eta\downarrow0$, the events $\{Z<\eta\}$ decrease to $\bigcap_{\eta>0}\{Z<\eta\}=\{Z\le0\}=\{Z=0\}$, where the last equality holds because $Z\ge0$. By continuity of the probability measure from above, $\Prob(Z<\eta)\to m<\alpha$. So there is $\eta>0$ with $\Prob(Z<\eta)<\alpha$, and for this $\eta$ the right-hand side above is positive. Hence $\CVaR_\alpha(Z)\ge g(\eta)>0$.

The two cases cover all possibilities, so $\CVaR_\alpha(Z)>0$ if and only if $m<\alpha$, that is, if and only if $\Prob(Z>0)=1-m>1-\alpha$.
\end{proof}

\begin{proof}[Proof of Theorem~\ref{thm:disc-cvar}]
\emph{Parameters.} Let $\alpha\in(0,1)$ be given. Choose rationals $a,b\in(0,1)$ with $a<1-\alpha<b$ (for example $a=\frac{1-\alpha}2$ and $b=\frac{2-\alpha}2$). For a two-counter machine $\mathcal C$, let $\rmdp_{\mathcal C}$ and $T$ be as in Lemma~\ref{lem:gap} for these $a$ and $b$. Recall that $T$ is absorbing and that the initial state of $\rmdp_{\mathcal C}$ is not in $T$. Give $\rmdp_{\mathcal C}$ the reward $1$ on the transitions from a state outside $T$ into $T$ and $0$ on all other transitions. Since $T$ is absorbing, at most one transition of a play earns the reward $1$. For a strategy $\sigma$, write $p^\sigma=\Prob^\sigma(\Diamond T)$.

\emph{Step 1: the sign of the discounted return.} Let $\gamma\in(0,1)$ be any discount factor, and let $\rho$ be a play of $\rmdp_{\mathcal C}$. If $\rho$ enters $T$ by its $\tau$-th transition, $\tau\ge1$, then the reward $1$ is earned at that transition and nowhere else, so $\Disc_\gamma(\rho)=\gamma^{\tau-1}>0$. If $\rho$ never enters $T$, then all rewards of $\rho$ are $0$ and $\Disc_\gamma(\rho)=0$. Since the initial state is not in $T$, a play enters $T$ if and only if it reaches $T$. Hence, for every discount factor $\gamma\in(0,1)$,
\[
\Disc_\gamma(\rho)\ge0\ \text{ for every play }\rho,\qquad\text{and}\qquad\{\rho:\Disc_\gamma(\rho)>0\}=\Diamond T ,\tag{1}
\]
where $\Diamond T$ denotes the set of plays that visit $T$.

\emph{Step 2: the key equivalence for POMDPs.} In a POMDP the environment has no choice, so $\Val_{\rmdp_{\mathcal C}}(\CVaR_\alpha,\Disc_\gamma)=\sup_\sigma\CVaR_\alpha\bigl(\law^\sigma_{\Disc_\gamma}\bigr)$. By~(1), $\Disc_\gamma$ is a nonnegative random variable with $\Prob^\sigma(\Disc_\gamma>0)=p^\sigma$, so Lemma~\ref{lem:cvar-positive} gives
\[
\CVaR_\alpha\bigl(\law^\sigma_{\Disc_\gamma}\bigr)>0\iff p^\sigma>1-\alpha .\tag{2}
\]
Since a supremum of real numbers is positive if and only if one of the numbers is positive,
\[
\Val_{\rmdp_{\mathcal C}}(\CVaR_\alpha,\Disc_\gamma)>0\iff\exists\sigma\colon p^\sigma>1-\alpha .\tag{3}
\]

\emph{Step 3: the two cases of the gap.} If $\mathcal C$ halts, Lemma~\ref{lem:gap} gives a strategy $\sigma$ with $p^\sigma>b>1-\alpha$, so the right-hand side of~(3) holds. If $\mathcal C$ does not halt, then $p^\sigma\le a<1-\alpha$ for every strategy $\sigma$, so the right-hand side of~(3) fails. Hence
\[
\Val_{\rmdp_{\mathcal C}}(\CVaR_\alpha,\Disc_\gamma)>0\iff\mathcal C\text{ halts}.
\]
The map $\mathcal C\mapsto\rmdp_{\mathcal C}$ is computable, and the halting problem of two-counter machines is undecidable. Therefore it is undecidable whether $\Val(\CVaR_\alpha,\Disc_\gamma)>0$ for POMDPs with a single observation, and hence for RPOMDPs, of which POMDPs are the special case with singleton uncertainty sets.

\emph{Step 4: ac-POSGs.} Let $\beta\in(0,1)$ be a rational discount factor of the game, and let $\sg=\fwd{\rmdp_{\mathcal C}}$, where $\rmdp_{\mathcal C}$ is regarded as an RPOMDP whose uncertainty sets are singletons. Take the discount factor $\gamma=\beta^2$, which is rational, in Steps 1 to 3. By Lemma~\ref{lem:fwd} with $X=\Disc_{\beta^2}$, $X^\sg=\Disc_\beta$, and $c=\beta$, there is a bijection $\sigma\mapsto\sigma^\sg$ from the agent strategies of $\rmdp_{\mathcal C}$ to the $\pmax$-strategies of $\sg$ such that, for every $\sigma$,
\[
\bigl\{\law^{\sigma^\sg,\pi'}_{\Disc_\beta}:\pi'\in\Pi_\sg\bigr\}=\bigl\{\push{\beta}\law^{\sigma,\pi}_{\Disc_{\beta^2}}:\pi\in\Pi_{\rmdp_{\mathcal C}}\bigr\}.
\]
Since $\CVaR_\alpha(\push{\beta}\mu)=\beta\,\CVaR_\alpha(\mu)$ by~\eqref{eq:scaling} in \ref{app:scale}, taking the infimum of $\CVaR_\alpha$ over both sets and then the supremum over $\sigma$ gives
\[
\Val_\sg(\CVaR_\alpha,\Disc_\beta)=\beta\cdot\Val_{\rmdp_{\mathcal C}}(\CVaR_\alpha,\Disc_{\beta^2}).
\]
As $\beta>0$, the left-hand side is positive if and only if $\Val_{\rmdp_{\mathcal C}}(\CVaR_\alpha,\Disc_{\beta^2})>0$, which holds if and only if $\mathcal C$ halts, by Step 3 applied with $\gamma=\beta^2$.  Since $\fwd{\rmdp_{\mathcal C}}$ is computable from $\mathcal C$ (as the composition of the computable map $\mathcal C\mapsto\rmdp_{\mathcal C}$ and the construction $\fwd{\cdot}$), the strict threshold problem for discounted CVaR is undecidable for ac-POSGs with every rational discount factor $\beta\in(0,1)$.

\end{proof}
\subsection{Proof of Proposition~\ref{prop:approx}} \label{app:propapprox}
\emph{(a)} Let $\rmdp$ be an RPOMDP whose uncertainty sets $\Unc(s,a)$ are convex polytopes with rational vertex sets $V_{s,a}$, let $N$ be the horizon, and let all rewards, discount factors, risk parameters ($\lambda$, $\alpha$, $\theta$) and the threshold $c$ be rational. The agent maximises and the environment minimises. Write $\rhd$ for either $\ge$ or $>$. We show that $\Val\rhd c$ is decidable. For ac-POSGs, the claim follows from Theorem~\ref{thm:umbrella}(b), or directly by the same argument (see the end of the proof).

\emph{Step 1: the strategy spaces are finite products of simplices.} Strategies depend only on observation histories, and only histories of length less than $N$ matter. Let $H$ be the finite set of observation histories of the agent of length less than $N$, and $G$ the finite set of histories of length less than $N$ available to the environment. A strategy of the agent is a family $x=(x_{h,a})$ of reals with $x_{h,a}\ge0$ and $\sum_ax_{h,a}=1$ for each $h\in H$. A strategy of the environment selects, for each $g\in G$ and each pair $(s,a)$, a point of $\Unc(s,a)=\mathrm{conv}(V_{s,a})$. Every such point has the form $\sum_{v}y_{g,s,a,v}\,v$ with $y_{g,s,a,v}\ge0$ and $\sum_vy_{g,s,a,v}=1$, and conversely each such family $y$ defines a point. (Randomising over points of a convex set yields the same transition law as the mean point, so nothing is lost.) Let $X$ and $Y$ be the sets of all such $x$ and $y$. They are compact, being finite products of simplices.

\emph{Step 2: probabilities of prefixes are polynomials.} Let $u=s_0a_0s_1\cdots a_{N-1}s_N$ be a play prefix of length $N$, let $h_i\in H$ and $g_i\in G$ be the histories of the agent and of the environment along $u$ before step $i$ (they are determined by $u$), and let $\mu_0$ be the initial distribution. Then
\[
p_u(x,y)\;=\;\Pr(\Cyl(u))\;=\;\mu_0(s_0)\prod_{i<N}x_{h_i,a_i}\cdot\sum_{v\in V_{s_i,a_i}}y_{g_i,s_i,a_i,v}\,v(s_{i+1}),
\]
a polynomial of degree at most $2N$ in the variables $x,y$ with rational coefficients. The return is a function of the prefix, which takes a rational value $x_u$ on $\Cyl(u)$. Let $\mathcal V$ be the (finite) set of these values. The law of the return under $(x,y)$ is $\sum_up_u(x,y)\,\delta_{x_u}$.

\emph{Step 3: the value of the profile is a continuous function $f(x,y)$.} Write $p_u=p_u(x,y)$.
\begin{itemize}
\item $\E$: $f=\sum_up_u\,x_u$.
\item $\MV_\lambda$: $f=\sum_up_u\,x_u-\lambda\bigl(\sum_up_u\,x_u^2-(\sum_up_u\,x_u)^2\bigr)$.
\item $\CVaR_\alpha$: $f=\max_{z\in\mathcal V}g_z$ with $g_z=z-\tfrac1\alpha\sum_up_u\,(z-x_u)^+$.
\item $\ERisk_\theta$ ($\theta\ne0$): $f=\tfrac1\theta\log S$ with $S=\sum_up_u\,e^{\theta x_u}$.
\end{itemize}
In the first two cases $f$ is a polynomial in the $p_u$. For $\CVaR_\alpha$ we use the Rockafellar-Uryasev formula $\CVaR_\alpha=\sup_{\eta\in\mathbb R}\varphi(\eta)$, where
\[
\varphi(\eta)=\eta-\tfrac1\alpha\sum_up_u\,(\eta-x_u)^+ ,
\]
and show that the supremum is attained at a point of $\mathcal V$. Let $z_1<\dots<z_m$ be the elements of $\mathcal V$, so that every $x_u$ is one of the $z_j$. There are three cases for $\eta$.
\begin{itemize}
\item $\eta\le z_1$. Then $\eta\le x_u$ for all $u$, so every term $(\eta-x_u)^+$ is $0$ and $\varphi(\eta)=\eta\le z_1=\varphi(z_1)$.
\item $\eta\ge z_m$. Then $(\eta-x_u)^+=\eta-x_u$ for all $u$, so $\varphi(\eta)=(1-\tfrac1\alpha)\,\eta+\tfrac1\alpha\sum_up_u\,x_u$. As $\alpha\le1$, the coefficient of $\eta$ is at most $0$, hence $\varphi(\eta)\le\varphi(z_m)$.

\item $z_j\le\eta\le z_{j+1}$. No $x_u$ lies strictly between $z_j$ and $z_{j+1}$, so for each $u$ either $x_u\le z_j$ and $(\eta-x_u)^+=\eta-x_u$ on the whole interval, or $x_u\ge z_{j+1}$ and $(\eta-x_u)^+=0$ on the whole interval. Hence $\varphi(\eta)=a+b\,\eta$ on $[z_j,z_{j+1}]$ for some constants $a,b$ (namely $b=1-\frac1\alpha\sum_{u:\,x_u\le z_j}p_u$). Write $\eta=(1-t)z_j+t\,z_{j+1}$ with $t\in[0,1]$. Then $\varphi(\eta)=(1-t)\varphi(z_j)+t\,\varphi(z_{j+1})\le\max\{\varphi(z_j),\varphi(z_{j+1})\}$, since a weighted average of two numbers is at most the larger of them.
\end{itemize}
In every case $\varphi(\eta)\le\max_j\varphi(z_j)$, and $\max_j\varphi(z_j)\le\sup\varphi$, so $\CVaR_\alpha=\max_{z\in\mathcal V}\varphi(z)=\max_{z\in\mathcal V}g_z$. Here $(z-x_u)^+$ is a rational constant because $z,x_u\in\mathcal V$, so each $g_z$ is a polynomial in the $p_u$. For $\ERisk_\theta$, $S\ge\min_ue^{\theta x_u}>0$, so $f$ is well defined and continuous. In all cases $f$ is continuous on $X\times Y$.

\emph{Step 4: quantifier form.} Since $Y$ is compact and $f$ is continuous, $g(x)=\min_{y\in Y}f(x,y)$ is attained and continuous in $x$, and since $X$ is compact, $\max_{x\in X}g(x)$ is attained. Hence $\Val=\max_x\min_yf(x,y)$, and
\[
\Val\rhd c\iff\exists x\in X\;\forall y\in Y:\;f(x,y)\rhd c,
\]
for both choices of $\rhd$. (For $>$: $\max_xg(x)>c$ iff some $x$ has $\min_yf(x,y)>c$ iff some $x$ has $f(x,y)>c$ for all $y$, because the minimum is attained.)

\emph{Step 5: $\E$, $\MV_\lambda$, $\CVaR_\alpha$.} The right-hand side of the equivalence is a sentence of the first-order theory of the reals. The memberships $x\in X$ and $y\in Y$ are conjunctions of polynomial (in)equalities with rational coefficients, and $f(x,y)\rhd c$ is a polynomial inequality in the first two cases and a finite disjunction $\bigvee_{z\in\mathcal V}\,g_z(x,y)\rhd c$ of polynomial inequalities in the third. The theory is decidable by Tarski's quantifier elimination~\cite{Tarski1951}.

\emph{Step 6: $\ERisk_\theta$ with rational $\theta$.} Tarski's theorem covers only formulas built from polynomials, whereas $\ERisk_\theta$ involves $\exp$ and $\log$. The first-order theory of the reals with the exponential function is only known to be decidable under Schanuel's conjecture, so we cannot apply it to the exponential sentence directly. Instead we observe that, for rational data, the exponential enters only through one fixed real number, and we reduce to a polynomial sentence in which this number appears as a single parameter.

\emph{Reduction to one parameter.} Since $\log$ is increasing, $\frac1\theta\log S\rhd c$ is equivalent to $S\rhd e^{\theta c}$ if $\theta>0$, and to the same inequality with the direction reversed if $\theta<0$. The numbers $\theta x$ with $x\in\mathcal V\cup\{c\}$ are rational, so there is a positive integer $D$ with $\theta x\in\frac1D\mathbb Z$ for all of them. Put $\tau=e^{1/D}>1$. Then $e^{\theta x_u}=\tau^{m_u}$ and $e^{\theta c}=\tau^{m_c}$ for integers $m_u,m_c$, and the condition reads
\[
\sum_up_u(x,y)\,\tau^{m_u}\;\rhd\;\tau^{m_c}
\]
(direction reversed if $\theta<0$). Choose $M\in\mathbb N$ with $m_u+M\ge0$ for all $u$ and $m_c+M\ge0$, and multiply by $\tau^M>0$ to remove negative exponents. Let $\varphi(z)$ be the formula
\[
\exists x\in X\;\forall y\in Y:\;\sum_up_u(x,y)\,z^{m_u+M}\rhd z^{m_c+M}
\]
(with the direction reversed if $\theta<0$), with one free variable $z$. It is a first-order formula over the reals with rational coefficients, and by Step 4, $\Val\rhd c$ holds if and only if $\varphi(\tau)$ holds. So it remains to decide $\varphi(\tau)$.

\emph{Why $\varphi(\tau)$ is decidable.} The idea is that $\varphi(z)$ is a polynomial sentence, so the set of $z$ satisfying it has a simple shape, and $\tau$ lies strictly inside one piece of that shape. We proceed in three parts.

\emph{(i) The shape of $Z=\{z>0:\varphi(z)\}$.} By Tarski's quantifier elimination~\cite{Tarski1951}, we can compute a quantifier-free formula $\psi(z)$ equivalent to $\varphi(z)$ over the reals. It is a Boolean combination of atoms $p_j(z)=0$ and $p_j(z)>0$ for finitely many polynomials $p_j$ with rational coefficients. Let $0=e_0<e_1<\dots<e_k$ be $0$ together with all positive real roots of  $p_j$ which is not the zero polynomial; these are algebraic numbers and can be computed. Let $J$ be an open interval between two consecutive points of $\{e_0,\dots,e_k,\infty\}$. Then no $p_j$ has a root in $J$. If some $p_j$ took values of both signs on $J$, say $p_j(a)>0>p_j(b)$ with $a,b\in J$, then by the intermediate value theorem $p_j$ would have a root between $a$ and $b$, hence in $J$, which is impossible. So every $p_j$ has constant sign on $J$, every atom has the same truth value on $J$, and so does $\psi$.

Fix an interval $J$ as above. Every atom $p_j(z)=0$ or $p_j(z)>0$ has the same truth value at all $z\in J$: this is clear if $p_j\equiv0$, and otherwise $p_j$ has constant sign on $J$ and no root there, so $p_j(z)=0$ is false throughout $J$ and $p_j(z)>0$ is either true throughout $J$ or false throughout $J$. The formula $\psi$ is built from these atoms using $\neg$, $\wedge$ and $\vee$, and these connectives preserve the property of having the same truth value at all points of $J$ (by induction on the structure of $\psi$, the truth value of a negation, conjunction or disjunction is determined by the truth values of its parts, which are constant on $J$). Hence $\psi$ has the same truth value at all points of $J$. Since $\psi$ is equivalent to $\varphi$ and $J\subseteq(0,\infty)$, either $\varphi(z)$ holds for all $z\in J$, that is, $J\subseteq Z$, or $\varphi(z)$ fails for all $z\in J$, that is, $J\cap Z=\emptyset$.

Hence either $J\subseteq Z$ or $J\cap Z=\emptyset$.

\emph{(ii) $\tau$ lies strictly inside one interval $J$.} The number $\tau=e^{1/D}$ is not algebraic: otherwise $e=\tau^D$ would be algebraic, contradicting the transcendence of $e$~\cite{Baker1975}. In particular $\tau\ne e_i$ for every $i$, so $\tau$ belongs to exactly one of the open intervals $J$ from (i), and $\tau\in Z$ if and only if that whole interval $J$ is contained in $Z$.

\emph{(iii) Finding the interval.} We need not know $\tau$ exactly, only which interval contains it. For $n=1,2,\dots$, compute a rational interval $I_n\subseteq(1,\infty)$ of length at most $2^{-n}$ that contains $\tau$ (for instance from the power series of $e^{1/D}$ with an explicit bound on the remainder), and refine rational isolating intervals of $e_1,\dots,e_k$ to length at most $2^{-n}$. Stop at the first $n$ for which $I_n$ is disjoint from all of these isolating intervals. This happens after finitely many steps, because by (ii) $\tau$ has positive distance from each $e_i$. At that point no $e_i$ lies in $I_n$, so $I_n$ is contained in a single interval $J$, namely the one that contains $\tau$. Let $q$ be the left endpoint of $I_n$. By (i), $\tau\in Z$ if and only if $q\in Z$, that is, if and only if the sentence $\varphi(q)$ holds. This sentence has only rational constants, so it is decided by Tarski's theorem.

(For entropic risk in a rational base, as in~\cite{BaierEntropicRisk2023}, the constants $b^{\theta x}$ are real algebraic, and Tarski's theorem applies directly.)

\emph{Conclusion.} The decision procedure for $\Val\rhd c$ with $\ERisk_\theta$ is therefore as follows. Compute $D$, the exponents $m_u,m_c$ and $M$, and the formula $\varphi(z)$. Compute the quantifier-free formula $\psi(z)$ and the algebraic numbers $e_1<\dots<e_k$. Compute the rational number $q$ as in (iii). Decide the sentence $\varphi(q)$ by Tarski's theorem and output its answer. This is correct because
\[
\Val\rhd c\;\iff\;\varphi(\tau)\;\iff\;\tau\in Z\;\iff\;q\in Z\;\iff\;\varphi(q),
\]
where the first equivalence is Step 4 together with the reduction above, the third holds because $\tau$ and $q$ lie in the same interval $J$ and $J\subseteq Z$ or $J\cap Z=\emptyset$ by (i), and the others are the definitions of $Z$. Every step is effective, and the procedure terminates because $\tau$ is not one of the $e_i$ by (ii). Hence $\Val\rhd c$ is decidable for $\ERisk_\theta$.

\emph{Step 7: ac-POSGs.} The same argument works directly for an ac-POSG with a finite horizon. Apply the claim for RPOMDPs to the game obtained from the equivalence of Theorem~\ref{thm:umbrella}(b).

\emph{(b)} We proceed in five steps. Let $R=\max|r|$ be the largest absolute reward and $M=R/(1-\gamma)$.

\emph{Step 1: truncation.} Let $X_N=\sum_{t<N}\gamma^tr_t$ be the discounted reward of the first $N$ steps. On every play,
\[
|\Disc_\gamma-X_N|=\Bigl|\sum_{t\ge N}\gamma^tr_t\Bigr|\le R\sum_{t\ge N}\gamma^t=\frac{\gamma^NR}{1-\gamma}=:\epsilon_N,
\]
and both $\Disc_\gamma$ and $X_N$ take values in $[-M,M]$.

\emph{Step 2: the functionals are Lipschitz.} Let $Z,Z'$ be random variables on the same probability space with values in $[-M,M]$ and $|Z-Z'|\le\epsilon$ everywhere; we will apply this to $Z=\Disc_\gamma$ and $Z'=X_N$ under the same profile, with $\epsilon=\epsilon_N$. We show $|\Psi(Z)-\Psi(Z')|\le K\epsilon$ (writing $\Psi(Z)$ for the value of $\Psi$ at the law of $Z$), with $K=1$ for $\E$, $\CVaR_\alpha$, and $\ERisk_\theta$, and $K=1+4\lambda M$ for $\MV_\lambda$.
\begin{itemize}
\item $|\E Z-\E Z'|\le\E|Z-Z'|\le\epsilon$.
\item $\CVaR_\alpha$ and $\ERisk_\theta$ are \emph{monotone}: $Z\le Z'$ everywhere implies $\Psi(Z)\le\Psi(Z')$. For $\CVaR_\alpha$, this holds because $(\eta-Z)^+\ge(\eta-Z')^+$ for every $\eta$, so the expression under the supremum in its definition is smaller for $Z$. For $\ERisk_\theta$, this holds because $\frac1\theta\log\E[e^{\theta Z}]$ increases with $Z$: for $\theta>0$, both $z\mapsto e^{\theta z}$ and $y\mapsto\frac1\theta\log y$ are increasing, and for $\theta<0$, both are decreasing. They are also \emph{translation equivariant}: $\Psi(Z+m)=\Psi(Z)+m$ for every constant $m$, as one sees by substituting $\eta=\eta'+m$ in the definition of $\CVaR_\alpha$, and from $\E[e^{\theta(Z+m)}]=e^{\theta m}\E[e^{\theta Z}]$ for $\ERisk_\theta$. Since $Z'-\epsilon\le Z\le Z'+\epsilon$, we get $\Psi(Z')-\epsilon\le\Psi(Z)\le\Psi(Z')+\epsilon$.
\item For the variance, $|\Var Z-\Var Z'|\le|\E Z^2-\E Z'^2|+|(\E Z)^2-(\E Z')^2|$. Since $|Z^2-Z'^2|=|Z-Z'|\cdot|Z+Z'|\le2M\epsilon$, the first term is at most $2M\epsilon$; likewise, $|(\E Z)^2-(\E Z')^2|=|\E Z-\E Z'|\cdot|\E Z+\E Z'|\le2M\epsilon$. Hence $|\MV_\lambda(Z)-\MV_\lambda(Z')|\le\epsilon+4\lambda M\epsilon$.
\end{itemize}

\emph{Step 3: the values are close.} By Step~2, $|\Psi(\law^{\sigma,\pi}_{\Disc_\gamma})-\Psi(\law^{\sigma,\pi}_{X_N})|\le K\epsilon_N$ for every profile $(\sigma,\pi)$. If two functions on profiles differ by at most $\delta$ everywhere, then so do their infima over $\pi$ and the suprema of these over $\sigma$. Hence $|\Val(\Psi,\Disc_\gamma)-\Val(\Psi,X_N)|\le K\epsilon_N$.

\emph{Step 4: the truncated values are computable.} Let $X_N=\sum_{t<N}\gamma^t r_t$ be the discounted return truncated after $N$ steps, where $\gamma$ is rational and $r_t$ is the reward of the $t$-th transition. It depends only on the prefix of length $N$ of the play, and on each such prefix it takes a rational value. The decision procedure of~(a) uses the return only through these values $x_u$ (Step~2 of~(a)), so it applies to $X_N$ unchanged, even though the reward earned by a transition now depends on the time step. Hence, for every rational $q$, we can decide whether $\Val(\Psi,X_N)\ge q$.

Moreover, the value lies in a rational interval $[l_0,u_0]$ that we can compute. Let $M$ be the largest absolute reward and $R=M\sum_{t<N}\gamma^t$, so that $|X_N|\le R$ on every play. For $\E$, $\CVaR_\alpha$ and $\ERisk_\theta$, the value lies between the smallest and the largest value of the return, so $[l_0,u_0]=[-R,R]$ works. For $\MV_\lambda$, the variance is at most $R^2$, so $[l_0,u_0]=[-R-\lambda R^2,R]$ works.

Now fix a precision $\delta>0$. We run bisection. Maintain rationals $l\le\Val(\Psi,X_N)\le u$, starting with $l=l_0$ and $u=u_0$. In each round let $m=\frac{l+u}2$ and use~(a) to decide whether $\Val(\Psi,X_N)\ge m$. If so, set $l:=m$, and otherwise set $u:=m$. The invariant is preserved, and the length $u-l$ halves in each round. After $\lceil\log_2((u_0-l_0)/\delta)\rceil$ rounds, $u-l\le\delta$, so $l$ approximates $\Val(\Psi,X_N)$ to within $\delta$.

\emph{Step 5: approximation and semi-decidability.} Given $\delta>0$, choose $N$ with $K\epsilon_N\le\delta/2$, which is possible since $\epsilon_N\to0$, and compute $\Val(\Psi,X_N)$ up to precision $\delta/2$ by Step~4; by Step~3, the result is within $\delta$ of $\Val(\Psi,\Disc_\gamma)$. For the strict threshold problem, we claim that $\Val(\Psi,\Disc_\gamma)>c$ holds if and only if $\Val(\Psi,X_N)>c+K\epsilon_N$ for some $N$. If $\Val(\Psi,X_N)>c+K\epsilon_N$, then $\Val(\Psi,\Disc_\gamma)\ge\Val(\Psi,X_N)-K\epsilon_N>c$ by Step~3. Conversely, if $\Val(\Psi,\Disc_\gamma)>c$, choose $N$ with $2K\epsilon_N<\Val(\Psi,\Disc_\gamma)-c$; then $\Val(\Psi,X_N)\ge\Val(\Psi,\Disc_\gamma)-K\epsilon_N>c+K\epsilon_N$. Since $K$ and $\epsilon_N$ are rational, each condition $\Val(\Psi,X_N)>c+K\epsilon_N$ is a strict threshold problem with a rational threshold, which is decidable by Step~4. Checking these conditions for $N=1,2,\dots$ and accepting as soon as one holds is a procedure that halts exactly when $\Val(\Psi,\Disc_\gamma)>c$, so the strict threshold problem is recursively enumerable.

\emph{ac-POSGs.} Steps~3 to~5 use the model only through part~(a), and through the bound $|\Disc_\gamma-X_N|\le K\epsilon_N$ on every play, which does not depend on the model. Part~(a) holds for ac-POSGs (Step~7 of its proof), so the same argument gives the approximation and the semi-decidability of the strict threshold problem for ac-POSGs, with the same $K$, $\epsilon_N$ and the same discount factor $\gamma$.

\subsection{Proof of Theorem~\ref{thm:ssg}}\label{app:thmssg}

We reduce the SSG problem to both threshold problems for $\Psi$ and $X$ on RMDPs. We first view the SSG as a TBSG, shift its threshold (only for CVaR), and add rewards; we then translate the TBSG into an RMDP, determine the value of the RMDP, and finally compute a rational threshold.

\emph{The SSG as a TBSG.} An SSG~\cite{Condon1992} has $\pmax$-, $\pmin$-, and random vertices, a $0$-sink, and a $1$-sink;
$\pmax$ wants to reach the $1$-sink, and $\pmin$ wants to prevent this. We view it as a TBSG in the sense of section~\ref{sec:posg}: a random vertex becomes a $\pmax$-state with a single action that moves to each of its two successors with probability $\frac12$, and the $1$-sink becomes an absorbing target state $T$,
and the $0$-sink becomes an absorbing non-target state. Thus $\pmax$ wins exactly on the plays that reach $T$.

 By a transformation of \cite{Condon1992}, which replaces every edge by a gadget of polynomially many random vertices, we may assume that the game is \emph{stopping}, i.e., that it reaches a sink with probability~$1$ under all strategies. Then the value $x^*$ of the initial vertex, i.e., the probability of reaching $T$ that $\pmax$ can guarantee, is a rational number with denominator at most $4^n$, where $n$ is the number of vertices~\cite{Condon1992}. The task is to decide whether $x^*>\frac12$.

\emph{Threshold shift.} The CVaR profile $\beta^{\CVaR_\alpha}_v(p)=v\max\{0,1-(1-p)/\alpha\}$ is $0$ for all $p\le1-\alpha$, so it cannot distinguish small probabilities. For $\Psi=\CVaR_\alpha$, we therefore add a new initial vertex, a $\pmax$-state with a single action that moves to $T$ with probability $\mu=1-\alpha$ and to the old initial vertex otherwise; the game remains stopping. Let $x'$ be the value of the new initial vertex. Then $x'=\mu+(1-\mu)x^*\in[1-\alpha,1]$, which lies in the range where the CVaR profile is strictly increasing. As $x'$ is a strictly increasing affine function of $x^*$, we have $x^*>\frac12$ if and only if $x'>q$ for $q=\mu+(1-\mu)\frac12=1-\frac\alpha2$. For the other objectives, let $\mu=0$, $x'=x^*$, and $q=\frac12$.

\emph{Rewards.} Let $v=1/(1+\lambda)$ for $\MV_\lambda$ and $v=1$ otherwise. For $\Tot$, we redirect every transition into $T$ to a fresh state $g$ that moves to an absorbing sink $z$; every transition into $g$ earns $v$, and all other transitions earn $0$. This ensures that the reward $v$ is collected exactly once on plays that reach the target, since the absorbing state $T$ itself would be entered again at every step.

For $\LimAvg$, every transition into a state of the absorbing set $T$ earns $v$, and all other transitions earn $0$. In both cases, rewards depend only on the target state of a transition. On every play that reaches $T$ (for $\Tot$: that reaches $g$), the return $X$ equals $v$: for $\Tot$, the reward $v$ is collected exactly once, and for $\LimAvg$, all rewards from some point on equal $v$. On all other plays, all rewards are $0$, so $X=0$. Hence every profile $(\sigma,\pi)$ induces the law $\Ber_v(p^{\sigma,\pi})$, where $p^{\sigma,\pi}$ is the probability of reaching $T$.

Here, the TBSG constructed need not be alternating, and for this we use the following variant of the backward construction \ref{def:bwd}. The construction normally records the action of $\pmin$ in the state it leads to; this record only serves to transfer rewards that depend on the action of $\pmin$. In our case, all rewards depend only on the target state of a transition, so we omit the record and keep the states of the TBSG $\sg$: the agent plays its actions in the $\pmax$-states, where the environment has no choice, and the dummy action $\anew$ in every $\pmin$-state $t$, where the environment chooses from $\conv\{\delta(t,b):b\in A(t)\}$; the transition from $t$ to $s'$ earns the reward of the target state $s'$.

As both players observe the states of the TBSG, the result is an RMDP $\rmdp$ with the same achievable return laws as $\sg$; we explain why.

\emph{Why $\rmdp$ and $\sg$ have the same achievable return laws.} The states, the actions of the agent in the $\pmax$-states, and the rewards of $\rmdp$ and $\sg$ coincide, and the agent observes the current state in both. The two models differ only in the $\pmin$-states $t$: in $\sg$, $\pmin$ chooses a mixed action $\lambda$ on $A(t)$, and the next state is distributed according to $\sum_b\lambda(b)\,\delta(t,b)$; in $\rmdp$, the environment chooses a point $\mu\in\conv\{\delta(t,b):b\in A(t)\}$ directly, and the next state is distributed according to $\mu$. Every $\lambda$ yields such a point, and every such point arises from some $\lambda$. Hence a strategy of $\pmin$ in $\sg$ induces a strategy of the environment in $\rmdp$ (map each history to the point obtained from the mixed action), and conversely a strategy of the environment induces a strategy of $\pmin$ (map each history to a fixed decomposition $\lambda$ of the selected point, chosen once for each point). In both directions the strategy of the agent is unchanged, and the probability of every finite play prefix is the product of the initial probability, the probabilities of the agent's actions, and the next-state probabilities, which are the same in both models. Hence the two profiles induce the same probability measure on plays. Since the rewards depend only on the states, they induce the same law of the return, so $\rmdp$ and $\sg$ have the same achievable return laws.

\emph{From the TBSG to an RMDP.} Applying this variant to the shifted TBSG above, which need not be alternating, yields an RMDP $\rmdp$ with the same achievable return laws as the shifted TBSG. Together with the rewards defined above, every profile of $\rmdp$ therefore induces the law $\Ber_v(p)$, where $p$ is the probability of reaching $T$ under the corresponding profile $(\sigma,\pi)$ of the shifted TBSG; we write $p^{\sigma,\pi}$ for it.

\emph{The value of $\rmdp$.} Write $\beta=\beta^\Psi_v$. We compute $\Val_\rmdp(\Psi,X)$ in four steps.
\begin{enumerate}
\item \emph{Reduce to a function of $p$.} By the definition of the profile $\beta$ and of the value,
\[
\Val_\rmdp(\Psi,X)=\sup_\sigma\inf_\pi\Psi\bigl(\Ber_v(p^{\sigma,\pi})\bigr)=\sup_\sigma\inf_\pi\beta(p^{\sigma,\pi}),
\]
where $\sigma$ and $\pi$ range over the strategies of the TBSG.
\item \emph{$\beta$ is continuous and nondecreasing.} For $\E$, $\CVaR_\alpha$ and $\ERisk_\theta$ this is the content of Lemma~\ref{lem:bernoulli-standard}. For $\MV_\lambda$ it holds because $\lambda v<1$: the derivative of $\beta^{\MV_\lambda}_v$ is $v-\lambda v^2(1-2p)\ge v(1-\lambda v)>0$.
\item \emph{Move $\beta$ outside.} By~(F1), a continuous nondecreasing function commutes with suprema and infima. Applying this to the infimum over $\pi$ for each fixed $\sigma$, and then to the supremum over $\sigma$, we get
\[
\Val_\rmdp(\Psi,X)=\beta\Bigl(\sup_\sigma\inf_\pi p^{\sigma,\pi}\Bigr).
\]
\item \emph{Identify the argument of $\beta$.} The quantity $\sup_\sigma\inf_\pi p^{\sigma,\pi}$ is the value of the reachability game for our classes of strategies. In the stopping game, both players have optimal MD strategies~\cite{Condon1992}, and these belong to our classes, since both players observe the current state of the TBSG. By~(F2), the value for our classes therefore equals the classical value of the game, that is, the value $x'$ of its initial vertex in the (shifted) TBSG. Hence $\Val_\rmdp(\Psi,X)=\beta(x')$.
\end{enumerate}

\emph{Thresholds.} By the previous step, $\Val_\rmdp(\Psi,X)=\beta(x')$, where $x'\ge\mu$ is the value of the shifted game, and we know that $x^*>\frac12$ if and only if $x'>q$. It remains to translate the question whether $x'>q$ into threshold questions about $\beta(x')$.

\emph{(i) A gap for the value of the SSG.} The value $x^*$ is a rational number $p/d$ with $d\le4^n$~\cite{Condon1992}. If $x^*>\frac12$, then $2p>d$, so $2p-d\ge1$ as an integer, and hence $x^*-\frac12=(2p-d)/(2d)\ge1/(2\cdot4^n)$. Thus
\[
x^*>\tfrac12\iff x^*\ge\tfrac12+\tfrac1{2\cdot4^n}.
\]

\emph{(ii) The gap for $x'$.} Recall that $x'=\mu+(1-\mu)x^*$ and $q=\mu+(1-\mu)\frac12$. The map $x\mapsto\mu+(1-\mu)x$ is strictly increasing, since $\mu<1$, and it sends $\frac12$ to $q$ and $\frac12+\frac1{2\cdot4^n}$ to $q+g$ with $g=(1-\mu)/(2\cdot4^n)$. Applying it to both sides of the equivalence in (i), we get
\[
x'>q\iff x'\ge q+g .
\]
Moreover, $\mu\le q<q+g\le1$, because $\frac12+\frac1{2\cdot4^n}\le1$. So $q$, $q+g$, and $x'$ all lie in $[\mu,1]$, which is the interval on which $\beta$ is strictly increasing (it is $[0,1]$ for $\E$, $\MV_\lambda$, and $\ERisk_\theta$ since $\mu=0$ there, and $[1-\alpha,1]$ for $\CVaR_\alpha$).

\emph{(iii) Thresholds for $\E$, $\MV_\lambda$, and $\CVaR_\alpha$.} Since $\beta$ is strictly increasing on $[\mu,1]$, $a>b$ holds if and only if $\beta(a)>\beta(b)$, and $a\ge b$ holds if and only if $\beta(a)\ge\beta(b)$, for $a,b\in[\mu,1]$. Together with (ii) and $\Val_\rmdp(\Psi,X)=\beta(x')$ this gives
\[
x^*>\tfrac12\iff x'>q\iff\Val_\rmdp(\Psi,X)>\beta(q)\iff\Val_\rmdp(\Psi,X)\ge\beta(q+g).
\]
For $\E$, $\MV_\lambda$, and $\CVaR_\alpha$, the profile $\beta$ is a polynomial, respectively $\beta(p)=pv$, $\beta(p)=pv-\lambda v^2p(1-p)$, and $\beta(p)=v(1-(1-p)/\alpha)$ on $[1-\alpha,1]$, with rational coefficients. The numbers $q$ and $g$ are rationals of bit size $O(n)$, so $\beta(q)$ and $\beta(q+g)$ are rationals of polynomial size, obtained by a constant number of arithmetic operations. Hence the instances ``is $\Val_\rmdp(\Psi,X)>\beta(q)$?'' of the strict threshold problem and ``is $\Val_\rmdp(\Psi,X)\ge\beta(q+g)$?'' of the non-strict one are computed in polynomial time, and each is a yes-instance if and only if $x^*>\frac12$.

\emph{(iv) Thresholds for $\ERisk_\theta$.} Here $\beta(p)=\frac1\theta\log(1-p+pe^{\theta v})$, and the numbers $\beta(q)$ and $\beta(q+g)$ are in general irrational, while the threshold of a threshold problem must be rational. We therefore choose a rational $c$ between them. The derivative
\[
\beta'(p)=\frac{e^{\theta v}-1}{\theta\,(1-p+pe^{\theta v})}
\]
is positive on $[0,1]$, since the numerator and $\theta$ have the same sign and $1-p+pe^{\theta v}$ is a convex combination of $1$ and $e^{\theta v}$. Therefore $\beta'(p)\ge C:=|e^{\theta v}-1|/(|\theta|\max\{1,e^{\theta v}\})>0$, a constant that depends only on $\theta$ and $v$, and fix a rational $0<C'\le C$. By the mean value theorem, $\beta(q+g)-\beta(q)\ge C'g$. Compute rationals $b_1,b_2$ with $|b_1-\beta(q)|\le C'g/4$ and $|b_2-\beta(q+g)|\le C'g/4$, which takes polynomial time since $g\ge2^{-O(n)}$ requires only $O(n)$ bits of precision, and let $c=(b_1+b_2)/2$. Let $m=(\beta(q)+\beta(q+g))/2$. Then $|c-m|\le C'g/4$, whereas $m-\beta(q)=\beta(q+g)-m\ge C'g/2$, so
\[
\beta(q)<c<\beta(q+g).
\]
Now if $x^*>\frac12$, then $x'\ge q+g$ by (ii), so $\Val_\rmdp(\Psi,X)=\beta(x')\ge\beta(q+g)>c$. If $x^*\le\frac12$, then $x'\le q$, so $\Val_\rmdp(\Psi,X)=\beta(x')\le\beta(q)<c$. Hence both ``is $\Val_\rmdp(\Psi,X)>c$?'' and ``is $\Val_\rmdp(\Psi,X)\ge c$?'' are yes-instances if and only if $x^*>\frac12$.

In all cases, both threshold problems for $\rmdp$ decide whether $x^*>\frac12$, with a threshold that is computed in polynomial time.

\emph{Complexity of the reduction.} All steps run in polynomial time. For $\E$, $\MV_\lambda$, and $\CVaR_\alpha$, they even run in logarithmic space: the transformation into a stopping game, the threshold shift, and the new reward states are local modifications of the game; the construction of the RMDP from the shifted TBSG is computable in logarithmic space, like the backward construction (Theorem~\ref{thm:umbrella}(c)); and the thresholds are obtained by a constant number of additions and multiplications on rationals of polynomial size. For $\ERisk_\theta$, all steps except the computation of the threshold $c$ are as above; the computation of $c$ approximates logarithms and exponentials, which is in polynomial time. \qed

\subsection{Proofs of Theorem~\ref{cor:expected}} \label{app:thmcorexp}

\begin{proof}
\emph{The source problem.} A \emph{discounted TBSG} is a tuple $G=(\Smax,\Smin,A,\delta,r,s_0,\gamma)$ with disjoint finite sets of states $\Smax,\Smin$ ($S=\Smax\cup\Smin$), finite action set $A$ and let $A(x)$ denote the set of effective actions at state $x$, transitions $\delta(x,a)\in\Dist(S)$, rewards $r(x,a,y)\in\Q$, an initial state $s_0$, and a rational discount factor $\gamma\in(0,1)$ given in binary. The owner of state $x$ chooses the action in $x$, and the play collects $\sum_{n\ge0}\gamma^nr_n$. The players need not alternate. For $V\in\R^S$, $x\in S$, and $a\in A(x)$, let
\[
f_{x,a}(V)=\sum_y\delta(x,a)(y)\bigl(r(x,a,y)+\gamma V(y)\bigr),
\]
and let $\mathcal G\colon\R^S\to\R^S$ be given by $\mathcal G(V)(x)=\max_{a\in A(x)}f_{x,a}(V)$ for $x\in\Smax$ and $\mathcal G(V)(x)=\min_{a\in A(x)}f_{x,a}(V)$ for $x\in\Smin$. Since $\delta(x,a)$ is a distribution, $|f_{x,a}(V)-f_{x,a}(V')|\le\gamma\|V-V'\|_\infty$. Moreover, for real numbers $u_i,v_i$ we have $|\max_iu_i-\max_iv_i|\le\max_i|u_i-v_i|$, and the same for $\min$. Hence $\|\mathcal G(V)-\mathcal G(V')\|_\infty\le\gamma\|V-V'\|_\infty$, so $\mathcal G$ is a contraction of $(\R^S,\|\cdot\|_\infty)$ and, by Banach's fixed-point theorem, has a unique fixed point $W$ (Shapley's equations). By \cite{Shapley1953}, $W(s_0)$ is the value of the game $G$, and we write $\Val_G=W(s_0)$. The \emph{threshold problem for TBSGs} asks, given $G$ and a rational $c$, whether $\Val_G\ge c$.

\cite{AnderssonMiltersen2009} study discounted games of this kind (in their terminology, discounted Gillette games, where the reward depends only on the state and the action), with discount factor, rewards, and probabilities in binary. They prove that \emph{solving} such a game, that is, computing optimal pure positional strategies, is polynomial-time equivalent to solving simple stochastic games, and they note that solving a game is polynomial-time Turing equivalent to comparing its value with a given number. Hence SSGs reduce to the threshold problem for discounted TBSGs as follows: deciding whether an SSG has value above $\frac12$ reduces to solving the SSG, which reduces to solving a discounted game, which reduces under polynomial-time Turing reductions to threshold queries for discounted games. In their games, the reward depends only on the state and the action. Such a game is a discounted TBSG in our sense, with $r(x,a,y):=r(x,a)$ for all successors $y$, and it has the same value.

\emph{Step 1: the RMDP.} Given $G$, we build an RMDP $\rmdp_G$ with the same discount factor $\gamma$, in which the choice of $\pmin$ becomes a choice of the environment. Its states are $Q=\{(x,\tau):x\in S,\ \tau\in\{\bot\}\cup\Amin\}$, where $\Amin=\bigcup_{t\in\Smin}A(t)$, and the tag $\tau$ records the action of $\pmin$ in the preceding move, or $\bot$ if that move was not made by $\pmin$. The initial state is $(s_0,\bot)$.  Both observation functions are injective (we take $\Oa=\Oe=Q$ and $\Za=\Ze=\mathrm{id}$), so $\rmdp_G$ is an RMDP, and the agent and the environment see the whole state, including the tag. The agent therefore knows the last action of $\pmin$, which $\pmax$ does not see in $G$. This does no harm, as Step~2 computes the value of $\rmdp_G$ directly: the optimal agent strategy ignores the tag, and the bound for the optimal environment strategy holds against all agent strategies.

\begin{itemize}
\item For $x\in\Smax$ and every tag $\tau$, the agent has the actions $A(x)$, and $\Unc((x,\tau),a)=\{\hat\delta_{x,a}\}$ with $\hat\delta_{x,a}((y,\bot))=\delta(x,a)(y)$. The transition from $(x,\tau)$ under $a$ to $(y,\bot)$ earns $r(x,a,y)$.
\item For $t\in\Smin$ and every tag $\tau$, the agent has the single action $\anew$, and
\[
\Unc((t,\tau),\anew)=\conv\{\hat\delta_{t,b}:b\in A(t)\},\qquad \hat\delta_{t,b}((y,b))=\delta(t,b)(y).
\]
The transition from $(t,\tau)$ to $(y,b)$ earns $r(t,b,y)$. This is well defined because the target determines $b$.
\end{itemize}
Every polytope has at most $|A(t)|$ vertices, so $\rmdp_G$ has polynomial size, and it is computable in logarithmic space, since every entry is a local copy of an entry of $G$.
A point of $\Unc((t,\tau),\anew)$ is a mixture of actions of $\pmin$, and the tag of the successor $(y,b)$ records which action $b$ was drawn. The tag is needed for the reward: the transition earns $r(t,b,y)$, which depends on $b$, and in an RMDP the reward of a transition may depend only on its source state, its action, and its target state. With the tag in the target, the reward is paid on the same transition, with no extra step, which would change the timing and hence the discounting.

\emph{Step 2: the value of $\rmdp_G$.} We show $\Val_{\rmdp_G}(\E,\Disc_\gamma)=W(s_0)$. Since $\Psi=\E$, $\Val_{\rmdp_G}(\E,\Disc_\gamma)=\sup_\sigma\inf_\pi\E^{\sigma,\pi}[\Disc_\gamma]$. Let $U(x,\tau)=W(x)$ and $R=\max|r|$. Fix an agent strategy $\sigma$ and an environment strategy $\pi$, let $x_0a_0x_1a_1\cdots$ be the random play, let $r_n$ be the reward of its $n$-th transition, let $\mathcal F_n$ be generated by the prefix $x_0a_0\cdots x_n$, and let
\[
Z_n=\sum_{k<n}\gamma^kr_k+\gamma^nU(x_n).
\]
Given $\mathcal F_n$, the action $a_n$ is drawn from a distribution $\mu_n$ on $A(x_n)$ determined by $\sigma$; after $a_n=a$, the environment has fixed a point $p^a_n\in\Unc(x_n,a)$ determined by $\pi$, and $x_{n+1}\sim p^a_n$. Let $\Phi_n(a)=\sum_zp^a_n(z)\bigl(\hat r(x_n,a,z)+\gamma U(z)\bigr)$, where $\hat r$ is the reward function of $\rmdp_G$.
  Since $Z_{n+1}-Z_n=\gamma^n\bigl(r_n+\gamma U(x_{n+1})-U(x_n)\bigr)$ and $x_n$ is $\mathcal F_n$-measurable, it suffices to average $r_n+\gamma U(x_{n+1})$ given $\mathcal F_n$: first over $x_{n+1}\sim p_n^a$ for fixed $a_n=a$, which gives $\Phi_n(a)$, then over $a_n\sim\mu_n$. Hence
\[
\E[Z_{n+1}-Z_n\mid\mathcal F_n]=\gamma^n\Bigl(\sum_a\mu_n(a)\Phi_n(a)-U(x_n)\Bigr).\tag{1}
\]

We evaluate $\Phi_n$ in the two kinds of states.
\begin{itemize}
\item $x_n=(x,\tau)$ with $x\in\Smax$: $p^a_n=\hat\delta_{x,a}$, so $\Phi_n(a)=f_{x,a}(W)$. This is $\le W(x)$ for every $a$, because $W=\mathcal G(W)$, and it equals $W(x)$ for an action $a^*(x)$ attaining the maximum.
\item $x_n=(t,\tau)$ with $t\in\Smin$: there is only the action $\anew$, and $p_n=\sum_b\lambda_b\hat\delta_{t,b}$ for some distribution $\lambda$ on $A(t)$. Since the successor $(y,b)$ has $U=W(y)$ and the transition earns $r(t,b,y)$, we get $\Phi_n=\sum_b\lambda_bf_{t,b}(W)$. This is $\ge\min_bf_{t,b}(W)=W(t)$, and it equals $W(t)$ if $\lambda$ is concentrated on an action $b^*(t)$ attaining the minimum.
\end{itemize}
By the tower property, $\E[Z_{n+1}-Z_n]=\E\bigl[\E[Z_{n+1}-Z_n\mid\mathcal F_n]\bigr]$, and $Z_0=U(x_0)=W(s_0)$ is deterministic. Moreover, $Z_n\to\Disc_\gamma$ on every play: the partial sums converge to $\Disc_\gamma$ because $|r_k|\le R$ and $\sum_k\gamma^kR<\infty$, and $|\gamma^nU(x_n)|\le\gamma^n\|W\|_\infty\to0$, where $\|W\|_\infty$ is finite because $S$ is finite. Also $|Z_n|\le R/(1-\gamma)+\|W\|_\infty$ for all $n$. Hence, by dominated convergence, $\E[Z_n]\to\E[\Disc_\gamma]$. We now use~(1) in the two directions.

\emph{Lower bound.} Let $\sigma^*$ play $a^*(x)$ in every state $(x,\tau)$ with $x\in\Smax$. Fix any environment strategy $\pi$. At states of $\Smax$ the right-hand side of~(1) is $0$ (as $\Phi_n(a^*(x))=W(x)$), and at states of $\Smin$ it is $\ge0$ (as $\Phi_n\ge W(t)$). So $\E[Z_{n+1}-Z_n\mid\mathcal F_n]\ge0$, hence $\E[Z_{n+1}]\ge\E[Z_n]$, and by induction $\E[Z_n]\ge Z_0=W(s_0)$ for all $n$. Letting $n\to\infty$ gives $\E^{\sigma^*,\pi}[\Disc_\gamma]\ge W(s_0)$. Since $\pi$ was arbitrary, $\inf_\pi\E^{\sigma^*,\pi}[\Disc_\gamma]\ge W(s_0)$, and therefore $\Val_{\rmdp_G}\ge W(s_0)$.

\emph{Upper bound.} Let $\pi^*$ be the environment strategy that always picks the vertex $\hat\delta_{t,b^*(t)}$ in $(t,\tau)$. Fix any agent strategy $\sigma$, randomized and history-dependent. At states of $\Smax$ the right-hand side of~(1) is $\le0$ (as $\Phi_n(a)\le W(x)$ for every $a$, hence also for the average over $\mu_n$), and at states of $\Smin$ it is $0$ (as $\Phi_n=W(t)$ for the chosen vertex). So $\E[Z_{n+1}]\le\E[Z_n]$, hence $\E[Z_n]\le Z_0=W(s_0)$ for all $n$, and in the limit $\E^{\sigma,\pi^*}[\Disc_\gamma]\le W(s_0)$. Therefore $\inf_\pi\E^{\sigma,\pi}[\Disc_\gamma]\le\E^{\sigma,\pi^*}[\Disc_\gamma]\le W(s_0)$ for every $\sigma$, and $\Val_{\rmdp_G}\le W(s_0)$.

Hence $\Val_{\rmdp_G}(\E,\Disc_\gamma)=W(s_0)=\Val_G$.

\emph{Step 3: the reduction.} By Step~2, $\Val_G\ge c$ if and only if $\Val_{\rmdp_G}(\E,\Disc_\gamma)\ge c$, and $\rmdp_G$ is a polytopic RMDP with the same binary discount factor, computable in logarithmic space from $G$. Given a polynomial-time Turing reduction from SSGs to the threshold problem for discounted TBSGs, we replace each oracle query $(G,c)$ by the query $(\rmdp_G,c)$ to the threshold problem for discounted RMDPs. This is a polynomial-time Turing reduction from SSGs to the latter problem. The same holds for the strict threshold problem, because the values coincide: any reduction from SSGs to a variant of the TBSG threshold problem yields a reduction to the same variant for RMDPs.
\end{proof}

\end{document}